\documentclass{article}

\usepackage{microtype}
\usepackage{graphicx}
\usepackage{subcaption}
\usepackage{booktabs}
\usepackage[table]{xcolor} 
\usepackage{hyperref}

\usepackage{dblfloatfix}
\usepackage{placeins}

\usepackage[accepted]{icml2026}

\usepackage{amsmath}
\usepackage{amssymb}
\usepackage{mathtools}
\usepackage{amsthm}
\usepackage[capitalize,noabbrev]{cleveref}

\usepackage[normalem]{ulem}

\usepackage{algorithm}
\usepackage{algorithmic}

\newcommand{\sg}[1]{\operatorname{stopgrad}\!\left(#1\right)}

\usepackage{multirow}
\usepackage{multicol}

\theoremstyle{plain}

\theoremstyle{definition}

\theoremstyle{remark}

\icmltitlerunning{RapTB: Rooted Absorbed Prefix Trajectory Balance with Submodular Replay}

\begin{document}

\twocolumn[

\icmltitle{Rooted Absorbed Prefix Trajectory Balance with Submodular Replay \\ for GFlowNet Training}

\begin{icmlauthorlist}
  \icmlauthor{Xi Wang}{xw}
  \icmlauthor{Wenbo Lu}{xw}
  \icmlauthor{Shengjie Wang}{xw}
\end{icmlauthorlist}
\icmlaffiliation{xw}{Courant Institute School of Mathematics, Computing, and Data Science, New York University}
\icmlcorrespondingauthor{Shengjie Wang}{sw5973@nyu.edu}

\icmlkeywords{Generative Flow Networks, Molecular Generation, Submodular Optimization, Replay Buffer}

\vskip 0.3in
]

\printAffiliationsAndNotice{}  

\begin{abstract}
Generative Flow Networks (GFlowNets) enable fine-tuning large language models to approximate reward-proportional posteriors, but they remain prone to mode collapse, manifesting as prefix collapse and length bias. We attribute this to two factors: (i) weak credit assignment to early prefixes, and (ii) biased replay that induces a shifted, non-representative training flow distribution. We propose Rooted absorbed prefix Trajectory Balance (\textbf{RapTB}), an objective that anchors subtrajectory supervision at the root and propagates terminal rewards to intermediate prefixes via absorbed suffix-based backups, providing dense prefix-level learning signals. To mitigate replay-induced distribution shift, we further introduce \textbf{SubM}, a submodular replay refresh strategy that promotes both high reward and diversity.
Empirically, on tasks such as molecule generation with LLM using SMILES strings, RapTB combined with SubM consistently improves optimization performance and molecular diversity while preserving high validity. The code is released on https://github.com/ComDec/ChemGFN.
\end{abstract}

\section{Introduction}
Generative Flow Networks (GFlowNets) learn a stochastic policy on a directed acyclic graph (DAG) that constructs objects sequentially, so that completed trajectories are sampled with probability proportional to the rewards~\citep{bengio2021flow,bengio2023gflownet, hu2023gflownet}. 
In contrast to reward-maximizing reinforcement learning, the objective of GFlowNets is distributional: spread probability mass across many high-reward modes in proportion to reward, rather than concentrating on a single optimum~\citep{kaelbling1996reinforcement}. 
The GFlowNet objective is distributional: spread probability mass across many high-reward modes in proportion to reward, rather than concentrating on a single optimum.
This objective admits an equivalent entropy-regularized RL formulation~\citep{tiapkin2024gflownet,deleu2024discrete}; our contributions operate within this off-policy regime, improving credit assignment and replay coverage for TB-family balance objectives on terminable prefix trees.
This method extends naturally to large language models (LLMs) in a terminable prefix-tree formulation~\citep{hu2024amortizing}.
Every prefix state has an explicit termination edge to a terminal node.
The termination edge can be implemented as an EOS action. \looseness=-1

In practice, LLM-GFlowNets suffer from mode collapse. We identify two specific failures: (i) prefix collapse, where early-token entropy drops sharply and distinct terminals share near-identical prefixes; and (ii) length bias, where the model favors sequences that are systematically too short or too long. We trace these issues to two factors: (i) weak credit assignment, as terminal-only rewards provide high-variance and ambiguous feedback for intermediate steps~\citep{madan2023learning}; and (ii) replay bias, where training is confined to a tiny fraction of the search space, and repeated reinforcement of this narrow subset causes the distribution to collapse~\citep{pmlr-v202-shen23a}.

We address these failure modes with two complementary mechanisms: one that strengthens prefix-level credit assignment and the other broadens the support of replay.
\textbf{RapTB} retains terminal Trajectory Balance (TB)~\citep{malkin2022trajectory} as the primary constraint and adds a lightweight rooted-prefix objective that provides supervision at intermediate prefixes. 
Concretely, RapTB densifies training signals by propagating the terminal reward to each prefix via suffix-based backups.
In parallel, \textbf{SubM} refreshes the replay buffer by selecting a subset of trajectories that maximizes a submodular objective over candidates. The objective jointly encourages high reward, trajectory diversity, and length coverage, expanding the support of the training distribution. Across molecular and arithmetic generation tasks, we find that TB objective~\citep{malkin2022trajectory} can quickly over-concentrate on shared prefixes, while Subtrajectory Balance (SubTB)~\citep{madan2023learning, hu2024amortizing} can drift in its termination probabilities. RapTB mitigates both prefix collapse and length bias, and SubM further improves coverage and distribution matching.

\paragraph{Contributions.}
\begin{itemize}
    \vspace{-0.8em}
    \item We empirically characterize mode collapse in LLM-GFlowNets as a reproducible combination of prefix collapse and length bias.
    We provide evidence that it is driven by high-variance terminal credit assignment and replay-induced training distribution shifts.
    \item We propose RapTB, which augments TB with a rooted prefix objective that propagates terminal reward to intermediate prefixes via suffix-based backups.
    It provides dense training signals and reduces variance.
    \item We introduce Submodular Replay (SubM), a replay refresh rule that balances reward, diversity, and length coverage in one submodular objective.
    It improves replay coverage and stabilizes training.
    We validate RapTB+SubM across five tasks, model scales up to 32B, and comparisons against RL and alternative GFlowNet baselines (Appendices~\ref{app:additional_baselines}--\ref{app:scaling}).
\end{itemize}

\paragraph{Positioning.}
We focus on TB-family objectives. While TB suffers from high variance under terminal rewards~\citep{madan2023learning}, existing subtrajectory methods for LLMs induce \textit{termination drift} via conflicting overlapping constraints.
RapTB resolves this by restricting dense supervision to rooted prefixes using variance-reduced absorbed targets.
This formulation eliminates destabilizing boundary conditions and explicitly detaches auxiliary termination gradients to prevent drift, all while preserving the global TB anchor.
Orthogonally, SubM stabilizes replay via coverage-aware subset refresh to mitigate the reward-tilted collapse observed in prior work.

\begin{figure*}[t]
\centering
\includegraphics[width=0.95\textwidth]{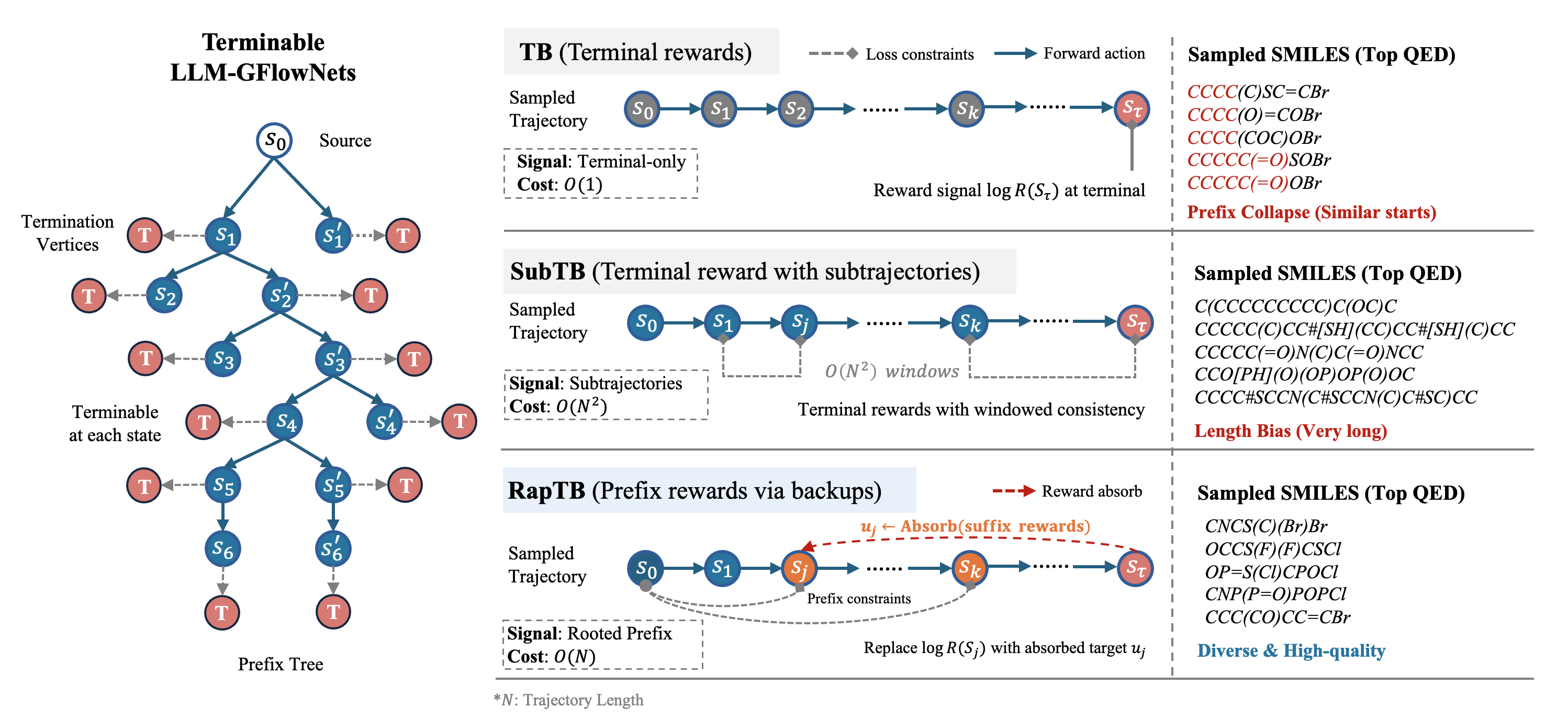}
\caption{\textbf{Training objectives for LLM-GFlowNets.}
TB uses only terminal reward $\log R(s_\tau)$ ($O(1)$). SubTB adds $O(N^2)$ windowed consistency constraints. RapTB replaces prefix stop-rewards with suffix-absorbed targets $u_j$ and applies $O(N)$ rooted prefix constraints. ($N$: trajectory length). The right column shows examples of generated molecules trained with different losses. QED (Quantitative Estimate of Drug-likeness) is a comprehensive metric for measuring a molecule's drug-likeness and is used as the task reward during training. \looseness=-1}

\label{fig:raptb}
\end{figure*}

\section{Related Work}
\textbf{GFlowNets and applications.}
GFlowNets learn generative policies whose terminal distribution is proportional to reward~\citep{bengio2021flow,bengio2023gflownet}. Trajectory Balance stabilizes training via global path consistency~\citep{malkin2022trajectory}, and subsequent work connects GFlowNets to variational inference and policy gradients~\citep{gflownet_variation,niu2024gflownet}. The framework has been applied broadly in scientific discovery~\citep{jain2022biological,jain2023multi,nguyen2023hierarchical,hernandez2023crystal}. To reduce variance, Subtrajectory Balance introduces dense subtrajectory constraints~\citep{madan2023learning}, but directly adapting it to terminable prefix trees for LLMs~\citep{hu2024amortizing} can introduce termination drift. RapTB targets this structural mismatch by restricting dense supervision to rooted prefixes while using suffix-absorbed targets for lower-variance prefix credit.

\textbf{Experience replay and exploration in GFlowNets.}
Experience replay improves sample efficiency in GFlowNets, but reward-prioritized replay can induce rich-get-richer collapse and reduce coverage~\citep{pmlr-v202-shen23a,vemgal2023an,hu2024amortizing}. Prior work mitigates this with heavier exploration mechanisms, including local search~\citep{kim2023local}, evolutionary population maintenance~\citep{ikram2024evolution}, MCTS-style rollouts~\citep{morozov2024improving}, and retrospective synthesis~\citep{he2024looking}. In contrast, we formulate replay refresh as lightweight submodular maximization with a greedy near-optimality guarantee~\citep{subM_bg1,subM_bg2,subM_bg3}, selecting a subset that jointly balances reward, diversity, and length coverage.
Concurrently, TBA~\citep{bartoldson2025tba} scales TB-based LLM post-training via asynchronous exploration and importance-weighted replay; its systems-level contributions are orthogonal to our objective-level improvements, and RapTB can in principle replace the TB loss within TBA's framework.
\section{Method}
\label{sec:method}

\subsection{GFlowNets on Terminable Prefix Trees}
\label{method:gfn-llm}
We consider a GFlowNet defined on a directed acyclic graph (DAG) with a unique source $s_0$ and terminal set $\mathcal{X}$.
A trajectory $\xi = (s_0 \to \cdots \to s_{\tau} \to x)$ ends at a terminal node $x$, and the target terminal distribution satisfies
$p^\star(x) \propto R(x)$~\citep{bengio2021flow,bengio2023gflownet}.

Following~\citet{hu2024amortizing}, we instantiate a GFlowNet on the prefix tree induced by an autoregressive language model augmented with a stop symbol $\top$ (i.e., each prefix has a termination action), such as an End-of-Sentence (EOS) token.
Given a prompt, the LLM generates a sequence of tokens. As the prompt is held constant for the generation process, we omit it in the following for simplicity. 
Therefore, the GFlowNet state is represented by a generated prefix $s_{0:i}$.

From state $s_{0:i}$, the forward policy either emits a token $s_{i+1}\in\mathcal{V}$ (the vocabulary) and transitions to $s_{0:i+1}$, or terminates by emitting $\top$.
We parameterize the forward policy by the language model distribution $q_\theta(\cdot\mid s_{0:i})$ over $\mathcal{V}\cup\{\top\}$, and identify
\[
\begin{aligned}
P_F^\theta(s_{0:i+1}\mid s_{0:i}) &= q_\theta(s_{i+1}\mid s_{0:i}), \\
P_F^\theta(s_{0:i}^\top\mid s_{0:i}) &= q_\theta(\top\mid s_{0:i}).
\end{aligned}
\]
where $s_{0:i}^{\top}$ denotes the terminated sequence of $s_{0:i}$ by appending the terminal token $\top$, i.e., $(s_0, s_1, \ldots,s_i,\top)$.

Because each non-root prefix has a unique parent, the backward kernel is deterministic, and the state space forms a tree. Therefore, the probability of sampling a terminated sequence $\xi = s_{0:\tau}^{\top}$ factorizes as
\[
q_\theta^{\top}(\xi)
=
\Big(\prod_{i=0}^{\tau-1} q_\theta(s_{i+1}\mid s_{0:i})\Big)\,
q_\theta(\top\mid s_{0:\tau}).
\]


\paragraph{Mixed reward.}
To stabilize exploration under sparse or high-variance task rewards, for any prefix $s_{0:i}$, we define the stop-reward as a mixture of
(i) a frozen reference language model prior and
(ii) an external task-specific score, following~\citet{hu2024amortizing}.
Let $P_{\mathrm{ref}}$ denote the probability assigned by a fixed, pre-trained reference LM to a terminated sequence, which serves as a prior that regularizes generation toward fluent and well-formed outputs.
For a prefix $s_{0:i}$, we define the mixed stop-reward as
\[
\log R(s_{0:i}^{\top})
=
\kappa \log P_{\mathrm{ref}}(s_{0:i}^{\top})
+
\lambda S(s_{0:i}^{\top}),
\]
where $S(s_{1:i}^{\top})$ is the task-only reward component and $\kappa,\lambda \ge 0$ control the mixture ratio.
The task-only component $S(\cdot)$ depends on the application; for example, in small-molecule generation it may correspond to a property score (e.g., binding affinity or drug-likeness), while in symbolic expression generation it may reflect functional correctness or coverage.
The exact reward definitions and ablation studies are provided in Appendix~\ref{app:reward_shaping}.

\subsection{Trajectory Balance: Global Path Consistency}

Trajectory Balance~\citep{malkin2022trajectory} enforces global consistency between forward trajectories and terminal rewards by introducing a learnable normalizer $Z_\theta>0$.
For a terminated trajectory $\xi = s_{0:\tau}^{\top}$, the TB log-residual is
\begin{equation}
\label{eq:tb_llm}
\begin{aligned}
\Delta^{\mathrm{TB}}(\xi)
&=
\log Z_\theta
+
\sum_{i=0}^{\tau-1}\log P_F^\theta(s_{i+1}\mid s_{0:i})
\\
&\quad+
\log P_F^\theta(\top\mid s_{0:\tau})
-
\log R(s_{0:\tau}^{\top}),
\end{aligned}
\end{equation}
and TB minimizes
\[
\mathcal{L}_{\mathrm{TB}}
=
\mathbb{E}_{\xi\sim P_F^\theta}
\big[\Delta^{\mathrm{TB}}(\xi)^2\big].
\]

TB provides a clean global anchor for reward-proportional sampling, but terminal-only rewards yield high-variance credit assignment on long horizons. In practice, a few high-reward trajectories dominate updates; since many trajectories share early prefixes, these prefixes are repeatedly reinforced while alternatives are under-trained, leading to self-reinforcing prefix collapse and reduced diversity~\citep{pmlr-v202-shen23a,madan2023learning}.

\subsection{Subtrajectory Balance: Dense but Over-Constrained Supervision}

Subtrajectory Balance~\citep{madan2023learning} reduces variance by enforcing TB-style consistency on subtrajectories. Following ~\citet{hu2023gflownet}, for any subtrajectory indexed by $i,j$, $0\le i<j\le \tau$, define $\Delta^{\mathrm{SubTB}}_{i\to j}(\xi)$ as
\begin{equation}
\label{eq:msubtb_residual}
\begin{aligned}
\sum_{k=i}^{j-1}\log P_F^\theta(s_{k+1}\mid s_{0:k}) + \log R(s_{0:i}^{\top}) - \log R(s_{0:j}^{\top})\\
+\log P_F^\theta(\top\mid s_{0:j})
-\log P_F^\theta(\top\mid s_{0:i}),
\end{aligned}
\end{equation}
The overall objective is the combination of all subtrajectory objectives, optionally weighted by $w_l$ based on the length of the subtrajectory:
\begin{equation}
\label{eq:msubtb_loss}
\mathcal{L}_{\mathrm{SubTB}}(\xi)
\triangleq
\frac{\sum_{\ell=1}^{\tau}\,w_\ell\sum_{i=0}^{\tau-\ell}\big(\Delta^{\mathrm{SubTB}}_{i\to i+\ell}(\xi)\big)^2}
{\sum_{\ell=1}^{\tau}\,w_\ell\sum_{i=0}^{\tau-\ell} 1},
\end{equation}

While SubTB provides dense supervision by enforcing consistency on many subtrajectories, in terminable prefix trees it introduces a large number of overlapping constraints that share the same termination head.
Each subtrajectory implicitly treats its endpoint as a pseudo-terminal, imposing a distinct boundary condition involving the termination probability $q_\theta(\top\mid s_{0:i})$.
These heterogeneous boundary conditions are difficult to satisfy simultaneously, and gradients from many windows accumulate on the shared termination logits.
As a result, the model can reduce SubTB residuals by adjusting termination probabilities rather than improving token-level transitions, leading to biased termination behavior such as systematic length drift.
This over-constraining effect motivates our more conservative approach to densifying prefix-level supervision.

\subsection{RapTB: Rooted Absorbed Prefix Trajectory Balance}

RapTB addresses the trade-off between TB's high variance and SubTB's over-constrained objective. It restricts dense supervision to rooted-prefix residuals, reducing destabilizing boundary conditions, and utilizes partial credit as additional training target by ``absorbing'' credits from suffixes.

\paragraph{Rooted Prefix Residuals.}
Instead of constraining all subtrajectories as in SubTB, we only focus on subtrajectories that are ``rooted,'' originating from $s_0$ (Appendix~\ref{app:deriv_rooted}). The rooted residual at step $k$ is defined as the difference between the TB residual of the current prefix and the root:
\begin{equation}
\label{eq:rooted}
\bar\Delta_k(\xi)
\triangleq
\Delta_k^{\mathrm{TB}}(\xi)
-
\Delta_0^{\mathrm{TB}}(\xi).
\end{equation}
By eliminating the global constant $\log Z_\theta$, this formulation creates a local consistency signal anchored to $s_0$. Unlike SubTB, which creates conflicting boundary conditions via overlapping windows, our approach provides incremental, step-by-step supervision.

\paragraph{Absorbed Suffix Rewards.}
To stabilize training against stochastic variance, we introduce the absorbed suffix reward. This approach constructs a lower-variance target by backing up the rewards from the observed suffix $s_{k:\tau}$, employing an aggregation mechanism (See details in Appendix~\ref{app:backup_details}). This distills hindsight information into a smoothed signal, guiding the policy more reliably than terminal feedback (See analysis in Appendix~\ref{app:vr_view}). For a trajectory $s_{0:\tau}$, let $u_j$ denote the task-only reward at position $j$, i.e., $\lambda S(s_{0:j}^{\top})$, and we define the absorbed target at $k$ as:

\begin{align}
\label{eq:backup_defs}
u_k^{\max}
&\triangleq \max_{j\in[k,\tau]} u_j,\\
u_k^{soft}
&\triangleq
\frac{1}{\beta}\log \sum_{j=k}^{\tau}\exp\!\Big(\beta u_j - \beta\rho\,(j-k)\Big),\\
u_k^{tgt}
&\triangleq
\alpha\,u_k^{\max} + (1-\alpha)\,u_k^{soft},\qquad \alpha\in[0,1].
\end{align}

The $u_k^{\max}$ provides a lower bound on prefix credit: the credit (or size of the flow in GFlowNet) in the prefix should be no smaller than the one in its continuation.
The $u_k^{soft}$ term smoothly aggregates multiple suffix rewards in log space, where $\beta>0$ controls the smoothness.
The distance penalty $\rho(j-k)$ downweights distant evidence, and $\rho\ge 0$ controls the penalty strength.

The absorbed target $u_k^{tgt}$ is the task-only partial credit assigned to the prefix $s_{0:k}$. We can then treat it as an ``estimated reward'' for $s_{0:k}$ and train the model to match it. \looseness=-1
Operationally, this is equivalent to recomputing the rooted TB residual using a surrogate stop-reward in which the task-only component $u_k$ is replaced by $u_k^{tgt}$ (details in Appendix~\ref{app:deriv}):

\begin{equation}
\label{eq:aux}
\mathcal{L}_{\mathrm{aux}}(\xi)
\triangleq
\frac{\sum_{k=1}^{\tau} w_k\,\big(\bar\Delta_k(\xi) + u_k-u_k^{tgt})^2}
{\sum_{k=1}^{\tau} w_k},
\end{equation}
where $w_k$ is the length weight.

\paragraph{Final Objective.}
The RapTB objective integrates global TB consistency with this dense guidance. 
\begin{equation}
\label{eq:raptb_final}
\mathcal{L}_{\mathrm{RapTB}}
=
\mathbb{E}_{\xi\sim P_F^\theta}
\Big[
\underbrace{\Delta^{\mathrm{TB}}(\xi)^2}_{\text{Anchor}}
+
\underbrace{\eta\,\mathcal{L}_{\mathrm{aux}}(\xi)}_{\text{Partial Credit}}
\Big],
\end{equation}
where $\eta$ balances global consistency with auxiliary term. The TB term remains the only exact balance condition whose optimum matches the reward-proportional target. The auxiliary term is a variance-reducing regularizer that can improves optimization.

\paragraph{Fixed-point property.}
Without absorption ($u_k^{tgt}{=}u_k$), the rooted residual $\bar\Delta_k$ vanishes whenever all prefix TB residuals are zero, so RapTB shares TB's global optimum exactly.
With absorption, $u_k^{tgt}\ge u_k$ (the suffix set includes $j{=}k$), so the absorbed residual is generally nonzero at the TB optimum.
Let $\theta^*$ satisfy $\mathcal{L}_{\mathrm{TB}}(\theta^*){=}0$ with finite auxiliary cost $C^*{=}\eta\mathcal{L}_{\mathrm{aux}}(\theta^*)$.
For any global minimizer $\hat\theta$ of $\mathcal{L}_{\mathrm{RapTB}}$:
$\mathcal{L}_{\mathrm{TB}}(\hat\theta)\le\mathcal{L}_{\mathrm{RapTB}}(\hat\theta)\le C^*$,
so the TB deviation is bounded by $\eta\mathcal{L}_{\mathrm{aux}}(\theta^*)$ and vanishes as $\eta\!\to\!0$.
The auxiliary term thus acts as a variance-reducing regularizer that does not destroy the global TB anchor (Appendix~\ref{app:vr_view}).

\paragraph{Compare three losses.}
TB uses a global objective with sparse supervision, while SubTB increases supervision density but over-constrains hinder optimization in prefix trees.
RapTB preserves TB as the sole exact balance constraint and adds (i) rooted prefix supervision that avoids heterogeneous window boundaries and (ii) absorbed suffix rewards that reduce variance and improve credit assignment.

\subsection{Submodular Replay: Diversity- and Length-balanced Experience Selection}
\label{sec:submodreplay}

RapTB addresses within-trajectory credit assignment.
In parallel, to explicitly enforce diversity in addition to the exploration of GFlowNet,
we maintain a fixed-size replay buffer of size $B$ and update it by selecting a representative, diverse, and length-balanced subset from the union of the current buffer and a newly collected batch (details in Appendix~\ref{app:subm_details}).

\paragraph{Submodular selection.}
At each buffer update step, we form the ground set $\mathcal{G}$ as the union of the current buffer and a new generated batch, then select $S\subseteq\mathcal{G}$ with $|S|=B$ by maximizing a monotone submodular function subject to a cardinality constraint~\citep{subM_bg3, subM_bg2, subM_bg1}.

Let $\mathrm{sim}(v,x)\in[0,1]$ be a task-appropriate similarity.
In experiments, we use Morgan fingerprints with Tanimoto similarity for SMILES, and $n$-gram shingle Jaccard similarity for text generation tasks.
We define the facility-location coverage operator, which reflects how well the buffer represents a sample $v$. \looseness=-1
\begin{equation}
\label{eq:msim_cov}
\mathrm{msim}(v,S)\triangleq \max_{x\in S}\mathrm{sim}(v,x),
\qquad \mathrm{msim}(v,\emptyset)\triangleq 0.
\end{equation}
The overall submodular objective $f(S)$ is
\begin{equation}
\label{eq:setfunc}
\begin{aligned}
\underbrace{\sum_{x\in S}\mathrm{static}(x)}_{\text{quality / feasibility}}
\;+\;
\underbrace{\lambda_{\mathrm{div}} \sum_{v\in \mathcal{G}} \mathrm{msim}(v,S)}_{\text{facility-location coverage}}
\;+\;
\underbrace{\vphantom{\sum_{u\in \mathcal{G}}}\lambda_{\mathrm{len}}\, f_{\mathrm{len}}(S)}_{\text{length coverage}}.
\end{aligned}
\end{equation}
When weights are nonnegative, each term is monotone submodular and so is their sum. $\mathrm{static}(x)$ denotes a fixed per-sample quality/feasibility term.
Validity gating restricts the candidate pool used in the facility-location term, without changing the form of the objective.
For length coverage, we discretize samples into bins and use concave-over-counts histogram coverage:
\begin{equation}
\label{eq:lencover}
f_{\mathrm{len}}(S) \triangleq \sum_{b=1}^{B_{\mathrm{bin}}} \alpha_b \, g\big(c_b(S)\big),
\quad
g(c)\triangleq \log(1+c),
\end{equation}
where $c_b(S)$ is the count in bin $b$ and $\alpha_b\!\ge\!0$ can bias coverage toward desired lengths; implementation details are in Appendix~\ref{app:subm_details}.

\paragraph{Greedy update and efficiency.}
For every gradient step, we update the fixed-size buffer by optimizing $\max_{S\subseteq \mathcal{G}, |S| \leq B} f(S)$ using greedy algorithm.
With cached similarities and histogram counts, one update costs $O(B|\mathcal{G}|)$. In our settings $B$ and $|\mathcal{G}|$ are small, so the overhead is negligible ($\sim 10\mathrm{ms}$ extra time cost per update).

\section{Experiments}
\label{sec:experiments}

\subsection{Tasks}
\label{sec:tasks}

\paragraph{Scaffold-conditioned SMILES optimization.}
We study conditional molecular generation \citep{li2024geometric, pepflow, Xu_2025, huang2025cascade, fei2026agentic, huang2025skillpuzzler} where the conditioning input is a fixed molecular scaffold and the model generates a completion by adding fragments.
Each terminal sequence is a SMILES string $x \in \mathcal{X}$ that must be chemically valid.
We optimize a property objective based on the Estimate of Drug-likeness ~\citep{qed, Xu_2024}.
Training aims to learn a GFlowNet sampler whose induced terminal distribution assigns higher probability to high-reward scaffold-consistent molecules while maintaining diversity among valid completions. \looseness=-1

\paragraph{Expr24 arithmetic expression generation.}
We also evaluate on a discrete, fully verifiable sparse-reward task: generating an arithmetic expression whose value equals $24$.
A terminal $x \in \mathcal{X}$ is a variable-length token sequence consisting of digits and operators $\{+,-,\times,\div\}$, evaluated with standard operator precedence.
The task score is sparse and exact:
\[
R(x)=\mathbb{I}[\mathrm{eval}(x)=24].
\]
This task isolates exploration, credit assignment, termination/length bias, and collapse behavior without domain-specific feasibility issues (e.g., chemical validity).

\paragraph{CommonGen: Concept-to-Sentence Generation.}
To ascertain whether the termination drift identified in the \textbf{synthetic} Expr24 task generalizes to \textbf{realistic} scenarios, we employ a diagnostic subset of CommonGen~\citep{lin2020commongen}. Unlike the synthetic setting, this task introduces strong pre-trained linguistic priors. We treat the reference model as a anchor, allowing us to strictly isolate objective-induced termination drift---deviations from natural stopping priors driven solely by reward maximization.

\subsection{Compared objectives and replay strategies}
\label{sec:baselines}

We compare three objectives adapted to terminable LLM-GFlowNets: TB , SubTB, and RapTB. Unless otherwise specified, methods employ the standard reward-prioritized replay (RP) from~\citet{hu2024amortizing}; for ablation purposes, we also include Reward-prioritized replay training (PRT)~\citep{pmlr-v202-shen23a}. We also introduce our proposed submodular replay (SubM), which acts to explicitly promote diversity among stored high-reward trajectories.\looseness=-1

\paragraph{Implementation summary.}
We fine-tune Llama-3.2-1B with LoRA (rank $16$) using AdamW (lr $10^{-4}$). All methods share the same model architecture, tokenizer, decoding constraints, and optimizer configuration.
Replay-buffer sizes, SubM refresh settings, and decoding constraints are task-specific and reported in Appendix~\ref{app:repro}.

\subsection{Metrics}
\label{sec:metrics}

We evaluate (i) feasibility and quality (Acc/Score/BLEU~\citep{papineni2002bleu}), (ii) diversity (Entropy for text; FPDiv for SMILES), (iii) length/termination calibration, and (iv) prefix-collapse diagnostics (Surv/PefEnt/Top1 versus depth). For Expr24, we additionally report coverage (Unique$_\checkmark$, NormCov) and distributional fidelity (KL/JS) against the full set of exact solutions. Unless stated otherwise, we report means over multiple seeds with 95\% confidence intervals; all metric definitions and computation details are provided in Appendix~\ref{app:metrics}.

\subsection{Results on Scaffold-conditioned SMILES Optimization}
\label{sec:results_smiles}



\subsubsection{Overall performance}
\label{sec:results_smiles_overall}

Table~\ref{tab:smiles_main} reports all compared methods on scaffold-conditioned SMILES generation.
RL baselines (PPO, GRPO) achieve high validity but collapse to near-zero diversity: PPO concentrates on a single mode, and GRPO achieves Entropy $\le 0.98$.
This confirms the fundamental gap between reward maximization, which concentrates mass on the single best mode, and reward-proportional sampling, which spreads mass across all high-reward modes~\citep{hu2024amortizing}.

Among GFlowNet objectives, SubTB suffers from severe validity degradation (Acc $0.328$), consistent with the termination drift analyzed in Section~\ref{sec:why_subtb_abnormal}.
TB achieves near-perfect validity but concentrates on short sequences (Len $3.065$), yielding weaker reward quality and diversity.
RapTB+SubM achieves the best quality--diversity trade-off while maintaining high validity.
As detailed in Appendix~\ref{app:results_smiles}, applying SubM to TB also yields substantial improvements by broadening replay coverage, underscoring SubM's role as a generic component for GFlowNets.
We additionally compare against AvgPrefixTB, an alternative prefix-level GFlowNet objective, in Appendix~\ref{app:avgprefixtb}.

\begin{table}[!htbp]
\centering
\small
\setlength{\tabcolsep}{4pt}
\renewcommand{\arraystretch}{1.08}
\caption{
\textbf{SMILES generation performance.}
Unless specified, metrics are computed on valid samples. \texttt{Len} represents average token length.
Results are averaged over seeds; see Appendix~\ref{app:results_smiles} for 95\% CIs and per-length breakdowns.
}
\label{tab:smiles_main}
\vspace{-0.35em}
\begin{tabular}{@{}lccccc@{}}
\toprule
Method & Acc $\uparrow$ & Score $\uparrow$ & Entropy $\uparrow$ & FPDiv $\uparrow$ & Len \\
\midrule
PPO  & 1.000 & 0.604 & $\approx$0 & -- & -- \\
GRPO & 0.997 & 0.661 & 0.98       & -- & 10.0 \\
\midrule
TB             & \textbf{0.998} & 0.717 & 2.503 & 0.807 & 3.065 \\
SubTB          & 0.328 & 0.755 & 2.127 & 0.836 & 8.354 \\
RapTB          & 0.996 & 0.740 & 2.448 & 0.860 & 6.142 \\
RapTB + SubM   & 0.988 & \textbf{0.844} & \textbf{2.726} & \textbf{0.898} & 7.435 \\
\bottomrule
\end{tabular}
\end{table}

\subsubsection{Per-length analysis}
\label{sec:results_smiles_length}

\begin{figure}[t]
\centering
\begin{subfigure}[t]{\linewidth}
  \centering
  \includegraphics[width=0.92\linewidth]{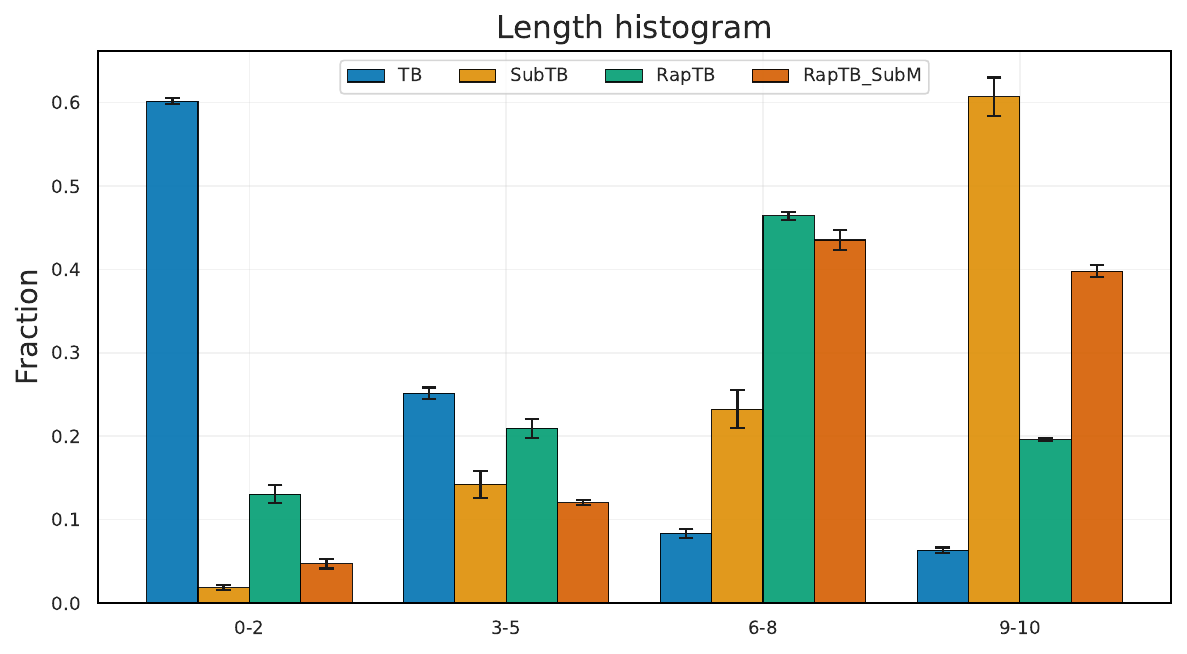}
  \caption{Valid-only length histogram.}
  \label{fig:smiles_len_hist_L10}
\end{subfigure}

\vspace{0.55em}

\begin{subfigure}[t]{\linewidth}
  \centering
  \includegraphics[width=0.92\linewidth]{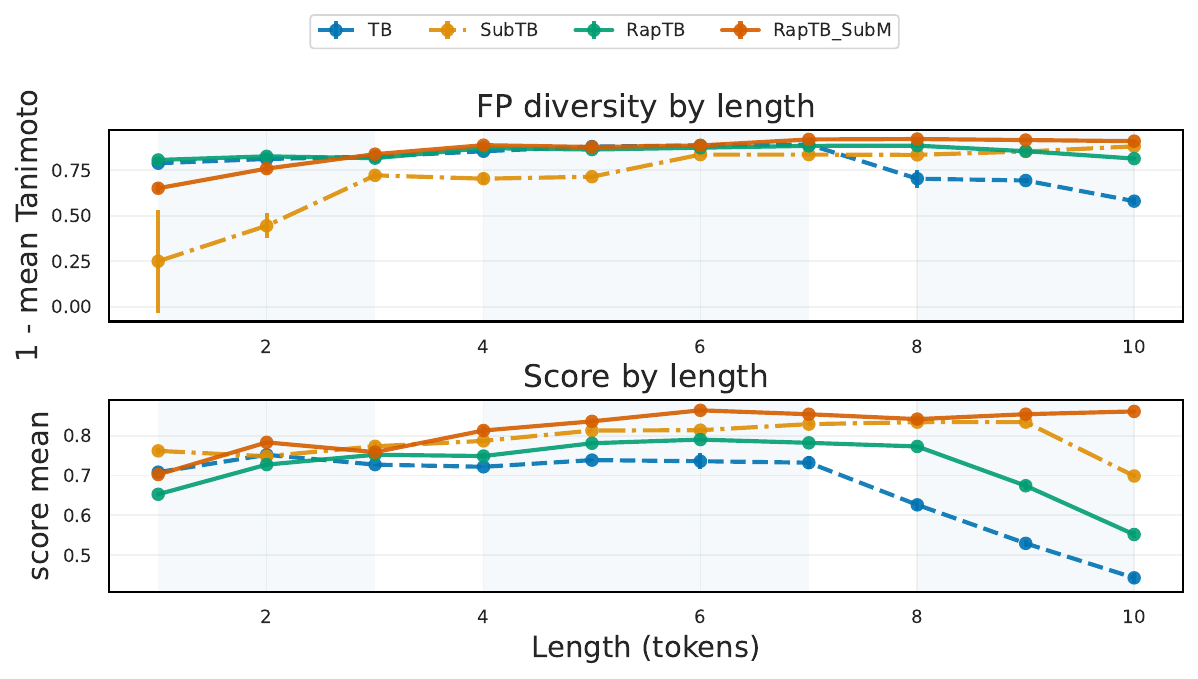}
  \caption{Valid-only score and FPDiv versus length.}
  \label{fig:smiles_perf_by_len_L10}
\end{subfigure}

\vspace{0.35em}
\caption{\textbf{Length-stratified analysis on SMILES ($L_{\max}=10$).}
(a) Distribution of valid generation lengths.
(b) Mean Score and FPDiv conditioned on length.
}
\label{fig:smiles_length_breakdown_L10}
\vspace{-0.35em}
\end{figure}

Aggregate metrics can be confounded by length shifts (e.g., diversity is more likely to be larger for longer sequences).
Figure~\ref{fig:smiles_length_breakdown_L10} reports the valid-only length histogram and length-conditioned score/diversity.
RapTB+SubM remains strong across most lengths, whereas TB concentrates on short lengths and degrades in the long-length regime. \looseness=-1

\subsubsection{Prefix collapse analysis}
\label{sec:results_smiles_L10_prefix}

\begin{figure*}[!htbp]
\centering
\includegraphics[width=0.9\textwidth]{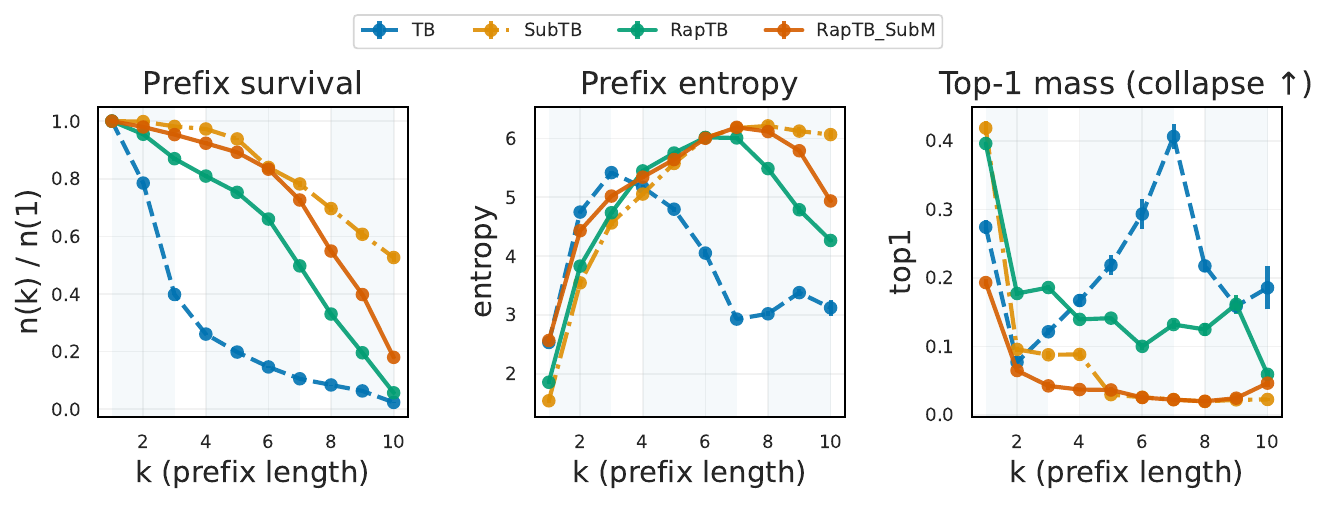}
\caption{
\textbf{Prefix-collapse diagnostics on SMILES ($L_{\max}=10$).}
Metrics vs.\ prefix length $k$ computed on correct samples: prefix survival (fraction of samples reaching length $k$), prefix entropy (diversity of prefix), and top-1 mass (frequency of the most common prefix).
}
\label{fig:smiles_prefix_L10}
\vspace{-0.35em}
\end{figure*}

Diverse terminals can still share highly concentrated early prefixes and only branch late, a failure mode we refer to as \emph{prefix collapse}.
We therefore compute position-wise prefix statistics on all valid samples.
Figure~\ref{fig:smiles_prefix_L10} reports prefix diagnostics by prefix length $k$.
TB displays rapid survival decay (failure to sustain generation) and sharply increasing top-1 mass at longer prefixes (dominance of a single prefix) , indicating severe concentration on a few shared partial trajectories.
RapTB sustains higher prefix entropy and lower top-1 mass deeper into the trajectory, consistent with earlier and broader branching among correct samples.

\subsubsection{Longer-horizon stress test }
\label{sec:results_smiles_L15}
Increasing $L_{\max}$ to $15$ exposes severe length collapse in TB, which concentrates mass on short trajectories and fails to reach the long-horizon regime (Table~\ref{tab:smiles_L15_stress}).
While SubTB improves coverage at the cost of validity, RapTB effectively mitigates this length bias, unlocking access to extended trajectories without compromising accuracy.
Crucially, RapTB+SubM achieves the most robust performance: it maximizes long-horizon coverage (Frac(11+)), yields the best quality-diversity trade-off (Score \& MacroFPDiv), and exhibits the lowest prefix concentration (Top1), demonstrating resistance to mode collapse. \looseness=-1

\begin{table*}[!htbp]
\centering
\small
\setlength{\tabcolsep}{4.5pt} 
\renewcommand{\arraystretch}{1.15} 
\caption{\textbf{Longer-horizon stress test on SMILES ($L_{\max}=15$).}
We report length fractions of valid samples (Length Dist.) and diversity metrics.
Prefix diagnostics are averaged over $k$: Surv (Prefix survival), PefEnt (prefix entropy), and Top1 (Top-1 mass).
MacroFP macro-averages FPDiv across length bins 0--5, 6--10, and 11+.
RapTB+SubM achieves the best balance of long-horizon coverage and diversity. CIs omitted for space (Appendix~\ref{app:results_smiles}).
}
\label{tab:smiles_L15_stress}
\vspace{-0.5em}
\begin{tabular}{@{}lccccccccccc@{}}
\toprule
& \multicolumn{2}{c}{\textbf{Performance}} & \multicolumn{3}{c}{\textbf{Length Dist.}} & \multicolumn{3}{c}{\textbf{Prefix Diagnostics}} & \multicolumn{2}{c}{\textbf{Diversity}} \\
\cmidrule(lr){2-3} \cmidrule(lr){4-6} \cmidrule(lr){7-9} \cmidrule(lr){10-11}
Method & Acc & Score & 0--5$\downarrow$ & 6--10 & 11+$\uparrow$ & Surv$\uparrow$ & PefEnt$\uparrow$ & Top1$\downarrow$ & MacroFP$\uparrow$ & FPDiv$\uparrow$ \\
\midrule
TB & \textbf{0.999} & 0.716 & 0.858 & 0.129 & 0.013 & 0.207 & 2.99 & 0.303 & 0.653 & 0.813 \\
SubTB & 0.636 & 0.742 & 0.286 & 0.442 & 0.271 & 0.561 & 4.82 & 0.085 & 0.716 & 0.770 \\
RapTB & 0.988 & 0.768 & 0.113 & 0.318 & 0.568 & 0.681 & \textbf{5.59} & 0.084 & 0.793 & 0.810 \\
RapTB+SubM & 0.972 & \textbf{0.849} & \textbf{0.094} & 0.205 & \textbf{0.701} & \textbf{0.751} & 5.32 & \textbf{0.071} & \textbf{0.805} & \textbf{0.868} \\
\bottomrule
\end{tabular}
\vspace{-0.5em}
\end{table*}

\subsection{Variable-length Expr24 under sparse rewards}
\label{sec:expr24}

\paragraph{Enumerable solutions enable controlled diagnostics.}
\texttt{Expr24} disentangles two factors affecting performance: external coverage (the capacity to discover modes) and internal credit assignment (the fidelity of probability mass allocation).
Because the full correct set $\mathcal{Y}^*$ is enumerable, we can (i) control coverage via an oracle replay buffer sampling from it,
and (ii) directly compare the learned terminal distribution $\pi$ to a reference $p^*$ using bidirectional KL and token-wise JS,
where KL($p^*\!\to\!\pi$) highlights mode dropping and KL($\pi\!\to\!p^*$) reflects over-concentration.

\paragraph{Expr24 Results.}
Table~\ref{tab:expr24_main} shows that RapTB achieves a superior trade-off between correctness and coverage.
Under standard RP, TB suffers from severe mode collapse (Unique$_\checkmark\!\approx\!5$), whereas RapTB significantly improves diversity without compromising accuracy.
This advantage is amplified by SubM: RapTB+SubM doubles the normalized coverage of the strongest baseline ($0.209$ vs.\ $0.100$) while maintaining near-perfect accuracy ($>$0.99).
Finally, the Oracle setting verifies the objective's effectiveness: RapTB outperforms TB in both accuracy ($0.945$ vs.\ $0.919$) and distribution matching (lower KL/JS), indicating that RapTB learns a more precise distribution.

\paragraph{Complementarity of RapTB and SubM.}
Table~\ref{tab:expr24_main} reveals complementary roles.
In coverage-limited regimes, SubM alone can outperform RapTB alone: TB+SubM achieves NormCov $0.100$ while RapTB+RP reaches only $0.039$.
Once coverage is sufficient, RapTB's credit-assignment benefit becomes dominant: with SubM, RapTB doubles NormCov to $0.209$.
Under Oracle replay, RapTB still improves Acc ($0.945$ vs.\ $0.919$) and JS ($0.013$ vs.\ $0.016$), confirming its benefit in distributional fidelity.

\paragraph{Additional tasks and scaling.}
We further evaluate on AMP biological sequence generation~\citep{jain2022biological} (amino-acid vocabulary, non-differentiable reward) and scale to 3B, 8B (Llama-3.2), and 32B (Qwen3) on SMILES.
On AMP, RapTB+SubM achieves the best performance--diversity--novelty trade-off within 3K steps, while SubTB collapses to maximum length ($49.3$), inflating its diversity (Appendix~\ref{app:amp}).
Across model scales, SubTB's termination drift persists at every size (Acc: $0.311$/$0.391$/$0.795$ at 3B/8B/32B), confirming the failure is structural and architecture-independent.
RapTB+SubM consistently achieves the best quality--diversity trade-off at all scales (Appendix~\ref{app:scaling}).

\begin{table*}[t]
\centering
\small
\setlength{\tabcolsep}{4.5pt}
\renewcommand{\arraystretch}{1.1}
\caption{\textbf{Expr24 results under different replay schemes.} Results are averaged over different random seeds; 95\% confidence intervals are reported in Appendix~\ref{app:results_expr24}.}
\label{tab:expr24_main}
\begin{tabular}{llcccccc}
\toprule
& & \multicolumn{2}{c}{\textbf{Diversity}} & \textbf{Quality} & \multicolumn{3}{c}{\textbf{Distributional Fidelity}} \\
\cmidrule(lr){3-4} \cmidrule(lr){5-5} \cmidrule(lr){6-8}
\textbf{Replay} & \textbf{Objective} & Unique$_\checkmark$$\uparrow$ & NormCov$\uparrow$ & Acc$\uparrow$ &
KL($\pi\!\to\!p^*$)$\downarrow$ & KL($p^*\!\to\!\pi$)$\downarrow$ & JS$_{\text{tok}}$$\downarrow$ \\
\midrule
\multirow{2}{*}{\centering N/A}
& PPO   & 1     & 0.000 & 0.003 & -- & -- & -- \\
& GRPO  & 1     & 0.000 & 0.002 & -- & -- & -- \\
\midrule
\multirow{3}{*}{\centering PRT}
& TB    & 103.7 & 0.016 & 0.999 & 1.105 & 7.803 & 0.292 \\
& SubTB & 292.0 & 0.046 & 0.311 & 0.424 & 0.672 & 0.107 \\
& RapTB & 129.3 & 0.020 & 0.992 & 0.908 & 5.538 & 0.230 \\
\midrule
\multirow{3}{*}{\centering RP}
& TB    & 5.3   & 0.001 & 1.000 & 1.297 & 11.403 & 0.339 \\
& SubTB & 324.7 & 0.051 & 0.229 & 0.455 & 0.865  & 0.109 \\
& RapTB & 246.7 & 0.039 & 0.991 & 0.561 & 4.480  & 0.147 \\
\midrule
\multirow{3}{*}{\centering SubM}
& TB    & 642.0  & 0.100 & 0.996 & 0.182 & 0.441 & 0.049 \\
& SubTB & 331.3  & 0.052 & 0.061 & 0.149 & 0.286 & 0.040 \\
& RapTB & 1337.3 & 0.209 & 0.994 & 0.169 & 0.623 & 0.048 \\
\midrule
\multirow{3}{*}{\centering Oracle}
& TB    & 5198.0 & 0.812 & 0.919 & 0.062 & 0.066 & 0.016 \\
& SubTB & 35.7   & 0.006 & 0.006 & 0.266 & 1.491 & 0.071 \\
& RapTB & 5220.7 & 0.816 & 0.945 & 0.052 & 0.056 & 0.013 \\
\bottomrule
\end{tabular}
\end{table*}

\paragraph{Diagnosis of SubTB's abnormal behaviors.}
\label{sec:why_subtb_abnormal}
In the variable-length Expr24 setting, we observe pronounced termination drift: the log-probability of sampled termination, $\log p_{\text{term}}(\tau)$, degrades to extremely negative values (Table~\ref{tab:expr24_json_metrics}). This severely impacts the hit rate when the stopping condition is a decision variable. We attribute this phenomenon to the enforcement of numerous arbitrary-start windows, which can be partially satisfied by a global shift in $\log q_\theta(\top\mid s_{0:\tau})$ (see analysis in Appendix~\ref{app:deriv_naive_subtb}). To investigate whether termination drift is the dominant failure mode, we employ \textsc{RootSubTBLogZ}, which restricts SubTB windows to be rooted and reintroduces a learnable global normalizer $Z_\theta$. As shown in Table~\ref{tab:expr24_json_metrics}, this modification mitigates termination drift and restores accuracy to nearly 100\%.

\begin{table}[t]
\centering
\small
\renewcommand{\arraystretch}{1.05}
\caption{\textbf{Termination/length calibration diagnostic on Expr24.}
More negative values indicate overly suppressed termination.
$\log p_{\text{term}}(\tau)$ is the termination log-probability at the sampled stop step, computed from the model's raw $q_\theta(\top\mid s_{0:\tau})$.}
\label{tab:expr24_json_metrics}
\colorlet{rowgray}{black!10}
\begin{tabular}{lcccc}
\toprule
Method & Acc$\uparrow$ & NormCov$\uparrow$ & $\log p_{\text{term}}(\tau)$ & $\log Z$ \\
\midrule
\multicolumn{5}{l}{\textbf{RP Replay}}\\
TB & 0.999 & 0.001 & -0.000  & 0.063 \\
\rowcolor{rowgray} SubTB & 0.229 & 0.051 & -79.638   & -- \\
RapTB & 0.991 & 0.039 & -0.065  & 0.062 \\
\rowcolor{rowgray} RootSubTBLogZ & 0.999 & 0.023 & -0.068  & 0.062 \\
\midrule
\multicolumn{5}{l}{\textbf{Oracle Replay}}\\
TB & 0.922 & 0.813 & -0.436  & 0.037 \\
\rowcolor{rowgray} SubTB & 0.006 & 0.006 & -86.415   & -- \\
RapTB & 0.945 & 0.816 & -0.644  & 0.038 \\
\rowcolor{rowgray} RootSubTBLogZ & 0.885 & 0.727 & -1.432   & 0.036 \\
\bottomrule
\end{tabular}
\end{table}

\subsection{Results on CommonGen}
\label{sec:results_commongen}

Table~\ref{tab:commongen_results} confirms that optimization pressure can indeed override linguistic priors. Despite the model's natural tendency to stop, SubTB deviates catastrophically from the anchor, saturating length ($20.00$) by suppressing stopping logits ($\Delta \log p_{\text{term}} \approx -28.32$). Conversely, RapTB maintains calibration ($\approx -0.94$). Notably, RapTB+SubM achieves superior BLEU (33.23) at natural lengths ($11.83$), demonstrating robust performance.

\begin{table}[!htbp]
\centering
\small
\setlength{\tabcolsep}{4pt}
\renewcommand{\arraystretch}{1.1}
\caption{\textbf{CommonGen performance.} $\Delta \log p_{\text{term}}$ measures the deviation of the learned policy's termination logits from the pre-trained reference anchor.}
\label{tab:commongen_results}
\vspace{-0.5em}
\begin{tabular}{@{}lcccc@{}}
\toprule
Method & Entropy$\uparrow$ & BLEU-4$\uparrow$  & Len  & $\Delta \log p_{\text{term}}$ \\
\midrule
TB & 2.966 & 5.95 & 13.86 & -1.34 \\
SubTB & 3.719 & 24.39 & 20.00 & -28.32  \\
RapTB & 3.933 & 11.75 & 15.63 & -0.94 \\
RapTB+SubM & \textbf{4.102} & \textbf{33.23} & 11.83 & 4.89  \\
\bottomrule
\end{tabular}
\end{table}

\subsection{Ablations.}
Table~\ref{tab:ablation} ablates the key design choices of RapTB and SubM on SMILES.
Removing reward absorption degrades both score and diversity, suggesting that suffix evidence provides useful prefix credit.
Using only max or only soft backups increases score but reduces diversity, while the mixed backup improves the balance.
Detaching termination gradients in the auxiliary branch is also important. Without it, the model collapses to very short sequences (Len 3.40) and the score drops.
SubM components are complementary. Reward-only improves score, diversity-only improves FPDiv, and length-only increases long-horizon coverage.
Combining them yields the strongest overall trade-off.

Additional sensitivity analysis over $(\beta,\rho,\eta,k_{\min})$ across both tasks is provided in Appendix~\ref{app:hparam_sensitivity}; across all 18 configurations tested, no catastrophic failure, length collapse, or termination drift is observed.

\begin{table}[t]
\centering
\small
\setlength{\tabcolsep}{4pt}
\caption{\textbf{Ablation study on SMILES generation.}}
\label{tab:ablation}
\begin{tabular}{@{}lcccc@{}}
\toprule
Variant & Score $\uparrow$ & FPDiv $\uparrow$ & Entropy $\uparrow$ & Len \\
\midrule
RapTB                               & 0.740 & 0.860 & 2.448 & 6.142 \\
\midrule
RapTB w/o reward absorb              & 0.716 & 0.805 & 2.031 & 5.296 \\
\midrule
RapTB absorb max-only                & 0.821 & 0.775 & 1.716 & 7.431 \\
RapTB absorb soft-only                & 0.819 & 0.748 & 1.516 & 7.710 \\
\midrule
RapTB + SubM (Full)                  & 0.844 & 0.898 & \textbf{2.726} & 7.435 \\
\quad + length-only SubM             & 0.773 & 0.885 & 1.914 & 6.569 \\
\quad + diversity-only SubM          & 0.741 & \textbf{0.942} & 2.602 & 5.459 \\
\quad + reward-only SubM             & \textbf{0.878} & 0.884 & 1.876 & 8.122 \\
\midrule
RapTB w/o detach $p_{\mathrm{term}}$ & 0.714 & 0.892 & 2.072 & 3.403 \\
\bottomrule
\end{tabular}
\vspace{-0.3em}
\end{table}

\section{Discussion and Practical Guidance}

\paragraph{Mitigating Replay-Induced Collapse.}
Standard reward-prioritized replay often induces ``rich-get-richer'' dynamics~\citep{pmlr-v202-shen23a}, where the training distribution collapses onto a narrow set of repeated high-reward modes. SubM explicitly counters this by enforcing structural diversity within the buffer. This prevents near-duplicate dominance and ensures the policy learns from a broad, representative landscape rather than degenerate subsets.

\paragraph{Credit Assignment and Consistency.}
A key limitation of SubTB is that enforcing constraints on arbitrary windows creates conflicting boundary conditions, effectively hardening optimization and destabilizing termination. RapTB resolves this by grounding dense supervision to rooted prefixes, ensuring all partial trajectory updates remain consistent with the global partition function $Z$. A future direction is adaptive subtrajectory selection, where the model learns to identify and prioritize essential substructures online. \looseness=-1

\section{Conclusion}
We studied mode collapse in terminable LLM-GFlowNets and identified two coupled and reproducible failure modes: prefix collapse and length bias.
To address unstable credit assignment without inducing termination drift, we proposed RapTB, which augments terminal Trajectory Balance with rooted prefix constraints and suffix-absorbed reward shaping.
To address replay-induced distribution shift and limited external coverage, we introduced SubM, a submodular replay refresh strategy that balances reward with diversity and length support.
Across tasks, RapTB+SubM improves long-horizon stability and coverage, yielding better reward--diversity trade-offs and substantially reduced prefix collapse.
Scaling experiments up to 32B parameters across two architecture families and an additional biological sequence task confirm that these improvements are consistent across domains and model scales (Appendices~\ref{app:amp}--\ref{app:scaling}).
We hope these results encourage future objectives that explicitly couple coverage-aware replay with selective, adaptively weighted subtrajectory learning for robust autoregressive GFlowNet training.

\clearpage

\section*{Impact Statement}
This paper advances learning methods for sampling diverse high-quality solutions from various reward distributions.
In molecular generation, improved exploration and property optimization may accelerate candidate discovery, but generated molecules are only hypotheses and require downstream validation.
We do not foresee direct negative societal impacts from the method itself beyond the general risks of misuse of generative models; appropriate safeguards and domain expert oversight remain necessary.


\bibliography{refs}
\bibliographystyle{icml2026}

\newpage
\appendix
\onecolumn
\newcommand{\NA}{\textcolor{gray}{--}}

\captionsetup{font=small,labelfont=bf}

\section{Additional Results}
\label{app:results}

\subsection{SMILES}
\label{app:results_smiles}
Molecule generation is a central task in the application of AI to chemistry, encompassing two main challenges. The first is molecular representation: without an effective representation, the vast chemical space becomes difficult to navigate. In this work, we adopt SMILES as our primary representation. Nevertheless, a variety of alternative approaches have demonstrated strong potential. For instance, quantum chemistry-based and geometric descriptors offer physically grounded encodings of molecular structure~\cite{shi2025unifying}, while chirality-aware representations have proven essential for distinguishing stereoisomers in downstream tasks~\cite{peng2025chiralcat}. Reaction-targeted descriptors, such as substrate-aware features automatically extracted for performance prediction, provide another powerful lens through which to characterize molecular reactivity~\cite{yu2025machine}. More recently, round-trip molecule-text alignment frameworks have shown that bridging modalities can yield richer and more transferable representations~\cite{chen2025rtmol}. The second challenge concerns the choice of generative model. Here, we employ autoregressive modeling for sequential molecule construction. Diffusion-based methods have also emerged as a powerful paradigm in this space, with recent work demonstrating their effectiveness even in data-scarce regimes such as atropisomer generation through multi-task pretraining and classifier guidance~\cite{chen2025atropdiff}.

\subsubsection{Per-length SMILES performance.}
Tables~\ref{tab:bylen_valid_core}--\ref{tab:bylen_valid_div} report valid-only metrics stratified by terminal length $L$,
where the largest differences concentrate in the longest bins ($L\ge 8$) under rewards.
Submodular Replay (SubM) substantially improves length-wise coverage for TB, largely eliminating its long-length degradation:
for $L=8$--$10$, TB's Score increases from $0.626/0.529/0.442$ to $0.861/0.880/0.875$,
with diversity recovering simultaneously (Entropy $\approx 2$ and FPDiv $\approx 0.91$).
Under this stronger coverage regime, the gap between RapTB and TB naturally narrows (a ceiling effect),
yet RapTB+SubM remains competitive across lengths and tends to allocate more mass to the longest bin
(higher Frac/Count at $L\ge 9$) while maintaining high quality, consistent with RapTB improving the learning signal for suffix credit assignment
whereas SubM primarily improves external replay coverage across length/diversity/reward.
In contrast, SubTB (with or without SubM) exhibits pronounced length skew, placing disproportionately large mass at $L=9$--$10$
alongside a substantial validity drop, consistent with the abnormal termination dynamics discussed in the main text.

\paragraph{All-length averaged summary.}
Table~\ref{tab:smiles_all_length_avg} aggregates metrics across all generated lengths ($L_{\max}=10$).
SubM dramatically changes the effective training distribution of TB: the median length shifts from $2.15$ to $7.23$ and
Frac[$0$--$2$] decreases from $0.601$ to $0.143$, while Frac[$9$--$10$] increases from $0.064$ to $0.323$,
yielding a large gain in valid-only average Score ($0.717\!\to\!0.842$) and diversity ($2.503\!\to\!2.775$).
With SubM, RapTB+SubM and TB+SubM achieve very similar averaged Score ($0.844$ vs.\ $0.842$),
but RapTB+SubM allocates significantly more probability mass to the longest bin (Frac[$9$--$10$] $0.402$ vs.\ $0.323$),
at a small but noticeable cost in overall validity (Acc $0.988$ vs.\ $0.996$).
Finally, for SubTB variants, the reported valid-only averages should be interpreted cautiously due to low Acc ($\approx 0.30$):
SubM further concentrates length mass at $L=9$--$10$ (Frac $>0.90$) without resolving the validity collapse.
\label{app:results_smiles_L10_main}

\begin{table*}[!htbp]
\centering
\scriptsize
\setlength{\tabcolsep}{3.2pt}
\renewcommand{\arraystretch}{1.06}
\resizebox{\textwidth}{!}{%
\begin{tabular}{lccccccccccc}
\toprule
Method & Acc & Score & Entropy & FPDiv & Len$_{\mu}$ & Len$_{50}$ & Len$_{90}$ & Frac[0--2] & Frac[3--5] & Frac[6--8] & Frac[9--10] \\
\midrule
TB & 0.998$\pm$0.001 & 0.717$\pm$0.001 & 2.503$\pm$0.026 & 0.807$\pm$0.003 & 3.06$\pm$0.02 & 2.15$\pm$0.03 & 6.42$\pm$0.13 & 0.601$\pm$0.004 & 0.252$\pm$0.007 & 0.084$\pm$0.005 & 0.064$\pm$0.004 \\
TB+SubM & 0.996$\pm$0.000 & 0.842$\pm$0.001 & 2.775$\pm$0.002 & 0.889$\pm$0.002 & 6.56$\pm$0.09 & 7.23$\pm$0.07 & 9.84$\pm$0.06 & 0.143$\pm$0.006 & 0.186$\pm$0.015 & 0.348$\pm$0.013 & 0.323$\pm$0.008 \\
RapTB & 0.996$\pm$0.001 & 0.740$\pm$0.004 & 2.448$\pm$0.017 & 0.860$\pm$0.001 & 6.14$\pm$0.03 & 6.58$\pm$0.03 & 9.01$\pm$0.04 & 0.134$\pm$0.007 & 0.204$\pm$0.009 & 0.466$\pm$0.017 & 0.197$\pm$0.015 \\
RapTB+SubM & 0.988$\pm$0.003 & 0.844$\pm$0.001 & 2.726$\pm$0.017 & 0.898$\pm$0.001 & 7.44$\pm$0.05 & 7.91$\pm$0.03 & 9.84$\pm$0.02 & 0.046$\pm$0.005 & 0.119$\pm$0.003 & 0.433$\pm$0.012 & 0.402$\pm$0.007 \\
SubTB & 0.328$\pm$0.016 & 0.755$\pm$0.004 & 2.127$\pm$0.037 & 0.836$\pm$0.003 & 8.35$\pm$0.06 & 9.22$\pm$0.05 & 9.97$\pm$0.03 & 0.006$\pm$0.001 & 0.047$\pm$0.005 & 0.076$\pm$0.005 & 0.871$\pm$0.001 \\
SubTB+SubM & 0.298$\pm$0.006 & 0.736$\pm$0.005 & 2.165$\pm$0.006 & 0.851$\pm$0.002 & 8.73$\pm$0.06 & 9.59$\pm$0.08 & 9.99$\pm$0.01 & 0.005$\pm$0.001 & 0.029$\pm$0.003 & 0.057$\pm$0.001 & 0.909$\pm$0.003 \\
\bottomrule
\end{tabular}
}
\caption{All-length averaged SMILES performance and induced length distribution ($L_{\max}=10$; mean$\pm$95\% CI over 6 runs).
Acc is computed over all samples; Score/Entropy/FPDiv are computed on valid samples only; Frac[$\cdot$] is computed over all samples.}
\label{tab:smiles_all_length_avg}
\end{table*}

\begin{table*}[!htbp]
\centering
\renewcommand{\arraystretch}{1.06}
\resizebox{\textwidth}{!}{%
\begin{tabular}{ccccccccccccccccccccccccc}
\toprule
$L$ & \multicolumn{4}{c}{TB} & \multicolumn{4}{c}{SubTB} & \multicolumn{4}{c}{RapTB} & \multicolumn{4}{c}{TB+SubM} & \multicolumn{4}{c}{SubTB+SubM} & \multicolumn{4}{c}{RapTB+SubM} \\
\midrule
 & Acc & Score & Frac & Count & Acc & Score & Frac & Count & Acc & Score & Frac & Count & Acc & Score & Frac & Count & Acc & Score & Frac & Count & Acc & Score & Frac & Count \\
\midrule
1 & 1$\pm$0 & 0.708$\pm$0.004 & 0.215$\pm$0.007 & 687$\pm$22.7 & 1$\pm$0 & 0.762$\pm$0.007 & 0.002$\pm$0.001 & 1.7$\pm$0.7 & 1$\pm$0 & 0.651$\pm$0.004 & 0.042$\pm$0.002 & 133$\pm$4.9 & 1$\pm$0 & 0.712$\pm$0.007 & 0.055$\pm$0.007 & 175$\pm$22.6 & 1$\pm$0 & 0.766$\pm$0.001 & 0.004$\pm$0.003 & 3.7$\pm$2.6 & 1$\pm$0 & 0.702$\pm$0.006 & 0.02$\pm$0.002 & 64.3$\pm$7.5 \\
2 & 0.999$\pm$0.001 & 0.752$\pm$0.002 & 0.387$\pm$0.008 & 1235.7$\pm$28.1 & 1$\pm$0 & 0.748$\pm$0.003 & 0.017$\pm$0.003 & 17.7$\pm$3.5 & 0.999$\pm$0.002 & 0.725$\pm$0.004 & 0.092$\pm$0.006 & 294.7$\pm$17.4 & 1$\pm$0 & 0.79$\pm$0 & 0.088$\pm$0.003 & 282$\pm$7.9 & 1$\pm$0 & 0.75$\pm$0.014 & 0.012$\pm$0.004 & 11.7$\pm$4 & 1$\pm$0 & 0.783$\pm$0.005 & 0.026$\pm$0.003 & 83.3$\pm$10.8 \\
3 & 0.998$\pm$0.002 & 0.727$\pm$0.002 & 0.138$\pm$0.003 & 439.3$\pm$10.5 & 1$\pm$0 & 0.774$\pm$0.01 & 0.009$\pm$0.004 & 9.3$\pm$4 & 1$\pm$0 & 0.754$\pm$0.001 & 0.056$\pm$0.002 & 177$\pm$6.8 & 1$\pm$0 & 0.787$\pm$0.003 & 0.058$\pm$0.008 & 184.3$\pm$26.7 & 1$\pm$0 & 0.759$\pm$0.008 & 0.008$\pm$0.003 & 8$\pm$3 & 0.996$\pm$0.008 & 0.759$\pm$0.006 & 0.03$\pm$0.006 & 95$\pm$19.3 \\
4 & 0.998$\pm$0.003 & 0.722$\pm$0.001 & 0.062$\pm$0.002 & 199.3$\pm$6.8 & 1$\pm$0 & 0.787$\pm$0.003 & 0.035$\pm$0.003 & 36.3$\pm$2.4 & 0.995$\pm$0.006 & 0.748$\pm$0.003 & 0.058$\pm$0.004 & 185.7$\pm$13.6 & 0.997$\pm$0.005 & 0.793$\pm$0.003 & 0.046$\pm$0.005 & 145.3$\pm$16.8 & 1$\pm$0 & 0.789$\pm$0.006 & 0.024$\pm$0.003 & 23.3$\pm$2.4 & 1$\pm$0 & 0.813$\pm$0.012 & 0.031$\pm$0.002 & 99.3$\pm$5.3 \\
5 & 0.996$\pm$0.008 & 0.739$\pm$0.003 & 0.052$\pm$0.002 & 165.3$\pm$5.8 & 1$\pm$0 & 0.813$\pm$0.001 & 0.099$\pm$0.011 & 103.7$\pm$9.9 & 0.998$\pm$0.004 & 0.78$\pm$0.009 & 0.09$\pm$0.005 & 288.3$\pm$14.6 & 0.997$\pm$0.002 & 0.851$\pm$0.004 & 0.083$\pm$0.004 & 263.7$\pm$13.6 & 1$\pm$0 & 0.815$\pm$0.001 & 0.066$\pm$0.011 & 63$\pm$10.4 & 0.995$\pm$0.006 & 0.836$\pm$0.003 & 0.059$\pm$0.004 & 185.3$\pm$12.1 \\
6 & 0.995$\pm$0.005 & 0.736$\pm$0.019 & 0.041$\pm$0.005 & 131$\pm$17.4 & 1$\pm$0 & 0.814$\pm$0.001 & 0.057$\pm$0.015 & 60$\pm$13.8 & 0.997$\pm$0.001 & 0.79$\pm$0.004 & 0.155$\pm$0.007 & 494.7$\pm$23.1 & 0.998$\pm$0.002 & 0.844$\pm$0.005 & 0.094$\pm$0.004 & 301$\pm$13.1 & 1$\pm$0 & 0.819$\pm$0.001 & 0.051$\pm$0.007 & 48.7$\pm$6.2 & 0.995$\pm$0.005 & 0.864$\pm$0.003 & 0.107$\pm$0.008 & 338.3$\pm$24.9 \\
7 & 0.995$\pm$0.009 & 0.732$\pm$0.017 & 0.021$\pm$0.002 & 68.7$\pm$7.4 & 1$\pm$0 & 0.829$\pm$0.003 & 0.086$\pm$0.009 & 90$\pm$9.8 & 0.997$\pm$0.002 & 0.785$\pm$0.004 & 0.17$\pm$0.003 & 540.3$\pm$8.2 & 0.997$\pm$0.002 & 0.86$\pm$0.004 & 0.122$\pm$0.006 & 387.7$\pm$18.9 & 1$\pm$0 & 0.831$\pm$0.005 & 0.064$\pm$0.002 & 60.7$\pm$3.5 & 0.993$\pm$0.003 & 0.854$\pm$0.002 & 0.177$\pm$0.009 & 560$\pm$26.8 \\
8 & 0.995$\pm$0.011 & 0.626$\pm$0.012 & 0.021$\pm$0.002 & 66.3$\pm$5.7 & 1$\pm$0 & 0.834$\pm$0.002 & 0.089$\pm$0.007 & 93.7$\pm$4.7 & 0.994$\pm$0.005 & 0.775$\pm$0.006 & 0.141$\pm$0.009 & 449$\pm$29.3 & 0.998$\pm$0.002 & 0.861$\pm$0.001 & 0.133$\pm$0.007 & 422.7$\pm$23.6 & 1$\pm$0 & 0.837$\pm$0.005 & 0.077$\pm$0.006 & 73$\pm$5.2 & 0.992$\pm$0.006 & 0.842$\pm$0.001 & 0.151$\pm$0.015 & 477.3$\pm$46.2 \\
9 & 1$\pm$0 & 0.529$\pm$0.008 & 0.04$\pm$0.002 & 127.3$\pm$5.1 & 1$\pm$0 & 0.834$\pm$0.002 & 0.081$\pm$0.002 & 84.7$\pm$4 & 0.996$\pm$0.001 & 0.666$\pm$0.005 & 0.141$\pm$0.012 & 448.3$\pm$37.9 & 0.998$\pm$0.003 & 0.88$\pm$0.003 & 0.139$\pm$0.004 & 443.3$\pm$13.6 & 1$\pm$0 & 0.838$\pm$0.005 & 0.051$\pm$0.01 & 48.3$\pm$10.3 & 0.997$\pm$0.003 & 0.854$\pm$0.002 & 0.218$\pm$0.007 & 689.3$\pm$24.3 \\
10 & 0.979$\pm$0.028 & 0.442$\pm$0.013 & 0.023$\pm$0.003 & 74.7$\pm$8.5 & 0.205$\pm$0.019 & 0.699$\pm$0.004 & 0.527$\pm$0.024 & 554$\pm$52 & 0.98$\pm$0.019 & 0.561$\pm$0.01 & 0.055$\pm$0.006 & 176$\pm$19.8 & 0.988$\pm$0.002 & 0.875$\pm$0.002 & 0.183$\pm$0.012 & 583.3$\pm$38.4 & 0.215$\pm$0.007 & 0.69$\pm$0.009 & 0.643$\pm$0.016 & 614.3$\pm$22.5 & 0.956$\pm$0.007 & 0.861$\pm$0.002 & 0.18$\pm$0.003 & 568.3$\pm$7.9 \\
\bottomrule
\end{tabular}
}%

\caption{Per-length valid-only core metrics of SMILES generation (mean$\pm$95\% CI, $L_{\max}=10$).}
\label{tab:bylen_valid_core}
\end{table*}

\begin{table*}[!htbp]
\centering
\scriptsize
\setlength{\tabcolsep}{2.2pt}
\renewcommand{\arraystretch}{1.06}
\resizebox{\textwidth}{!}{%
\begin{tabular}{ccccccccccccc}
\toprule
$L$ & \multicolumn{2}{c}{TB} & \multicolumn{2}{c}{SubTB} & \multicolumn{2}{c}{RapTB} & \multicolumn{2}{c}{TB+SubM} & \multicolumn{2}{c}{SubTB+SubM} & \multicolumn{2}{c}{RapTB+SubM} \\
\midrule
 & Entropy & FPDiv & Entropy & FPDiv & Entropy & FPDiv & Entropy & FPDiv & Entropy & FPDiv & Entropy & FPDiv \\
\midrule
1 & 2.39$\pm$0.07 & 0.789$\pm$0.012 & 0.23$\pm$0.45 & 0.25$\pm$0.283 & 2.28$\pm$0.11 & 0.824$\pm$0.005 & 1.35$\pm$0.09 & 0.628$\pm$0.014 & 0.17$\pm$0.33 & 0.2$\pm$0.226 & 1.32$\pm$0.08 & 0.651$\pm$0.027 \\
2 & 2.54$\pm$0.01 & 0.812$\pm$0.005 & 1.05$\pm$0.33 & 0.445$\pm$0.069 & 2.33$\pm$0.03 & 0.827$\pm$0.003 & 2.12$\pm$0.07 & 0.84$\pm$0.004 & 0.99$\pm$0.09 & 0.435$\pm$0.056 & 1.82$\pm$0.13 & 0.759$\pm$0.022 \\
3 & 2.44$\pm$0.02 & 0.825$\pm$0.001 & 1.1$\pm$0.17 & 0.722$\pm$0.022 & 1.99$\pm$0.03 & 0.818$\pm$0.003 & 2.21$\pm$0.07 & 0.877$\pm$0.002 & 0.87$\pm$0.43 & 0.604$\pm$0.091 & 1.76$\pm$0.04 & 0.838$\pm$0.002 \\
4 & 2.46$\pm$0.05 & 0.854$\pm$0.002 & 1.35$\pm$0.14 & 0.703$\pm$0.03 & 2.02$\pm$0.06 & 0.867$\pm$0.005 & 2.39$\pm$0.01 & 0.909$\pm$0.005 & 1.31$\pm$0.36 & 0.739$\pm$0.05 & 2.19$\pm$0.07 & 0.888$\pm$0.007 \\
5 & 2.41$\pm$0.02 & 0.88$\pm$0.005 & 1.19$\pm$0.11 & 0.714$\pm$0.01 & 2.21$\pm$0.01 & 0.859$\pm$0.003 & 2.48$\pm$0.04 & 0.906$\pm$0.004 & 1.15$\pm$0.05 & 0.721$\pm$0.004 & 2.21$\pm$0.12 & 0.877$\pm$0.015 \\
6 & 2.42$\pm$0.08 & 0.886$\pm$0.002 & 1.7$\pm$0.06 & 0.836$\pm$0.008 & 2.15$\pm$0.04 & 0.88$\pm$0.005 & 2.44$\pm$0.05 & 0.92$\pm$0.001 & 1.69$\pm$0.04 & 0.845$\pm$0.001 & 2.37$\pm$0.03 & 0.886$\pm$0.007 \\
7 & 2.35$\pm$0.02 & 0.891$\pm$0.003 & 1.5$\pm$0.05 & 0.836$\pm$0.007 & 2.18$\pm$0.02 & 0.884$\pm$0.001 & 2.49$\pm$0.01 & 0.926$\pm$0.001 & 1.68$\pm$0.02 & 0.864$\pm$0.008 & 2.47$\pm$0.03 & 0.919$\pm$0.003 \\
8 & 1.61$\pm$0.23 & 0.703$\pm$0.049 & 1.62$\pm$0.03 & 0.834$\pm$0.012 & 2.17$\pm$0.02 & 0.882$\pm$0.001 & 2.54$\pm$0.01 & 0.918$\pm$0.001 & 1.55$\pm$0.06 & 0.829$\pm$0.01 & 2.59$\pm$0.03 & 0.921$\pm$0.001 \\
9 & 1.35$\pm$0.05 & 0.693$\pm$0.016 & 1.71$\pm$0.07 & 0.853$\pm$0.002 & 1.98$\pm$0.04 & 0.848$\pm$0.006 & 2.36$\pm$0.02 & 0.909$\pm$0.001 & 1.67$\pm$0.05 & 0.857$\pm$0.008 & 2.28$\pm$0.05 & 0.916$\pm$0.001 \\
10 & 1.06$\pm$0.11 & 0.58$\pm$0.023 & 2.16$\pm$0.04 & 0.881$\pm$0 & 1.75$\pm$0.03 & 0.815$\pm$0.004 & 2.08$\pm$0.02 & 0.906$\pm$0.001 & 2.18$\pm$0.01 & 0.884$\pm$0.003 & 2.16$\pm$0.02 & 0.91$\pm$0 \\
\bottomrule
\end{tabular}
}%
\caption{Per-length valid-only diversity metrics of SMILES generation(mean$\pm$95\% CI, $L_{\max}=10$).}
\label{tab:bylen_valid_div}
\end{table*}

\begin{table*}[!htbp]
\centering
\scriptsize
\setlength{\tabcolsep}{2.2pt}
\renewcommand{\arraystretch}{1.06}
\resizebox{\textwidth}{!}{%
\begin{tabular}{ccccccccccccccccccccccccc}
\toprule
$L$ & \multicolumn{4}{c}{TB} & \multicolumn{4}{c}{SubTB} & \multicolumn{4}{c}{RapTB} & \multicolumn{4}{c}{TB+SubM} & \multicolumn{4}{c}{SubTB+SubM} & \multicolumn{4}{c}{RapTB+SubM} \\
\midrule
 & UniqStr & UniqMol & UniqRateStr & UniqRateMol & UniqStr & UniqMol & UniqRateStr & UniqRateMol & UniqStr & UniqMol & UniqRateStr & UniqRateMol & UniqStr & UniqMol & UniqRateStr & UniqRateMol & UniqStr & UniqMol & UniqRateStr & UniqRateMol & UniqStr & UniqMol & UniqRateStr & UniqRateMol \\
\midrule
1 & 29.3$\pm$1.1 & 25.3$\pm$1.5 & 0$\pm$0 & 0$\pm$0 & 1.3$\pm$0.4 & 1.3$\pm$0.4 & 0.8$\pm$0.2 & 0.8$\pm$0.2 & 17.3$\pm$0.8 & 15.7$\pm$0.4 & 0.1$\pm$0 & 0.1$\pm$0 & 13.7$\pm$0.8 & 13.7$\pm$0.8 & 0.1$\pm$0 & 0.1$\pm$0 & 1.3$\pm$0.4 & 1.3$\pm$0.4 & 0.5$\pm$0.3 & 0.5$\pm$0.3 & 8.3$\pm$0.8 & 8.3$\pm$0.8 & 0.1$\pm$0 & 0.1$\pm$0 \\
2 & 184.7$\pm$2.1 & 115.7$\pm$3 & 0.1$\pm$0 & 0.2$\pm$0 & 6.3$\pm$0.8 & 6$\pm$0.7 & 0.4$\pm$0 & 0.3$\pm$0 & 77$\pm$1.9 & 71$\pm$2.5 & 0.3$\pm$0 & 0.2$\pm$0 & 60.3$\pm$3 & 59.7$\pm$2.9 & 0.2$\pm$0 & 0.2$\pm$0 & 5$\pm$0.7 & 5$\pm$0.7 & 0.4$\pm$0 & 0.4$\pm$0 & 27.3$\pm$1.5 & 27.3$\pm$1.5 & 0.3$\pm$0 & 0.3$\pm$0 \\
3 & 224$\pm$7.5 & 215.7$\pm$7.3 & 0.5$\pm$0 & 0.5$\pm$0 & 6.3$\pm$1.7 & 6.3$\pm$1.7 & 0.7$\pm$0.1 & 0.7$\pm$0.1 & 78.3$\pm$2.2 & 76.3$\pm$2.2 & 0.4$\pm$0 & 0.4$\pm$0 & 52.3$\pm$5.8 & 52.3$\pm$5.8 & 0.3$\pm$0 & 0.3$\pm$0 & 4.7$\pm$1.8 & 4.7$\pm$1.8 & 0.6$\pm$0.1 & 0.6$\pm$0.1 & 27.7$\pm$3 & 27.7$\pm$3 & 0.3$\pm$0 & 0.3$\pm$0 \\
4 & 166$\pm$4 & 164$\pm$3.6 & 0.8$\pm$0 & 0.8$\pm$0 & 17$\pm$1.2 & 16.7$\pm$0.8 & 0.5$\pm$0.1 & 0.5$\pm$0 & 111$\pm$3.1 & 107.3$\pm$3 & 0.6$\pm$0 & 0.6$\pm$0 & 58$\pm$0.7 & 57.7$\pm$1.1 & 0.4$\pm$0 & 0.4$\pm$0 & 14.7$\pm$2.2 & 13.7$\pm$1.8 & 0.6$\pm$0.1 & 0.6$\pm$0.1 & 41$\pm$6.1 & 41$\pm$6.1 & 0.4$\pm$0.1 & 0.4$\pm$0.1 \\
5 & 156$\pm$5.9 & 153.3$\pm$5.8 & 0.9$\pm$0 & 0.9$\pm$0 & 41.3$\pm$3.7 & 38.3$\pm$3.9 & 0.4$\pm$0 & 0.4$\pm$0 & 187$\pm$7.6 & 181.7$\pm$7.9 & 0.6$\pm$0 & 0.6$\pm$0 & 101.3$\pm$3.5 & 100.7$\pm$3.3 & 0.4$\pm$0 & 0.4$\pm$0 & 31.3$\pm$4.2 & 28.3$\pm$3.5 & 0.5$\pm$0 & 0.4$\pm$0 & 74$\pm$8.1 & 73.7$\pm$7.9 & 0.4$\pm$0 & 0.4$\pm$0 \\
6 & 129.7$\pm$10.2 & 128.7$\pm$9.5 & 1$\pm$0 & 1$\pm$0 & 43.3$\pm$6.6 & 40$\pm$6.6 & 0.7$\pm$0 & 0.7$\pm$0 & 332.3$\pm$4.2 & 322$\pm$5 & 0.7$\pm$0 & 0.7$\pm$0 & 130.3$\pm$12.8 & 130$\pm$12.4 & 0.4$\pm$0 & 0.4$\pm$0 & 39$\pm$5.4 & 37$\pm$5.2 & 0.8$\pm$0 & 0.8$\pm$0 & 123.3$\pm$6 & 123.3$\pm$6 & 0.4$\pm$0 & 0.4$\pm$0 \\
7 & 68$\pm$4.7 & 68$\pm$4.7 & 1$\pm$0 & 1$\pm$0 & 60$\pm$3.1 & 55$\pm$2.5 & 0.7$\pm$0 & 0.6$\pm$0 & 448.7$\pm$7.2 & 411.7$\pm$3.3 & 0.8$\pm$0 & 0.8$\pm$0 & 157$\pm$8.7 & 156.3$\pm$8.4 & 0.4$\pm$0 & 0.4$\pm$0 & 52.3$\pm$3 & 50.3$\pm$1.8 & 0.9$\pm$0 & 0.8$\pm$0 & 235$\pm$6.1 & 220.7$\pm$0.4 & 0.4$\pm$0 & 0.4$\pm$0 \\
8 & 35$\pm$5 & 35$\pm$5 & 0.5$\pm$0.1 & 0.5$\pm$0.1 & 73.3$\pm$8.1 & 69.3$\pm$7 & 0.8$\pm$0.1 & 0.7$\pm$0.1 & 409.7$\pm$15.1 & 400$\pm$12.5 & 0.9$\pm$0 & 0.9$\pm$0 & 163$\pm$6.8 & 162.3$\pm$6.5 & 0.4$\pm$0 & 0.4$\pm$0 & 58.7$\pm$3.2 & 56.3$\pm$3.2 & 0.8$\pm$0 & 0.8$\pm$0 & 299$\pm$13.6 & 298.3$\pm$13.7 & 0.6$\pm$0 & 0.6$\pm$0 \\
9 & 38.7$\pm$1.1 & 38.7$\pm$1.1 & 0.3$\pm$0 & 0.3$\pm$0 & 76$\pm$6.2 & 72$\pm$5.2 & 0.9$\pm$0.1 & 0.8$\pm$0 & 246.3$\pm$6 & 244.3$\pm$5.4 & 0.6$\pm$0 & 0.5$\pm$0 & 159.7$\pm$4.3 & 159.3$\pm$4.7 & 0.4$\pm$0 & 0.4$\pm$0 & 45.7$\pm$5.4 & 43.3$\pm$5.5 & 0.9$\pm$0 & 0.9$\pm$0 & 353.7$\pm$5 & 291.7$\pm$5.3 & 0.5$\pm$0 & 0.6$\pm$0 \\
10 & 33$\pm$4.4 & 33$\pm$4.4 & 0.4$\pm$0 & 0.4$\pm$0 & 482.3$\pm$31.8 & 435.3$\pm$14.1 & 0.9$\pm$0 & 0.9$\pm$0 & 96.3$\pm$3.3 & 95.7$\pm$2.9 & 0.5$\pm$0 & 0.5$\pm$0 & 156.3$\pm$4.6 & 145.3$\pm$3.7 & 0.3$\pm$0 & 0.3$\pm$0 & 558$\pm$8.7 & 467.3$\pm$4.2 & 0.9$\pm$0 & 0.9$\pm$0 & 242$\pm$5.9 & 225.3$\pm$2.9 & 0.4$\pm$0 & 0.4$\pm$0 \\
\bottomrule
\end{tabular}
}%
\caption{Per-length valid-only uniqueness metrics of SMILES generation (mean$\pm$95\% CI, $L_{\max}=10$).}
\label{tab:smiles_L10_bylen_valid_uniq}
\end{table*}

\subsubsection{Per-length prefix collapse analysis of SMILES generation.}
\label{app:results_smiles_L10_prefix}
Table~\ref{tab:prefix_bylen_base} reports prefix distributions at depth $k$ (mean$\pm$95\% CI), where Survival should be read jointly with concentration metrics (PefEnt/Eff/Top1/UniqueRate).
Without SubM, TB exhibits rapid attrition (e.g., Survival drops to $0.105$ at $k{=}7$ and $0.023$ at $k{=}10$) and increased concentration among the remaining deep prefixes (Top1 reaches $0.406$ at $k{=}7$), consistent with prefix collapse.
RapTB substantially improves Survival at larger $k$ (e.g., $0.497$ at $k{=}7$) while reducing deep-prefix concentration (Top1 $0.132$ at $k{=}7$), indicating more sustained branching beyond early decisions.
Enabling SubM (Table~\ref{tab:prefix_bylen_subm}) further alleviates collapse for both TB and RapTB by increasing deep-prefix diversity (higher PefEnt/Eff, lower Top1) and improving Survival at large $k$ (e.g., TB: $0.023\!\to\!0.183$ and RapTB: $0.055\!\to\!0.180$ at $k{=}10$).

\begin{table*}[!htbp]
\centering
\scriptsize
\setlength{\tabcolsep}{2.2pt}
\renewcommand{\arraystretch}{1.06}
\resizebox{\textwidth}{!}{%
\begin{tabular}{cccccccccccccccc}
\toprule
$k$ & \multicolumn{5}{c}{TB} & \multicolumn{5}{c}{SubTB} & \multicolumn{5}{c}{RapTB} \\
\midrule
 & Survival & PefEnt & Eff & Top1 & UniqueRate
 & Survival & PefEnt & Eff & Top1 & UniqueRate
 & Survival & PefEnt & Eff & Top1 & UniqueRate \\
\midrule
1 & 1$\pm$0 & 2.531$\pm$0.022 & 12.57$\pm$0.28 & 0.274$\pm$0.01 & 0.013$\pm$0.001 & 1$\pm$0 & 1.539$\pm$0.036 & 4.66$\pm$0.17 & 0.419$\pm$0.01 & 0.017$\pm$0.001 & 1$\pm$0 & 1.841$\pm$0.025 & 6.3$\pm$0.15 & 0.4$\pm$0.004 & 0.01$\pm$0 \\
2 & 0.785$\pm$0.005 & 4.748$\pm$0.019 & 115.34$\pm$2.25 & 0.077$\pm$0.001 & 0.135$\pm$0.002 & 0.998$\pm$0 & 3.544$\pm$0.063 & 34.69$\pm$2.15 & 0.096$\pm$0.004 & 0.096$\pm$0.002 & 0.958$\pm$0.001 & 3.838$\pm$0.023 & 46.46$\pm$1.05 & 0.174$\pm$0.005 & 0.076$\pm$0.002 \\
3 & 0.398$\pm$0.002 & 5.418$\pm$0.027 & 225.52$\pm$6.1 & 0.121$\pm$0.005 & 0.45$\pm$0.005 & 0.982$\pm$0.002 & 4.563$\pm$0.044 & 95.94$\pm$4.19 & 0.088$\pm$0.006 & 0.24$\pm$0.003 & 0.866$\pm$0.005 & 4.717$\pm$0.031 & 111.95$\pm$3.52 & 0.184$\pm$0.005 & 0.215$\pm$0.005 \\
4 & 0.261$\pm$0.002 & 5.17$\pm$0.038 & 176.02$\pm$6.81 & 0.167$\pm$0.009 & 0.566$\pm$0.012 & 0.973$\pm$0.004 & 5.054$\pm$0.035 & 156.78$\pm$5.48 & 0.088$\pm$0.006 & 0.355$\pm$0.003 & 0.81$\pm$0.006 & 5.45$\pm$0.047 & 233.1$\pm$11.25 & 0.135$\pm$0.007 & 0.357$\pm$0.01 \\
5 & 0.198$\pm$0.002 & 4.796$\pm$0.041 & 121.2$\pm$4.96 & 0.219$\pm$0.014 & 0.599$\pm$0.009 & 0.938$\pm$0.005 & 5.575$\pm$0.048 & 264.06$\pm$12.39 & 0.03$\pm$0.003 & 0.471$\pm$0.004 & 0.752$\pm$0.004 & 5.747$\pm$0.051 & 313.88$\pm$16.34 & 0.138$\pm$0.007 & 0.474$\pm$0.01 \\
6 & 0.146$\pm$0.003 & 4.051$\pm$0.075 & 57.65$\pm$4.37 & 0.293$\pm$0.021 & 0.533$\pm$0.007 & 0.839$\pm$0.011 & 6.012$\pm$0.046 & 408.87$\pm$18.66 & 0.026$\pm$0.001 & 0.638$\pm$0.006 & 0.662$\pm$0.006 & 6.027$\pm$0.044 & 414.97$\pm$17.93 & 0.102$\pm$0.004 & 0.575$\pm$0.012 \\
7 & 0.105$\pm$0.002 & 2.928$\pm$0.02 & 18.7$\pm$0.37 & 0.406$\pm$0.019 & 0.363$\pm$0.008 & 0.782$\pm$0.012 & 6.175$\pm$0.07 & 482.25$\pm$33.04 & 0.022$\pm$0.002 & 0.738$\pm$0.01 & 0.506$\pm$0.004 & 6.011$\pm$0.05 & 408.6$\pm$20.46 & 0.132$\pm$0.004 & 0.637$\pm$0.01 \\
8 & 0.084$\pm$0.001 & 3.018$\pm$0.042 & 20.48$\pm$0.84 & 0.217$\pm$0.008 & 0.246$\pm$0.012 & 0.696$\pm$0.011 & 6.213$\pm$0.081 & 501.27$\pm$39.04 & 0.02$\pm$0 & 0.803$\pm$0.016 & 0.337$\pm$0.003 & 5.492$\pm$0.072 & 243.57$\pm$17.66 & 0.126$\pm$0.011 & 0.6$\pm$0.013 \\
9 & 0.063$\pm$0.002 & 3.378$\pm$0.059 & 29.37$\pm$1.76 & 0.159$\pm$0.008 & 0.27$\pm$0.009 & 0.607$\pm$0.015 & 6.124$\pm$0.076 & 458.44$\pm$33.86 & 0.022$\pm$0.001 & 0.83$\pm$0.015 & 0.196$\pm$0.009 & 4.774$\pm$0.031 & 118.44$\pm$3.61 & 0.16$\pm$0.01 & 0.474$\pm$0.012 \\
10 & 0.023$\pm$0.002 & 3.119$\pm$0.131 & 22.88$\pm$3.02 & 0.185$\pm$0.032 & 0.442$\pm$0.045 & 0.527$\pm$0.015 & 6.065$\pm$0.076 & 431.97$\pm$31.82 & 0.022$\pm$0.003 & 0.871$\pm$0.01 & 0.055$\pm$0.004 & 4.276$\pm$0.022 & 71.97$\pm$1.57 & 0.049$\pm$0.007 & 0.549$\pm$0.023 \\
\bottomrule
\end{tabular}
}%
\caption{Prefix statistics by depth. Mean$\pm$95\% CI.}
\label{tab:prefix_bylen_base}
\end{table*}

\begin{table*}[!htbp]
\centering
\scriptsize
\setlength{\tabcolsep}{2.2pt}
\renewcommand{\arraystretch}{1.06}
\resizebox{\textwidth}{!}{%
\begin{tabular}{cccccccccccccccc}
\toprule
$k$ & \multicolumn{5}{c}{TB+SubM} & \multicolumn{5}{c}{SubTB+SubM} & \multicolumn{5}{c}{RapTB+SubM} \\
\midrule
 & Survival & PefEnt & Eff & Top1 & UniqueRate
 & Survival & PefEnt & Eff & Top1 & UniqueRate
 & Survival & PefEnt & Eff & Top1 & UniqueRate \\
\midrule
1 & 1$\pm$0 & 2.715$\pm$0.017 & 15.11$\pm$0.25 & 0.165$\pm$0.006 & 0.011$\pm$0 & 1$\pm$0 & 1.559$\pm$0.037 & 4.76$\pm$0.18 & 0.394$\pm$0.01 & 0.017$\pm$0.001 & 1$\pm$0 & 2.563$\pm$0.004 & 12.98$\pm$0.05 & 0.193$\pm$0.005 & 0.009$\pm$0 \\
2 & 0.945$\pm$0.004 & 4.778$\pm$0.004 & 118.92$\pm$0.46 & 0.052$\pm$0.003 & 0.094$\pm$0.001 & 0.996$\pm$0.002 & 3.669$\pm$0.041 & 39.26$\pm$1.58 & 0.094$\pm$0.005 & 0.108$\pm$0.004 & 0.98$\pm$0.001 & 4.43$\pm$0.015 & 83.96$\pm$1.27 & 0.065$\pm$0.001 & 0.076$\pm$0.001 \\
3 & 0.857$\pm$0.004 & 5.224$\pm$0.01 & 185.6$\pm$1.81 & 0.033$\pm$0.002 & 0.164$\pm$0.005 & 0.984$\pm$0.002 & 4.723$\pm$0.024 & 112.55$\pm$2.69 & 0.089$\pm$0.004 & 0.278$\pm$0.008 & 0.953$\pm$0.003 & 5.019$\pm$0.022 & 151.34$\pm$3.33 & 0.042$\pm$0.006 & 0.141$\pm$0.001 \\
4 & 0.799$\pm$0.005 & 5.434$\pm$0.014 & 229.16$\pm$3.26 & 0.034$\pm$0.002 & 0.211$\pm$0.006 & 0.975$\pm$0.004 & 5.275$\pm$0.021 & 195.48$\pm$4.12 & 0.09$\pm$0.004 & 0.437$\pm$0 & 0.923$\pm$0.007 & 5.342$\pm$0.036 & 209.02$\pm$7.37 & 0.037$\pm$0.002 & 0.196$\pm$0.004 \\
5 & 0.753$\pm$0.008 & 5.667$\pm$0.021 & 289.19$\pm$6 & 0.024$\pm$0.002 & 0.254$\pm$0.007 & 0.951$\pm$0.004 & 5.765$\pm$0.014 & 318.94$\pm$4.35 & 0.029$\pm$0.001 & 0.563$\pm$0 & 0.892$\pm$0.006 & 5.641$\pm$0.047 & 282.07$\pm$13.11 & 0.036$\pm$0.004 & 0.254$\pm$0.006 \\
6 & 0.671$\pm$0.01 & 5.75$\pm$0.019 & 314.29$\pm$5.93 & 0.019$\pm$0.001 & 0.29$\pm$0.006 & 0.885$\pm$0.007 & 6.136$\pm$0.02 & 462.51$\pm$9.06 & 0.025$\pm$0.002 & 0.704$\pm$0.014 & 0.833$\pm$0.004 & 6.002$\pm$0.019 & 404.41$\pm$7.56 & 0.025$\pm$0.001 & 0.326$\pm$0.002 \\
7 & 0.576$\pm$0.01 & 5.617$\pm$0.009 & 275.21$\pm$2.49 & 0.022$\pm$0.001 & 0.299$\pm$0.003 & 0.834$\pm$0.006 & 6.29$\pm$0.021 & 539.52$\pm$11.44 & 0.015$\pm$0.002 & 0.793$\pm$0.013 & 0.726$\pm$0.009 & 6.189$\pm$0.023 & 487.47$\pm$11.36 & 0.022$\pm$0.003 & 0.404$\pm$0.001 \\
8 & 0.455$\pm$0.007 & 5.363$\pm$0.015 & 213.33$\pm$3.24 & 0.028$\pm$0.001 & 0.293$\pm$0.002 & 0.771$\pm$0.005 & 6.292$\pm$0.016 & 540.19$\pm$8.66 & 0.015$\pm$0.002 & 0.832$\pm$0.012 & 0.549$\pm$0.005 & 6.116$\pm$0.022 & 453.29$\pm$9.84 & 0.02$\pm$0.002 & 0.456$\pm$0.003 \\
9 & 0.322$\pm$0.005 & 4.997$\pm$0.014 & 148.03$\pm$2.02 & 0.039$\pm$0.001 & 0.284$\pm$0.006 & 0.694$\pm$0.008 & 6.242$\pm$0.014 & 513.7$\pm$6.98 & 0.016$\pm$0.002 & 0.865$\pm$0.011 & 0.398$\pm$0.005 & 5.79$\pm$0.017 & 327.02$\pm$5.68 & 0.024$\pm$0.003 & 0.445$\pm$0.005 \\
10 & 0.183$\pm$0.008 & 4.447$\pm$0.025 & 85.38$\pm$2.14 & 0.053$\pm$0.007 & 0.268$\pm$0.012 & 0.643$\pm$0.01 & 6.262$\pm$0.013 & 524.33$\pm$6.63 & 0.01$\pm$0.001 & 0.909$\pm$0.01 & 0.18$\pm$0.002 & 4.936$\pm$0.021 & 139.26$\pm$2.9 & 0.046$\pm$0.006 & 0.426$\pm$0.007 \\
\bottomrule
\end{tabular}
}%
\caption{Prefix statistics by depth (Continue). Mean$\pm$95\% CI.}
\label{tab:prefix_bylen_subm}
\end{table*}

\subsubsection{Per-length SMILES performance on long horizon.}
Table~\ref{tab:smiles_L15_bylen_valid_core}--\ref{tab:smiles_L15_bylen_valid_uniq} show that increasing the horizon amplifies length-wise failure modes.
TB rapidly loses support on long valid trajectories: its Frac/Count becomes negligible beyond $L\ge 12$ (e.g., Frac $0.003$ at $L=12$ and effectively zero thereafter), and the corresponding Score degrades sharply at long lengths (e.g., $0.544/0.471/0.433/0.375$ for $L=9$--$12$), indicating severe long-horizon under-coverage.
RapTB maintains substantially higher terminal quality on long trajectories (Score $\approx 0.75$--$0.81$ for $L=10$--$14$ with near-perfect Acc), but still under-allocates mass to the extreme tail ($L=15$) as reflected by a small Frac/Count.
Combining RapTB with SubM shifts probability mass back to long lengths without sacrificing quality, yielding strong tail performance (at $L=15$, Acc $0.84$ and Score $0.85$ with a much larger Frac/Count), and also improves long-length diversity/uniqueness compared to RapTB alone (Tables~\ref{tab:smiles_L15_bylen_valid_div}, \ref{tab:smiles_L15_bylen_valid_uniq}).
In contrast, SubTB places substantial mass on very long lengths (e.g., large Frac at $L=15$) but exhibits low Acc/Score there, consistent with the termination/length instability discussed in the main text.
\label{app:results_smiles_L15_main}

\begin{table*}[!htbp]
\centering
\scriptsize
\setlength{\tabcolsep}{2.2pt}
\renewcommand{\arraystretch}{1.06}
\resizebox{\textwidth}{!}{%
\begin{tabular}{ccccccccccccccccc}
\toprule
$L$ & \multicolumn{4}{c}{TB} & \multicolumn{4}{c}{SubTB} & \multicolumn{4}{c}{RapTB} & \multicolumn{4}{c}{RapTB+SubM} \\
\midrule
 & Acc & Score & Frac & Count & Acc & Score & Frac & Count & Acc & Score & Frac & Count & Acc & Score & Frac & Count \\
\midrule
1 & 1$\pm$0 & 0.707$\pm$0.003 & 0.242$\pm$0.012 & 774.7$\pm$37.9 & 1$\pm$0 & 0.765$\pm$0 & 0.002$\pm$0.001 & 1.7$\pm$0.7 & 1$\pm$0 & 0.7$\pm$0.005 & 0.033$\pm$0.004 & 105.7$\pm$11.6 & 1$\pm$0 & 0.694$\pm$0.012 & 0.018$\pm$0.001 & 56.5$\pm$2.9 \\
2 & 1$\pm$0 & 0.748$\pm$0 & 0.34$\pm$0.01 & 1086.7$\pm$33.2 & 1$\pm$0 & 0.738$\pm$0.009 & 0.05$\pm$0.002 & 51.3$\pm$2.6 & 1$\pm$0 & 0.723$\pm$0.003 & 0.033$\pm$0.005 & 103$\pm$16.7 & 1$\pm$0 & 0.798$\pm$0.004 & 0.018$\pm$0.008 & 57.5$\pm$26.5 \\
3 & 0.999$\pm$0.002 & 0.724$\pm$0.007 & 0.128$\pm$0.005 & 409$\pm$16.3 & 1$\pm$0 & 0.723$\pm$0.016 & 0.03$\pm$0.006 & 31$\pm$6 & 1$\pm$0 & 0.747$\pm$0.012 & 0.011$\pm$0.001 & 34.3$\pm$2.4 & 1$\pm$0 & 0.824$\pm$0.003 & 0.016$\pm$0.004 & 50.5$\pm$12.7 \\
4 & 0.999$\pm$0.002 & 0.739$\pm$0 & 0.088$\pm$0.006 & 281.3$\pm$20.3 & 1$\pm$0 & 0.796$\pm$0.008 & 0.069$\pm$0.011 & 70.7$\pm$11.3 & 0.995$\pm$0.01 & 0.758$\pm$0.005 & 0.017$\pm$0.002 & 55.3$\pm$7.5 & 1$\pm$0 & 0.853$\pm$0.017 & 0.016$\pm$0.006 & 51$\pm$17.6 \\
5 & 0.997$\pm$0.003 & 0.749$\pm$0.001 & 0.06$\pm$0.003 & 192.3$\pm$10.3 & 1$\pm$0 & 0.795$\pm$0.003 & 0.135$\pm$0.004 & 137$\pm$5.2 & 1$\pm$0 & 0.779$\pm$0.01 & 0.019$\pm$0.008 & 60$\pm$25.5 & 1$\pm$0 & 0.888$\pm$0.01 & 0.025$\pm$0.001 & 77$\pm$2 \\
6 & 1$\pm$0 & 0.736$\pm$0.011 & 0.043$\pm$0.002 & 138.3$\pm$7.3 & 1$\pm$0 & 0.794$\pm$0.009 & 0.066$\pm$0.012 & 66.7$\pm$11.4 & 0.993$\pm$0.013 & 0.798$\pm$0.014 & 0.017$\pm$0.002 & 55.3$\pm$6.8 & 1$\pm$0 & 0.879$\pm$0.004 & 0.031$\pm$0.001 & 95.5$\pm$4.9 \\
7 & 0.995$\pm$0.01 & 0.741$\pm$0.021 & 0.022$\pm$0.002 & 69.3$\pm$7.5 & 0.997$\pm$0.005 & 0.811$\pm$0.004 & 0.138$\pm$0.018 & 140.7$\pm$17.4 & 0.99$\pm$0.011 & 0.799$\pm$0.012 & 0.03$\pm$0.002 & 95.7$\pm$5.7 & 0.996$\pm$0.008 & 0.893$\pm$0.006 & 0.034$\pm$0.008 & 107$\pm$23.5 \\
8 & 0.982$\pm$0.018 & 0.705$\pm$0.021 & 0.012$\pm$0.001 & 37$\pm$4.5 & 1$\pm$0 & 0.821$\pm$0.004 & 0.107$\pm$0.009 & 108.7$\pm$9.2 & 0.987$\pm$0.006 & 0.776$\pm$0.009 & 0.031$\pm$0.003 & 99$\pm$7.9 & 0.995$\pm$0.011 & 0.863$\pm$0.013 & 0.029$\pm$0.001 & 91$\pm$2 \\
9 & 0.997$\pm$0.005 & 0.544$\pm$0.024 & 0.032$\pm$0.006 & 101$\pm$19.7 & 1$\pm$0 & 0.808$\pm$0.003 & 0.07$\pm$0.003 & 71$\pm$3 & 0.997$\pm$0.005 & 0.706$\pm$0.003 & 0.084$\pm$0.009 & 266.3$\pm$28 & 1$\pm$0 & 0.871$\pm$0.02 & 0.045$\pm$0.009 & 141$\pm$27.4 \\
10 & 1$\pm$0 & 0.471$\pm$0.002 & 0.021$\pm$0.001 & 65.7$\pm$2.4 & 1$\pm$0 & 0.818$\pm$0.007 & 0.062$\pm$0.004 & 63.3$\pm$4 & 0.997$\pm$0.001 & 0.751$\pm$0.008 & 0.155$\pm$0.008 & 490.3$\pm$22.9 & 1$\pm$0 & 0.862$\pm$0.001 & 0.066$\pm$0.004 & 204$\pm$13.7 \\
11 & 1$\pm$0 & 0.433$\pm$0.034 & 0.009$\pm$0.002 & 28$\pm$6.3 & 1$\pm$0 & 0.826$\pm$0.011 & 0.041$\pm$0.006 & 42$\pm$5.9 & 0.995$\pm$0.001 & 0.757$\pm$0.002 & 0.154$\pm$0.003 & 487.7$\pm$10.5 & 0.988$\pm$0.017 & 0.847$\pm$0.008 & 0.094$\pm$0.004 & 292$\pm$11.8 \\
12 & 1$\pm$0 & 0.375$\pm$0.014 & 0.003$\pm$0.001 & 10$\pm$3.4 & 1$\pm$0 & 0.803$\pm$0.013 & 0.023$\pm$0.003 & 23.3$\pm$3.5 & 0.995$\pm$0.004 & 0.787$\pm$0.002 & 0.158$\pm$0.007 & 498.7$\pm$24.2 & 0.996$\pm$0.003 & 0.853$\pm$0.002 & 0.126$\pm$0.01 & 391.5$\pm$32.3 \\
13 & 1$\pm$0 & 0.711$\pm$0 & 0$\pm$0 & 1$\pm$0 & 1$\pm$0 & 0.797$\pm$0.013 & 0.011$\pm$0.003 & 11.7$\pm$3.6 & 0.995$\pm$0.001 & 0.807$\pm$0.002 & 0.155$\pm$0.011 & 489$\pm$33.3 & 0.993$\pm$0.002 & 0.832$\pm$0.005 & 0.16$\pm$0.003 & 499$\pm$9.8 \\
14 & 1$\pm$0 & 0.332$\pm$0.112 & 0.001$\pm$0 & 3.7$\pm$1.3 & 1$\pm$0 & 0.832$\pm$0.014 & 0.009$\pm$0.004 & 8.7$\pm$3.6 & 0.997$\pm$0.003 & 0.795$\pm$0.007 & 0.08$\pm$0.009 & 254$\pm$27.5 & 0.992$\pm$0.006 & 0.853$\pm$0.005 & 0.192$\pm$0.003 & 596$\pm$7.8 \\
15 & -- & -- & -- & -- & 0.247$\pm$0.01 & 0.494$\pm$0.02 & 0.187$\pm$0.007 & 190.3$\pm$7.7 & 0.748$\pm$0.041 & 0.763$\pm$0.013 & 0.022$\pm$0.001 & 68.7$\pm$4 & 0.84$\pm$0.023 & 0.85$\pm$0.006 & 0.129$\pm$0.005 & 400$\pm$17.6 \\
\bottomrule
\end{tabular}
}%
\caption{Per-length valid-only core metrics of SMILES generation (mean$\pm$95\% CI, $L_{\max}=15$).}
\label{tab:smiles_L15_bylen_valid_core}
\end{table*}

\begin{table*}[!htbp]
\centering
\scriptsize
\renewcommand{\arraystretch}{1.06}
\resizebox{\textwidth}{!}{%
\begin{tabular}{ccccccccc}
\toprule
$L$ & \multicolumn{2}{c}{TB} & \multicolumn{2}{c}{SubTB} & \multicolumn{2}{c}{RapTB} & \multicolumn{2}{c}{RapTB+SubM} \\
\midrule
 & Entropy & FPDiv & Entropy & FPDiv & Entropy & FPDiv & Entropy & FPDiv \\
\midrule
1 & 2.45$\pm$0.04 & 0.82$\pm$0.011 & 0$\pm$0 & 0$\pm$0 & 1.8$\pm$0.06 & 0.625$\pm$0.009 & 0.92$\pm$0.03 & 0.455$\pm$0.024 \\
2 & 2.57$\pm$0 & 0.813$\pm$0.004 & 0.83$\pm$0.21 & 0.344$\pm$0.071 & 2.07$\pm$0.12 & 0.802$\pm$0.004 & 1.04$\pm$0.2 & 0.661$\pm$0.025 \\
3 & 2.49$\pm$0 & 0.834$\pm$0 & 1.49$\pm$0.05 & 0.793$\pm$0.005 & 1.6$\pm$0.14 & 0.706$\pm$0.02 & 1.21$\pm$0.24 & 0.682$\pm$0.027 \\
4 & 2.18$\pm$0.03 & 0.791$\pm$0.009 & 1.21$\pm$0.08 & 0.673$\pm$0.012 & 1.27$\pm$0.23 & 0.748$\pm$0.04 & 1.17$\pm$0.06 & 0.653$\pm$0.026 \\
5 & 2.42$\pm$0.04 & 0.868$\pm$0.003 & 1.25$\pm$0.02 & 0.748$\pm$0.007 & 1.72$\pm$0.05 & 0.825$\pm$0.008 & 1.32$\pm$0.19 & 0.793$\pm$0.017 \\
6 & 2.5$\pm$0.06 & 0.879$\pm$0.004 & 1.62$\pm$0.02 & 0.841$\pm$0.005 & 1.53$\pm$0.19 & 0.79$\pm$0.028 & 1.73$\pm$0.04 & 0.879$\pm$0.003 \\
7 & 2.32$\pm$0 & 0.879$\pm$0.003 & 0.93$\pm$0.05 & 0.705$\pm$0.015 & 1.55$\pm$0.11 & 0.826$\pm$0.006 & 1.69$\pm$0.1 & 0.857$\pm$0.013 \\
8 & 2.23$\pm$0.03 & 0.881$\pm$0.005 & 1.19$\pm$0.09 & 0.796$\pm$0.007 & 1.82$\pm$0.02 & 0.835$\pm$0.003 & 2.05$\pm$0.1 & 0.873$\pm$0.007 \\
9 & 1.51$\pm$0.17 & 0.718$\pm$0.019 & 1.43$\pm$0.03 & 0.846$\pm$0.001 & 1.75$\pm$0.04 & 0.808$\pm$0.005 & 1.91$\pm$0.06 & 0.872$\pm$0.009 \\
10 & 1.18$\pm$0.04 & 0.648$\pm$0.006 & 1.2$\pm$0.13 & 0.782$\pm$0.019 & 1.72$\pm$0.02 & 0.833$\pm$0.003 & 2.17$\pm$0.03 & 0.891$\pm$0.007 \\
11 & 0.82$\pm$0.23 & 0.524$\pm$0.048 & 1.33$\pm$0.07 & 0.835$\pm$0.006 & 1.74$\pm$0.01 & 0.835$\pm$0.001 & 2.08$\pm$0 & 0.883$\pm$0.006 \\
12 & 0.46$\pm$0.03 & 0.422$\pm$0.029 & 1.38$\pm$0.14 & 0.846$\pm$0.003 & 1.57$\pm$0.02 & 0.819$\pm$0.001 & 2.05$\pm$0.01 & 0.9$\pm$0.001 \\
13 & 0$\pm$0 & -- & 1.24$\pm$0.3 & 0.83$\pm$0.017 & 1.38$\pm$0.02 & 0.797$\pm$0.003 & 2.02$\pm$0.06 & 0.894$\pm$0.003 \\
14 & 0.22$\pm$0.35 & 0.39$\pm$0.209 & 1.09$\pm$0.12 & 0.813$\pm$0.015 & 1.46$\pm$0.05 & 0.82$\pm$0.002 & 2$\pm$0.03 & 0.897$\pm$0 \\
15 & -- & -- & 2.29$\pm$0.05 & 0.884$\pm$0.001 & 1.57$\pm$0.07 & 0.825$\pm$0.007 & 1.85$\pm$0.02 & 0.889$\pm$0 \\
\bottomrule
\end{tabular}
}%
\caption{Per-length valid-only diversity metrics of SMILES generation (mean$\pm$95\% CI, $L_{\max}=15$).}
\label{tab:smiles_L15_bylen_valid_div}
\end{table*}

\begin{table*}[!htbp]
\centering
\scriptsize
\renewcommand{\arraystretch}{1.06}
\resizebox{\textwidth}{!}{%
\begin{tabular}{ccccccccccccccccc}
\toprule
$L$ & \multicolumn{4}{c}{TB} & \multicolumn{4}{c}{SubTB} & \multicolumn{4}{c}{RapTB} & \multicolumn{4}{c}{RapTB+SubM} \\
\midrule
 & UniqStr & UniqMol & UniqRateStr & UniqRateMol
 & UniqStr & UniqMol & UniqRateStr & UniqRateMol
 & UniqStr & UniqMol & UniqRateStr & UniqRateMol
 & UniqStr & UniqMol & UniqRateStr & UniqRateMol \\
\midrule
1 & 31$\pm$1.2 & 24.7$\pm$3 & 0$\pm$0 & 0$\pm$0 & 1$\pm$0 & 1$\pm$0 & 0.7$\pm$0.2 & 0.7$\pm$0.2 & 14$\pm$1.4 & 11.3$\pm$1.5 & 0.1$\pm$0 & 0.1$\pm$0 & 5.5$\pm$0.6 & 5.5$\pm$0.6 & 0.1$\pm$0 & 0.1$\pm$0 \\
2 & 189$\pm$3.6 & 128.7$\pm$4.3 & 0.2$\pm$0 & 0.3$\pm$0 & 7.3$\pm$0.4 & 6.3$\pm$0.4 & 0.1$\pm$0 & 0.1$\pm$0 & 35.7$\pm$4.2 & 33$\pm$4.5 & 0.3$\pm$0 & 0.3$\pm$0 & 8.5$\pm$1.7 & 8.5$\pm$1.7 & 0.1$\pm$0 & 0.1$\pm$0 \\
3 & 215.7$\pm$2.7 & 203.7$\pm$1.1 & 0.5$\pm$0 & 0.5$\pm$0 & 14.3$\pm$0.8 & 14.3$\pm$0.8 & 0.5$\pm$0.1 & 0.5$\pm$0.1 & 20.7$\pm$1.5 & 19.7$\pm$1.8 & 0.6$\pm$0 & 0.6$\pm$0 & 8.5$\pm$0.6 & 8.5$\pm$0.6 & 0.2$\pm$0 & 0.2$\pm$0 \\
4 & 170$\pm$9.3 & 164.7$\pm$8.6 & 0.6$\pm$0 & 0.6$\pm$0 & 22$\pm$1.4 & 20.7$\pm$1.8 & 0.3$\pm$0 & 0.3$\pm$0 & 32$\pm$4.5 & 31.3$\pm$4.8 & 0.6$\pm$0.1 & 0.6$\pm$0.1 & 15.5$\pm$1.7 & 15.5$\pm$1.7 & 0.3$\pm$0 & 0.3$\pm$0 \\
5 & 178.7$\pm$6.6 & 176$\pm$6.8 & 0.9$\pm$0 & 0.9$\pm$0 & 47$\pm$1.2 & 42.3$\pm$0.4 & 0.3$\pm$0 & 0.3$\pm$0 & 51.3$\pm$11 & 48.7$\pm$7.7 & 0.9$\pm$0 & 0.8$\pm$0.1 & 23$\pm$0 & 23$\pm$0 & 0.3$\pm$0 & 0.3$\pm$0 \\
6 & 133$\pm$5.4 & 131.7$\pm$5.7 & 1$\pm$0 & 1$\pm$0 & 43.7$\pm$3 & 43$\pm$2.9 & 0.7$\pm$0 & 0.7$\pm$0.1 & 41.3$\pm$5 & 40$\pm$5.7 & 0.7$\pm$0.1 & 0.7$\pm$0.1 & 36.5$\pm$0.6 & 36.5$\pm$0.6 & 0.4$\pm$0 & 0.4$\pm$0 \\
7 & 69.3$\pm$4.8 & 69.3$\pm$4.8 & 1$\pm$0 & 1$\pm$0 & 55.3$\pm$3 & 48$\pm$3.8 & 0.4$\pm$0 & 0.3$\pm$0 & 76.7$\pm$2.3 & 72.3$\pm$3.3 & 0.8$\pm$0 & 0.8$\pm$0 & 45$\pm$7.9 & 44.5$\pm$7.4 & 0.4$\pm$0 & 0.4$\pm$0 \\
8 & 37$\pm$2.9 & 36.7$\pm$2.5 & 1$\pm$0 & 1$\pm$0 & 68$\pm$2.6 & 63.3$\pm$1.7 & 0.6$\pm$0 & 0.6$\pm$0 & 94.7$\pm$3.5 & 93$\pm$2.6 & 1$\pm$0 & 0.9$\pm$0 & 39.5$\pm$4 & 39.5$\pm$4 & 0.4$\pm$0 & 0.4$\pm$0 \\
9 & 40$\pm$3.3 & 40$\pm$3.3 & 0.4$\pm$0 & 0.4$\pm$0 & 55.3$\pm$1.5 & 54.7$\pm$1.7 & 0.8$\pm$0 & 0.8$\pm$0 & 167.3$\pm$4.7 & 157.7$\pm$2.2 & 0.6$\pm$0 & 0.6$\pm$0 & 63$\pm$4.5 & 63$\pm$4.5 & 0.4$\pm$0 & 0.4$\pm$0 \\
10 & 26.3$\pm$1.5 & 26.3$\pm$1.5 & 0.4$\pm$0 & 0.4$\pm$0 & 41.7$\pm$1.5 & 41.3$\pm$1.7 & 0.7$\pm$0 & 0.7$\pm$0 & 358.7$\pm$7.2 & 324.7$\pm$6.5 & 0.7$\pm$0 & 0.7$\pm$0 & 101$\pm$0 & 101$\pm$0 & 0.5$\pm$0 & 0.5$\pm$0 \\
11 & 14.7$\pm$1.1 & 14.7$\pm$1.1 & 0.5$\pm$0.1 & 0.5$\pm$0.1 & 35.7$\pm$3.6 & 35.7$\pm$3.6 & 0.8$\pm$0 & 0.8$\pm$0 & 415.7$\pm$9.9 & 397.3$\pm$8.8 & 0.9$\pm$0 & 0.8$\pm$0 & 139.5$\pm$6.2 & 139.5$\pm$6.2 & 0.5$\pm$0 & 0.5$\pm$0 \\
12 & 5.7$\pm$0.8 & 5.7$\pm$0.8 & 0.6$\pm$0.1 & 0.6$\pm$0.1 & 22.3$\pm$2.3 & 22.3$\pm$2.3 & 1$\pm$0.1 & 1$\pm$0.1 & 391$\pm$10.3 & 360$\pm$10 & 0.8$\pm$0 & 0.7$\pm$0 & 226.5$\pm$4 & 226.5$\pm$4 & 0.6$\pm$0 & 0.6$\pm$0 \\
13 & 1$\pm$0 & 1$\pm$0 & 1$\pm$0 & 1$\pm$0 & 11.3$\pm$2.2 & 11.3$\pm$2.2 & 1$\pm$0 & 1$\pm$0 & 357.7$\pm$12.8 & 328.7$\pm$11.5 & 0.7$\pm$0 & 0.7$\pm$0 & 334.5$\pm$18.7 & 334$\pm$19.2 & 0.7$\pm$0 & 0.7$\pm$0 \\
14 & 2.3$\pm$0.4 & 2.3$\pm$0.4 & 0.7$\pm$0.2 & 0.7$\pm$0.2 & 8.7$\pm$2.3 & 8.7$\pm$2.3 & 1$\pm$0 & 1$\pm$0 & 219.7$\pm$12.6 & 211.3$\pm$12.5 & 0.9$\pm$0 & 0.8$\pm$0 & 303$\pm$9.1 & 261.5$\pm$11.9 & 0.5$\pm$0 & 0.5$\pm$0 \\
15 & -- & -- & -- & -- & 186$\pm$5.7 & 186$\pm$5.7 & 1$\pm$0 & 1$\pm$0 & 67$\pm$2.1 & 66.7$\pm$1.8 & 1$\pm$0 & 1$\pm$0 & 217.5$\pm$11.9 & 217.5$\pm$11.9 & 0.5$\pm$0 & 0.5$\pm$0 \\
\bottomrule
\end{tabular}
}%
\caption{Per-length valid-only uniqueness metrics of SMILES generation (mean$\pm$95\% CI, $L_{\max}=15$).}
\label{tab:smiles_L15_bylen_valid_uniq}
\end{table*}

\subsubsection{Per-depth prefix collapse on long horizon ($L_{\max}=15$).}
\label{app:results_smiles_L15_prefix}
Tables~\ref{tab:smiles_L15_prefix_base}--\ref{tab:smiles_L15_prefix_raptb} report prefix statistics at depth $k$, where concentration metrics (PefEnt/Eff/Top1/UniqueRate) must be interpreted jointly with Survival.
On the long horizon, TB exhibits a textbook collapse pattern: Survival drops from $0.758$ at $k{=}2$ to $0.034$ at $k{=}10$ and is essentially zero for $k\!\ge\!12$, while Top1 peaks at $0.424/0.457$ for $k{=}7/8$, indicating that only a tiny fraction of trajectories reach deep prefixes and those prefixes are highly shared.
RapTB substantially improves deep-prefix survival (e.g., $0.723$ at $k{=}10$) and keeps Top1 low at depth ($\approx0.06$--$0.07$ for $k{=}7$--$10$), consistent with sustained branching and reduced prefix concentration; RapTB+SubM further lowers deep-prefix Top1 while maintaining high Survival, suggesting that improved replay coverage helps prevent the buffer from over-focusing on a few dominant prefixes under long-horizon generation.
At $L_{\max}=15$, we only report RapTB+SubM as the combined setting due to the additional training cost of re-running all baseline+SubM variants at this horizon.
Finally, although SubTB shows strong prefix-level dispersion (low Top1 with high PefEnt/Eff and high Survival), this alone does not imply better terminal quality or validity; thus these prefix statistics should be interpreted together with the per-length terminal metrics in Tables~\ref{tab:smiles_L15_bylen_valid_core}--\ref{tab:smiles_L15_bylen_valid_uniq}, where SubTB degrades on long-length performance.

\begin{table*}[!htbp]
\centering
\scriptsize
\renewcommand{\arraystretch}{1.06}
\resizebox{\textwidth}{!}{%
\begin{tabular}{ccccccccccc}
\toprule
$k$ & \multicolumn{5}{c}{TB} & \multicolumn{5}{c}{SubTB} \\
\midrule
 & Survival & PefEnt & Eff & Top1 & UniqueRate
 & Survival & PefEnt & Eff & Top1 & UniqueRate \\
\midrule
1  & 1$\pm$0 & 2.637$\pm$0.021 & 13.97$\pm$0.3 & 0.236$\pm$0.009 & 0.013$\pm$0
   & 1$\pm$0 & 1.235$\pm$0.016 & 3.44$\pm$0.05 & 0.508$\pm$0.007 & 0.012$\pm$0.001 \\
2  & 0.758$\pm$0.007 & 4.775$\pm$0.027 & 118.61$\pm$3.24 & 0.082$\pm$0.005 & 0.152$\pm$0.008
   & 0.998$\pm$0 & 3.085$\pm$0.009 & 21.87$\pm$0.2 & 0.164$\pm$0.009 & 0.07$\pm$0.003 \\
3  & 0.418$\pm$0.005 & 5.495$\pm$0.048 & 243.8$\pm$11.8 & 0.108$\pm$0.008 & 0.447$\pm$0.01
   & 0.948$\pm$0.002 & 3.987$\pm$0.031 & 53.93$\pm$1.7 & 0.098$\pm$0 & 0.179$\pm$0.004 \\
4  & 0.29$\pm$0.006 & 5.452$\pm$0.052 & 233.74$\pm$12.43 & 0.145$\pm$0.008 & 0.591$\pm$0.016
   & 0.917$\pm$0.002 & 4.466$\pm$0.035 & 87.04$\pm$3 & 0.082$\pm$0.004 & 0.262$\pm$0.003 \\
5  & 0.202$\pm$0.004 & 5.178$\pm$0.092 & 178.4$\pm$16.91 & 0.208$\pm$0.014 & 0.682$\pm$0.02
   & 0.848$\pm$0.007 & 5.041$\pm$0.093 & 155.54$\pm$14.12 & 0.058$\pm$0.008 & 0.375$\pm$0.016 \\
6  & 0.142$\pm$0.006 & 4.373$\pm$0.141 & 80.31$\pm$11.33 & 0.294$\pm$0.02 & 0.607$\pm$0.025
   & 0.714$\pm$0.008 & 5.393$\pm$0.093 & 221.07$\pm$20.27 & 0.066$\pm$0.011 & 0.522$\pm$0.016 \\
7  & 0.099$\pm$0.005 & 3.278$\pm$0.175 & 27.03$\pm$4.53 & 0.424$\pm$0.027 & 0.457$\pm$0.031
   & 0.648$\pm$0.008 & 5.46$\pm$0.069 & 235.93$\pm$16.3 & 0.071$\pm$0.012 & 0.593$\pm$0.011 \\
8  & 0.077$\pm$0.004 & 2.639$\pm$0.112 & 14.11$\pm$1.54 & 0.457$\pm$0.002 & 0.316$\pm$0.031
   & 0.51$\pm$0.005 & 5.702$\pm$0.033 & 299.69$\pm$10.07 & 0.03$\pm$0.006 & 0.727$\pm$0.008 \\
9  & 0.065$\pm$0.003 & 2.835$\pm$0.08 & 17.1$\pm$1.37 & 0.309$\pm$0.03 & 0.28$\pm$0.024
   & 0.403$\pm$0.004 & 5.654$\pm$0.036 & 285.55$\pm$10.24 & 0.038$\pm$0.007 & 0.815$\pm$0.012 \\
10 & 0.034$\pm$0.001 & 2.977$\pm$0.088 & 19.73$\pm$1.72 & 0.154$\pm$0.018 & 0.333$\pm$0.031
   & 0.333$\pm$0.002 & 5.547$\pm$0.039 & 256.65$\pm$9.97 & 0.045$\pm$0.007 & 0.864$\pm$0.013 \\
11 & 0.013$\pm$0.002 & 2.415$\pm$0.052 & 11.21$\pm$0.59 & 0.326$\pm$0.015 & 0.441$\pm$0.031
   & 0.271$\pm$0.004 & 5.51$\pm$0.03 & 247.38$\pm$7.36 & 0.019$\pm$0.001 & 0.933$\pm$0.008 \\
12 & 0.004$\pm$0.001 & 1.737$\pm$0.216 & 5.86$\pm$1.35 & 0.365$\pm$0.045 & 0.517$\pm$0.046
   & 0.23$\pm$0.004 & 5.398$\pm$0.029 & 221.18$\pm$6.45 & 0.021$\pm$0.003 & 0.967$\pm$0.005 \\
13 & 0.001$\pm$0 & 0.347$\pm$0.43 & 1.61$\pm$0.76 & 0.833$\pm$0.207 & 0.428$\pm$0.205
   & 0.207$\pm$0.006 & 5.304$\pm$0.039 & 201.39$\pm$7.93 & 0.024$\pm$0.004 & 0.976$\pm$0.004 \\
14 & 0.001$\pm$0 & 0.745$\pm$0.224 & 2.18$\pm$0.52 & 0.6$\pm$0.172 & 0.689$\pm$0.215
   & 0.195$\pm$0.005 & 5.252$\pm$0.037 & 191.04$\pm$7.2 & 0.024$\pm$0.003 & 0.978$\pm$0.005 \\
15 & 0$\pm$0 & 0$\pm$0 & 0$\pm$0 & 0$\pm$0 & --
   & 0.187$\pm$0.004 & 5.205$\pm$0.035 & 182.38$\pm$6.38 & 0.025$\pm$0.003 & 0.977$\pm$0.005 \\
\bottomrule
\end{tabular}
}%
\caption{Prefix statistics by depth on SMILES generation (mean$\pm$95\% CI, $L_{\max}=15$): TB vs.\ SubTB.}
\label{tab:smiles_L15_prefix_base}
\end{table*}

\begin{table*}[!htbp]
\centering
\scriptsize
\renewcommand{\arraystretch}{1.06}
\resizebox{\textwidth}{!}{%
\begin{tabular}{ccccccccccc}
\toprule
$k$ & \multicolumn{5}{c}{RapTB} & \multicolumn{5}{c}{RapTB+SubM} \\
\midrule
 & Survival & PefEnt & Eff & Top1 & UniqueRate
 & Survival & PefEnt & Eff & Top1 & UniqueRate \\
\midrule
1  & 1$\pm$0 & 1.693$\pm$0.014 & 5.44$\pm$0.08 & 0.375$\pm$0.004 & 0.008$\pm$0
   & 1$\pm$0 & 1.938$\pm$0.001 & 6.94$\pm$0.01 & 0.342$\pm$0 & 0.007$\pm$0 \\
2  & 0.967$\pm$0.002 & 3.438$\pm$0.033 & 31.14$\pm$1.03 & 0.204$\pm$0.006 & 0.051$\pm$0.002
   & 0.982$\pm$0.001 & 3.659$\pm$0.011 & 38.83$\pm$0.43 & 0.125$\pm$0.009 & 0.039$\pm$0 \\
3  & 0.934$\pm$0.001 & 4.542$\pm$0.04 & 93.95$\pm$3.74 & 0.106$\pm$0.006 & 0.138$\pm$0.006
   & 0.963$\pm$0.005 & 4.227$\pm$0.033 & 68.53$\pm$2.26 & 0.126$\pm$0.01 & 0.09$\pm$0.003 \\
4  & 0.923$\pm$0.001 & 5.249$\pm$0.026 & 190.42$\pm$4.86 & 0.071$\pm$0.002 & 0.238$\pm$0.004
   & 0.947$\pm$0.003 & 4.829$\pm$0.009 & 125.11$\pm$1.11 & 0.094$\pm$0.003 & 0.148$\pm$0.002 \\
5  & 0.906$\pm$0.001 & 5.822$\pm$0.019 & 337.73$\pm$6.41 & 0.07$\pm$0.005 & 0.345$\pm$0.003
   & 0.931$\pm$0 & 5.184$\pm$0.001 & 178.44$\pm$0.24 & 0.095$\pm$0.003 & 0.204$\pm$0.002 \\
6  & 0.887$\pm$0.005 & 6.239$\pm$0.013 & 512.44$\pm$6.71 & 0.071$\pm$0.005 & 0.445$\pm$0.004
   & 0.906$\pm$0 & 5.701$\pm$0.007 & 299.31$\pm$2.04 & 0.044$\pm$0.002 & 0.263$\pm$0.002 \\
7  & 0.869$\pm$0.005 & 6.559$\pm$0.021 & 705.68$\pm$14.92 & 0.073$\pm$0.005 & 0.542$\pm$0.002
   & 0.875$\pm$0.001 & 6.07$\pm$0.015 & 432.83$\pm$6.29 & 0.02$\pm$0.001 & 0.322$\pm$0.002 \\
8  & 0.839$\pm$0.004 & 6.78$\pm$0.016 & 879.89$\pm$14.09 & 0.058$\pm$0.004 & 0.614$\pm$0.001
   & 0.841$\pm$0.004 & 6.151$\pm$0.022 & 469.45$\pm$10.24 & 0.018$\pm$0.002 & 0.357$\pm$0.003 \\
9  & 0.808$\pm$0.003 & 6.89$\pm$0.013 & 982.52$\pm$12.91 & 0.053$\pm$0.004 & 0.658$\pm$0.002
   & 0.812$\pm$0.003 & 6.203$\pm$0.001 & 494.29$\pm$0.41 & 0.019$\pm$0.002 & 0.382$\pm$0.004 \\
10 & 0.723$\pm$0.008 & 7.041$\pm$0.005 & 1142.49$\pm$5.93 & 0.024$\pm$0.001 & 0.718$\pm$0.006
   & 0.766$\pm$0.008 & 6.313$\pm$0.017 & 551.78$\pm$9.32 & 0.02$\pm$0.002 & 0.432$\pm$0.001 \\
11 & 0.568$\pm$0.011 & 6.93$\pm$0.016 & 1022.69$\pm$16.25 & 0.024$\pm$0 & 0.769$\pm$0.007
   & 0.701$\pm$0.006 & 6.386$\pm$0.007 & 593.32$\pm$4.26 & 0.022$\pm$0.002 & 0.481$\pm$0.002 \\
12 & 0.414$\pm$0.01 & 6.638$\pm$0.006 & 763.56$\pm$4.76 & 0.03$\pm$0.004 & 0.775$\pm$0.007
   & 0.607$\pm$0.008 & 6.371$\pm$0.018 & 584.42$\pm$10.32 & 0.016$\pm$0.003 & 0.52$\pm$0.004 \\
13 & 0.257$\pm$0.012 & 6.201$\pm$0.01 & 493.35$\pm$5.11 & 0.046$\pm$0.007 & 0.793$\pm$0.009
   & 0.481$\pm$0.003 & 6.205$\pm$0.02 & 495.45$\pm$9.87 & 0.02$\pm$0.004 & 0.548$\pm$0 \\
14 & 0.102$\pm$0.005 & 5.587$\pm$0.029 & 267.14$\pm$7.85 & 0.024$\pm$0.004 & 0.889$\pm$0.021
   & 0.32$\pm$0.001 & 5.677$\pm$0.083 & 292.79$\pm$24.31 & 0.031$\pm$0.005 & 0.511$\pm$0.019 \\
15 & 0.022$\pm$0.001 & 4.195$\pm$0.03 & 66.4$\pm$2.02 & 0.029$\pm$0.001 & 0.976$\pm$0.005
   & 0.129$\pm$0.003 & 4.899$\pm$0.098 & 134.67$\pm$13.23 & 0.077$\pm$0.015 & 0.543$\pm$0.016 \\
\bottomrule
\end{tabular}
}%
\caption{Prefix statistics by depth on SMILES generation (mean$\pm$95\% CI, $L_{\max}=15$): RapTB vs.\ RapTB+SubM.}
\label{tab:smiles_L15_prefix_raptb}
\end{table*}

\subsection{Expr24}
\label{app:results_expr24}
We provide results with 95\% CI and per-length termination probability analysis of different objectives under various replay strategies.

\begin{table*}[t]
\centering
\small
\setlength{\tabcolsep}{5pt}
\renewcommand{\arraystretch}{1.08}
\caption{\textbf{Expr24 results under four replay schemes.} All the experiments are run under 3 different random seeds with 95\% CI. Per-run sample size is 6400.}
\label{tab:expr24_main_ci}
\begin{tabular}{llcccccc}
\toprule
Replay & Objective & Unique$_\checkmark$ & NormCov & Acc & KL($\pi\!\to\!p^*$) & KL($p^*\!\to\!\pi$) & JS$_{\text{tok}}$ \\
\midrule
\multirow{3}{*}{PRT} & TB & 103.7$\pm$3.2 & 0.016$\pm$0.001 & 0.999$\pm$0.000 & 1.105$\pm$0.002 & 7.803$\pm$0.060 & 0.292$\pm$0.001 \\
 & SubTB & 292.0$\pm$2.9 & 0.046$\pm$0.000 & 0.311$\pm$0.002 & 0.424$\pm$0.010 & 0.672$\pm$0.077 & 0.107$\pm$0.003 \\
 & RapTB & 129.3$\pm$0.4 & 0.020$\pm$0.000 & 0.992$\pm$0.001 & 0.908$\pm$0.003 & 5.538$\pm$0.005 & 0.230$\pm$0.001 \\
\midrule
\multirow{3}{*}{RP} & TB & 5.3$\pm$0.4 & 0.001$\pm$0.000 & 1.000$\pm$0.000 & 1.297$\pm$0.001 & 11.403$\pm$0.282 & 0.339$\pm$0.000 \\
 & SubTB & 324.7$\pm$2.7 & 0.051$\pm$0.000 & 0.229$\pm$0.005 & 0.455$\pm$0.005 & 0.865$\pm$0.083 & 0.109$\pm$0.002 \\
 & RapTB & 246.7$\pm$7.1 & 0.039$\pm$0.001 & 0.991$\pm$0.000 & 0.561$\pm$0.001 & 4.480$\pm$0.002 & 0.147$\pm$0.000 \\
\midrule
\multirow{3}{*}{SubM} & TB & 642.0$\pm$5.6 & 0.100$\pm$0.001 & 0.996$\pm$0.001 & 0.182$\pm$0.001 & 0.441$\pm$0.005 & 0.049$\pm$0.000 \\
 & SubTB & 331.3$\pm$22.7 & 0.052$\pm$0.004 & 0.061$\pm$0.005 & 0.149$\pm$0.008 & 0.286$\pm$0.070 & 0.040$\pm$0.002 \\
 & RapTB & 1337.3$\pm$7.5 & 0.209$\pm$0.001 & 0.994$\pm$0.001 & 0.169$\pm$0.001 & 0.623$\pm$0.004 & 0.048$\pm$0.000 \\
\midrule
\multirow{3}{*}{Oracle} & TB & 5198.0$\pm$5.2 & 0.812$\pm$0.001 & 0.919$\pm$0.001 & 0.062$\pm$0.001 & 0.066$\pm$0.001 & 0.016$\pm$0.000 \\
 & SubTB & 35.7$\pm$2.9 & 0.006$\pm$0.000 & 0.006$\pm$0.000 & 0.266$\pm$0.009 & 1.491$\pm$0.413 & 0.071$\pm$0.003 \\
 & RapTB & 5220.7$\pm$4.3 & 0.816$\pm$0.001 & 0.945$\pm$0.001 & 0.052$\pm$0.001 & 0.056$\pm$0.001 & 0.013$\pm$0.000 \\
\bottomrule
\end{tabular}
\end{table*}

\begin{table}[!htbp]
\centering
\small
\setlength{\tabcolsep}{5pt}
\renewcommand{\arraystretch}{1.05}
\begin{tabular}{llcccc}
\toprule
Replay & Objective & $\ell=3$ & $\ell=5$ & $\ell=7$ & $\ell=9$ \\
\midrule
\multirow{4}{*}{RP} & TB            & --     & --     & -0.000 & -0.001 \\
                       & SubTB         & -5.220 & -1.292 & -1.064 & -79.638 \\
                       & RapTB         & --     & -2.451 & -2.319 & -0.065 \\
                       & RootSubTBLogZ & -0.709 & -0.606 & -0.411 & -0.068 \\
\midrule
\multirow{4}{*}{Oracle} & TB            & --     & -4.391 & -1.779 & -0.441 \\
                        & SubTB         & -1.017 & -2.803 & -4.530 & -86.415 \\
                        & RapTB         & --     & -8.312 & -3.341 & -0.644 \\
                        & RootSubTBLogZ & -0.442 & -0.417 & -0.354 & -1.432 \\
\bottomrule
\end{tabular}
\caption{Per-length $\log p_{\text{term}}$ on Expr24.}
\label{tab:expr24_pterm_by_len}
\end{table}

\subsection{CommonGen.}
\label{app:results_commongen}
To make the behavioral differences visible, we show a few representative validation probes where the contrast is the clearest.
Each row corresponds to the \emph{same} probe instance across methods (IDs omitted for compactness; see released CSV logs for exact indices).
We report the effective length (Len) before padding.

\begin{table*}[t]
\centering
\footnotesize
\setlength{\tabcolsep}{5pt}
\renewcommand{\arraystretch}{1.18}
\caption{\textbf{Samples from CommonGen task.}
SubTB often exhibits max-length / off-distribution artifacts, while RapTB+SubM yields shorter and cleaner sentences.
}
\label{tab:commongen_qual}
\vspace{-0.4em}
\begin{tabular}{p{0.24\textwidth} p{0.24\textwidth} p{0.24\textwidth} p{0.24\textwidth}}
\toprule
TB & SubTB & RapTB & RapTB+SubM \\
\midrule

{\scriptsize (Len=19)}\newline
The cat likes to sit on the couch with a pet that lives in a cat, because it.
&
{\scriptsize (Len=20)}\newline
My pet cat likes to sleep on a couch, I know, what a pet perv! This is.
&
{\scriptsize (Len=15)}\newline
The pet cat loves the couch, and the cat's mom was an mascot.
&
{\scriptsize (Len=10)}\newline
A pet cat likes to sleep on a couch.
\\ \addlinespace

{\scriptsize (Len=15)}\newline
A kid loves to dance in a room that a kid likes to dance in.
&
{\scriptsize (Len=20)}\newline
My kid loves to dance in her room with all of her clothes on except her persnare cal.
&
{\scriptsize (Len=7)}\newline
I dance in the room for kid.
&
{\scriptsize (Len=8)}\newline
kid loves to dance in her room!!
\\ \addlinespace

{\scriptsize (Len=20)}\newline
There is a hockey team that plays at a rink and plays on a team on a hockey Claus.
&
{\scriptsize (Len=20)}\newline
The team practiced hockey at the rink, and do freeze to ourselves towards the rinks touring doors.
&
{\scriptsize (Len=15)}\newline
A team played at the hockey rink with the best players on their own.
&
{\scriptsize (Len=11)}\newline
The team practiced hockey at the rink for four hours.
\\ \addlinespace

{\scriptsize (Len=20)}\newline
There was a flight of stairs, and a jump, a flight of stairs, and a jump,
&
{\scriptsize (Len=20)}\newline
I jump up a flight of stair, landing in a little flight-recorded levy down to current it.
&
{\scriptsize (Len=14)}\newline
I jump up a flight of stair to get to school in the morning.
&
{\scriptsize (Len=11)}\newline
I walk up a stair: jump up a flight.
\\

\bottomrule
\end{tabular}
\vspace{-0.8em}
\end{table*}

\section{Additional Baselines}
\label{app:additional_baselines}

\subsection{RL Baselines: PPO and GRPO}
\label{app:rl_baselines}

To contextualize RapTB within the broader LLM fine-tuning landscape, we compare against PPO and GRPO~\citep{shao2024grpo}.
Both methods are trained with the same backbone (Llama-3.2-1B, LoRA rank 16) and reward function.
As shown in the main-text Tables~\ref{tab:smiles_main} and~\ref{tab:expr24_main}, reward-maximizing RL methods achieve reasonable task scores but suffer severe diversity collapse: PPO collapses to a single mode on both tasks, and GRPO achieves Entropy $\le 0.98$ on SMILES.
This is consistent with the fundamental difference between reward maximization and reward-proportional sampling~\citep{hu2024amortizing}: RL concentrates mass on the single best mode, whereas GFlowNets target the full reward-proportional distribution.

\subsection{AvgPrefixTB: Uniform Prefix Trajectory Balance}
\label{app:avgprefixtb}

\paragraph{Definition.}
In a terminable prefix tree, every prefix $s_{0:k}$ can be regarded as a complete trajectory by appending a stop token $\top$.
AvgPrefixTB averages the standard TB residual over all such prefix terminations along a sampled trajectory:
\begin{equation}
\label{eq:avgprefixtb}
\mathcal{L}_{\mathrm{AvgPrefixTB}}(\xi)
\triangleq
\frac{1}{\tau}\sum_{k=1}^{\tau}\big(\Delta_k^{\mathrm{TB}}(\xi)\big)^2,
\end{equation}
where $\Delta_k^{\mathrm{TB}}(\xi)$ is the TB residual at prefix $s_{0:k}$ (Eq.~\ref{eq:app_stop_res_k_new}).
This provides $O(N)$ constraints like RapTB, but differs in two key respects:
(i) each residual retains the learnable $\log Z_\theta$ instead of canceling it via rooting, and
(ii) it uses the raw stop-reward $\log R(s_{0:k}^{\top})$ at each prefix rather than absorbed suffix targets.

\paragraph{Motivation as a baseline.}
AvgPrefixTB tests whether simply densifying TB supervision across all prefixes---without the structural design choices of RapTB (rooted residuals, absorbed targets, termination gradient stopping)---is sufficient to address prefix collapse and length bias.

\paragraph{Results.}
Table~\ref{tab:avgprefixtb_full} compares AvgPrefixTB against TB and RapTB on both SMILES and Expr24.

\begin{table}[!htbp]
\centering
\small
\setlength{\tabcolsep}{4pt}
\renewcommand{\arraystretch}{1.08}
\caption{\textbf{AvgPrefixTB comparison on SMILES and Expr24.}
All methods use the same backbone (Llama-3.2-1B, LoRA rank 16) and RP replay.
SMILES metrics are computed on valid samples; Len denotes mean token length ($L_{\max}=10$).}
\label{tab:avgprefixtb_full}
\vspace{-0.35em}
\begin{tabular}{@{}lcccccc@{}}
\toprule
& \multicolumn{4}{c}{\textbf{SMILES}} & \multicolumn{2}{c}{\textbf{Expr24}} \\
\cmidrule(lr){2-5} \cmidrule(lr){6-7}
Method & Acc$\uparrow$ & Score$\uparrow$ & Entropy$\uparrow$ & Len & Acc$\uparrow$ & Unique$_\checkmark$$\uparrow$ \\
\midrule
TB          & 0.998 & 0.717 & 2.503 & 3.06 & 1.000 & 5.3 \\
AvgPrefixTB & 1.000 & 0.661 & 0.665 & 2.89 & 0.998 & 142.0 \\
RapTB       & 0.996 & 0.740 & 2.448 & 6.14 & 0.991 & 246.7 \\
RapTB+SubM  & 0.988 & 0.844 & 2.726 & 7.44 & 0.994 & 1337.3 \\
\bottomrule
\end{tabular}
\end{table}

\paragraph{Analysis.}
AvgPrefixTB exhibits a pronounced short-sequence bias on SMILES: average length is $2.89$ with $54\%$ of generated mass concentrated on lengths $1$--$2$.
This is because averaging TB residuals uniformly across all prefixes creates a \emph{shortcut}: early prefixes already have near-zero residuals (they are close to the root and have few accumulated transitions), so the model can minimize the average loss cheaply by terminating early and concentrating probability on short, high-reward trajectories.
Score ($0.661$) and Entropy ($0.665$) are both substantially below TB ($0.717$ / $2.503$) and RapTB ($0.740$ / $2.448$).

On Expr24, AvgPrefixTB improves unique correct solutions over TB ($142.0$ vs.\ $5.3$), indicating that prefix-level supervision does help with mode diversity even in its simplest form.
However, it remains well below RapTB ($246.7$; NormCov $0.016$ vs.\ $0.039$), confirming that the specific design choices of RapTB---rooted residuals that cancel $\log Z_\theta$, absorbed suffix targets for informative prefix credit, and termination gradient stopping---are individually and jointly important.
The ablation in Table~\ref{tab:ablation} (main text) provides further evidence for each component.

\subsection{AMP Biological Sequence Generation}
\label{app:amp}

We evaluate on the antimicrobial peptide (AMP) generation task~\citep{jain2022biological}, a standard GFlowNet benchmark with amino-acid vocabulary and non-differentiable reward.
We additionally compare against DynaPPO, COMs, and a standard GFlowNet baseline from~\citet{jain2022biological}.

\begin{table}[!htbp]
\centering
\small
\setlength{\tabcolsep}{3.5pt}
\renewcommand{\arraystretch}{1.08}
\caption{\textbf{AMP biological sequence generation.}
$\dagger$: SubTB collapses to max length; its diversity/novelty is inflated by raw edit distance over unnaturally long sequences.}
\label{tab:amp_results}
\begin{tabular}{@{}lccccc@{}}
\toprule
Method & Perf$\uparrow$ & Div$\uparrow$ & Novelty$\uparrow$ & Len & Steps \\
\midrule
DynaPPO    & 0.938 & 12.12 & 9.31  & $\sim$20 & 10K \\
COMs       & 0.761 & 19.38 & 26.47 & $\sim$20 & 10K \\
GFlowNet   & 0.868 & 11.32 & 15.72 & $\sim$20 & 10K \\
\midrule
TB         & 0.927 & 7.39  & 10.65 & 17.4 & 10K \\
SubTB$^\dagger$ & 0.897 & 21.37 & 28.68 & 49.3 & 10K \\
RapTB      & 0.919 & 8.83  & 14.44 & 22.4 & 5K \\
RapTB+SubM & \textbf{0.916} & \textbf{16.92} & \textbf{15.77} & 25.6 & 3K \\
\bottomrule
\end{tabular}
\end{table}

RapTB+SubM achieves the best performance--diversity--novelty trade-off within only 3K training steps.
SubTB exhibits the same termination drift observed in SMILES and Expr24, collapsing to maximum length ($49.3$) and inflating diversity/novelty through unnaturally long sequences rather than genuine structural variation.
This confirms that the termination drift failure mode generalizes to biological sequence generation with a fundamentally different vocabulary and reward structure.

\section{Scaling Study}
\label{app:scaling}

To verify that the identified failure modes are structural rather than capacity-limited, we scale from Llama-3.2-1B to 3B, 8B (Llama-3.2), and 32B (Qwen3) on SMILES.
All runs use LoRA (rank 16) and the same reward/decoding configuration as the 1B experiments.

\begin{table}[!htbp]
\centering
\small
\setlength{\tabcolsep}{3pt}
\renewcommand{\arraystretch}{1.1}
\caption{\textbf{Scaling study on SMILES generation across model sizes and architectures.}}
\label{tab:scaling}
\begin{tabular}{@{}clccccc@{}}
\toprule
Scale & Method & Acc$\uparrow$ & Score$\uparrow$ & FPDiv$\uparrow$ & Ent$\uparrow$ & Len \\
\midrule
\multirow{4}{*}{3B (Llama)}
& TB         & 0.999 & 0.716 & 0.838 & 1.92 & 2.69 \\
& SubTB      & 0.311 & 0.222 & 0.854 & 2.56 & 9.52 \\
& RapTB      & 1.000 & 0.795 & 0.839 & 1.81 & 7.99 \\
& RapTB+SubM & 0.998 & 0.869 & 0.936 & 2.41 & 8.05 \\
\midrule
\multirow{4}{*}{8B (Llama)}
& TB         & 1.000 & 0.715 & 0.775 & 1.84 & 2.98 \\
& SubTB      & 0.391 & 0.307 & 0.869 & 2.72 & 9.22 \\
& RapTB      & 0.999 & 0.825 & 0.852 & 1.89 & 8.09 \\
& RapTB+SubM & 0.998 & 0.873 & 0.937 & 2.51 & 7.65 \\
\midrule
\multirow{4}{*}{32B (Qwen3)}
& TB         & 1.000 & 0.762 & 0.860 & 2.06 & 3.15 \\
& SubTB      & 0.795 & 0.626 & 0.864 & 2.19 & 7.31 \\
& RapTB      & 0.998 & 0.794 & 0.880 & 2.16 & 6.19 \\
& RapTB+SubM & 0.998 & 0.867 & 0.896 & 2.23 & 7.56 \\
\bottomrule
\end{tabular}
\end{table}

Three observations emerge:

(a) \textbf{SubTB termination drift persists across all scales.}
Accuracy remains low at every size ($0.311$/$0.391$/$0.795$ at 3B/8B/32B), and even 32B cannot fully resolve the issue.
This confirms the failure is structural (arising from the overlapping-window objective) rather than capacity-limited.

(b) \textbf{RapTB scales reliably.}
Accuracy stays above $0.998$ at all scales, with diversity improving with model size (FPDiv: $0.839 \to 0.852 \to 0.880$).

(c) \textbf{RapTB+SubM achieves the best quality--diversity trade-off at every scale.}
At 32B: highest QED ($0.867$), strong diversity (FPDiv $0.896$), and near-perfect validity (Acc $0.998$).
These results across two architecture families confirm that the benefits of RapTB+SubM are architecture-independent.

\section{Hyperparameter Sensitivity}
\label{app:hparam_sensitivity}

RapTB introduces several hyperparameters beyond TB: auxiliary weight $\eta$, soft-backup temperature $\beta$, distance penalty $\rho$, mix weight $\alpha$, distance discount $\gamma$, minimum prefix depth $k_{\min}$ (with schedule), and horizon cap $K$.
To assess robustness, we conduct cross-task sweeps over $(\beta,\rho)$ (18 configs total) plus separate $\eta$ and $k_{\min}$ ablations.
Across all settings, no catastrophic failure, length collapse, or termination drift is observed.

\subsection{Expr24}

\paragraph{$(\beta,\rho)$ grid (9 experiments).}
$\beta$ and $\rho$ govern the bias--variance trade-off of the absorbed target: $\beta$ controls the smoothness of the soft backup, and $\rho$ controls the distance-decay rate.

\begin{table}[!htbp]
\centering
\small
\setlength{\tabcolsep}{5pt}
\renewcommand{\arraystretch}{1.08}
\caption{$(\beta,\rho)$ sensitivity on Expr24 (RP replay).}
\label{tab:hparam_expr24_beta_rho}
\begin{tabular}{@{}cccc@{}}
\toprule
$\beta$ & $\rho$ & Acc$\uparrow$ & Diversity$\uparrow$ \\
\midrule
1 & 0.0 & 0.999 & 1.010 \\
1 & 0.1 & 0.999 & 0.993 \\
1 & 0.5 & 1.000 & 1.015 \\
3 & 0.0 & 0.990 & 0.950 \\
3 & 0.1 & 1.000 & 0.769 \\
3 & 0.5 & 0.997 & 0.978 \\
5 & 0.0 & 0.983 & 1.022 \\
5 & 0.1 & 0.998 & 0.968 \\
5 & 0.5 & 0.999 & 0.912 \\
\bottomrule
\end{tabular}
\end{table}

All 9 configs maintain Acc $\ge 0.983$.
The parameters primarily affect the accuracy--diversity balance rather than causing qualitative failures.

\paragraph{$\eta$ sweep ($\beta{=}3,\rho{=}0.5$).}
Higher $\eta$ strengthens the auxiliary prefix credit signal, improving diversity at a mild cost to accuracy.

\begin{table}[!htbp]
\centering
\small
\setlength{\tabcolsep}{5pt}
\renewcommand{\arraystretch}{1.08}
\caption{Auxiliary weight $\eta$ sensitivity on Expr24.}
\label{tab:hparam_expr24_eta}
\begin{tabular}{@{}ccc@{}}
\toprule
$\eta$ & Acc$\uparrow$ & Diversity$\uparrow$ \\
\midrule
0.10 & 0.999 & 0.983 \\
0.25 & 0.997 & 0.978 \\
0.50 & 0.987 & 1.149 \\
\bottomrule
\end{tabular}
\end{table}

\paragraph{$k_{\min}$ ablation ($\beta{=}3,\rho{=}0.5$).}
$k_{\min}$ controls the minimum prefix depth receiving auxiliary supervision.
Smaller $k_{\min}$ emphasizes shorter prefixes, creating a shortcut toward high-reward short sequences that improves accuracy but reduces diversity.
The linear schedule ($7{\to}3$) used in the paper provides the best balance.

\begin{table}[!htbp]
\centering
\small
\setlength{\tabcolsep}{5pt}
\renewcommand{\arraystretch}{1.08}
\caption{$k_{\min}$ schedule sensitivity on Expr24.}
\label{tab:hparam_expr24_kmin}
\begin{tabular}{@{}lcc@{}}
\toprule
$k_{\min}$ variant & Acc$\uparrow$ & Diversity$\uparrow$ \\
\midrule
Fixed $k_{\min}{=}3$ & 0.998 & 0.852 \\
Schedule $7{\to}3$ & 0.997 & 0.978 \\
Fixed $k_{\min}{=}7$ & 0.969 & 1.025 \\
\bottomrule
\end{tabular}
\end{table}

\subsection{SMILES}

\paragraph{$(\beta,\rho)$ grid (9 experiments).}
Eight of nine configs achieve Acc $\ge 0.991$; only $(\beta{=}10,\rho{=}0)$ shows mild degradation ($0.968$), corresponding to high temperature with zero distance penalty.
No config exhibits length collapse: all average lengths fall in $[5.5,7.5]$, far from TB's collapsed $3.06$.
Score ($\ge 0.72$) and FPDiv ($\ge 0.83$) remain robust across all settings.

\begin{table}[!htbp]
\centering
\small
\setlength{\tabcolsep}{5pt}
\renewcommand{\arraystretch}{1.08}
\caption{$(\beta,\rho)$ sensitivity on SMILES (RP replay).}
\label{tab:hparam_smiles_beta_rho}
\begin{tabular}{@{}cccc@{}}
\toprule
$\beta$ & $\rho$ & Acc$\uparrow$ & Entropy$\uparrow$ \\
\midrule
1  & 0.0 & 0.992 & 2.173 \\
1  & 0.1 & 0.994 & 2.161 \\
1  & 0.5 & 0.992 & 2.162 \\
5  & 0.0 & 0.995 & 1.997 \\
5  & 0.1 & 0.991 & 2.076 \\
5  & 0.5 & 0.997 & 2.079 \\
10 & 0.0 & 0.968 & 2.279 \\
10 & 0.1 & 0.999 & 1.986 \\
10 & 0.5 & 0.997 & 2.036 \\
\bottomrule
\end{tabular}
\end{table}

\section{Metrics: Formal Definitions and Protocol}
\label{app:metrics}
\newcommand{\valid}{\mathrm{valid}}

\paragraph{Sampling and aggregation protocol.}
For each run, we draw $N$ i.i.d. terminal samples $\{x_i\}_{i=1}^N$ from the learned sampler (For SMILES, $N=3200$, For Expr24, $N=6400$).
Let $\mathbb{I}_{\valid}(x)\in\{0,1\}$ indicate whether $x$ satisfies task constraints.
Let $\mathcal{D}$ denote the multiset of all samples and let
\[
\mathcal{D}_{\valid}\triangleq \{x_i\in\mathcal{D}:\mathbb{I}_{\valid}(x_i)=1\},\qquad
n_{\valid}\triangleq |\mathcal{D}_{\valid}|.
\]
Unless explicitly stated otherwise, all metrics \emph{except \texttt{Acc}} are computed on $\mathcal{D}_{\valid}$.
We report a metric as $0$ if its denominator is $0$ (e.g., $n_{\valid}=0$).

Across random seeds, we report the mean and a two-sided 95\% confidence interval.
With $S$ seeds and per-seed values $m_1,\dots,m_S$, we report
\[
\bar m \pm t_{S-1,0.975}\,\frac{\mathrm{sd}(m_1,\dots,m_S)}{\sqrt{S}}.
\]

\subsection{Terminal-level metrics}
\label{app:metrics_terminal}

\paragraph{Accuracy / validity rate (\texttt{Acc}).}
\texttt{Acc} measures the fraction of valid samples among all $N$ draws:
\begin{equation}
\texttt{Acc}\ \triangleq\ \frac{1}{N}\sum_{i=1}^{N}\mathbb{I}_{\valid}(x_i).
\end{equation}
For SMILES, $\mathbb{I}_{\valid}(x)=1$ requires chemical validity and scaffold consistency.
For Expr24, $\mathbb{I}_{\valid}(x)=1$ requires syntactic validity and $\mathrm{eval}(x)=24$.

\paragraph{Task score on valid samples (\texttt{Score}).}
Let $s(x)$ be the task score.
\begin{equation}
\texttt{Score}\ \triangleq\ \frac{1}{n_{\valid}}\sum_{x\in\mathcal{D}_{\valid}} s(x).
\end{equation}
For Expr24 with binary reward, $s(x)=\mathbb{I}_{\valid}(x)$, so \texttt{Score} numerically equals \texttt{Acc}.

\paragraph{Pre-EOS length and length statistics.}
Each terminal $x$ is a variable-length token sequence with pre-EOS length $\ell(x)$ (number of tokens before EOS/\,$\top$).
We report
\[
\texttt{Len}\ \triangleq\ \frac{1}{n_{\valid}}\sum_{x\in\mathcal{D}_{\valid}}\ell(x),\qquad
\texttt{Len}_{50},\ \texttt{Len}_{90}\ \text{as percentiles of }\{\ell(x):x\in\mathcal{D}_{\valid}\}.
\]

\paragraph{Length-bin fractions and counts.}
Given an integer bin $[a,b]$ (inclusive), define
\[
\texttt{Frac}[a\text{--}b]\ \triangleq\ \frac{1}{n_{\valid}}\sum_{x\in\mathcal{D}_{\valid}}\mathbb{I}[a\le \ell(x)\le b],\qquad
\texttt{Count}[a\text{--}b]\ \triangleq\ \sum_{x\in\mathcal{D}_{\valid}}\mathbb{I}[a\le \ell(x)\le b].
\]
For an open-ended bin $[a,+\infty)$, replace the indicator with $\mathbb{I}[\ell(x)\ge a]$.

\paragraph{Termination calibration (\texorpdfstring{$\log p_{\mathrm{term}}(\tau)$}{log p_term(tau)}).}
For each sampled trajectory, let $\tau_i$ be the sampled stop step (the position where EOS/\,$\top$ is taken).
Define the per-sample termination log-probability
\[
\log p_{\mathrm{term}}(\tau_i)\ \triangleq\ \log q_\theta(\top\mid s_{0:\tau_i}),
\]
evaluated from the model's raw termination head (no masking/renormalization).
We report the mean over all samples:
\[
\log p_{\mathrm{term}}(\tau)\ \triangleq\ \frac{1}{N}\sum_{i=1}^{N}\log p_{\mathrm{term}}(\tau_i).
\]
More negative values indicate overly suppressed termination.

\paragraph{Uniqueness metrics for SMILES.}
Let $\mathrm{canon}(x)$ denote canonicalization used in evaluation (e.g., canonical SMILES / canonical molecule identity).
We define
\[
\texttt{UniqStr}\triangleq |\{x: x\in\mathcal{D}_{\valid}\}|,\qquad
\texttt{UniqRateStr}\triangleq \texttt{UniqStr}/n_{\valid}.
\]
For molecule-level uniqueness, define $\texttt{UniqMol}\triangleq |\{\mathrm{canon}(x): x\in\mathcal{D}_{\valid}\}|$ and
$\texttt{UniqRateMol}\triangleq \texttt{UniqMol}/n_{\valid}$.

\subsection{Token entropy}
\label{app:metrics_Entropy}

\paragraph{Ragged token entropy (\texttt{Entropy}).}
Let $\mathcal{D}_{\valid}=\{x_i\}_{i=1}^{n_{\valid}}$ and let $\ell_i\triangleq\ell(x_i)$.
For each position $t\ge 1$, consider the survivor index set $\mathcal{I}_t\triangleq\{i:\ell_i\ge t\}$ with $n_t\triangleq|\mathcal{I}_t|$.
If $n_t\le 1$, skip this position. Otherwise define the empirical marginal
\[
\hat p_t(v)\triangleq \frac{1}{n_t}\sum_{i\in\mathcal{I}_t}\mathbb{I}[x_{i,t}=v]
\]
and its entropy (natural log)
\[
H_t\triangleq-\sum_v\hat p_t(v)\log(\hat p_t(v)+\epsilon),\qquad \epsilon=10^{-10}.
\]
Let $\mathcal{T}\triangleq\{t:n_t>1\}$ and report
\[
\texttt{Entropy}\triangleq \frac{1}{|\mathcal{T}|}\sum_{t\in\mathcal{T}} H_t.
\]

\paragraph{Length-bucketed token entropy (\texttt{Entropy}$(\ell)$).}
Group valid samples by their pre-EOS length $\ell$ and compute \texttt{Entropy} on each fixed-length bucket after truncation to $\ell$.
We set $\texttt{Entropy}(\ell)=0$ if the bucket has $\le 1$ sample or $\ell\le 0$.

\paragraph{Fingerprint diversity (\texttt{FPDiv}) for SMILES.}
Let $f(x)$ be a fingerprint and $\mathrm{sim}(x,x')$ a similarity (Tanimoto in our SMILES experiments).
We report
\begin{equation}
\texttt{FPDiv}\ \triangleq\ 1-\frac{2}{n_{\valid}(n_{\valid}-1)}\sum_{\substack{x,x'\in\mathcal{D}_{\valid}\\ x<x'}} \mathrm{sim}(x,x').
\end{equation}

\paragraph{Macro-averaged fingerprint diversity (\texttt{MacroFP}).}
Given length bins $\mathcal{B}$ (e.g., 0--5, 6--10, 11+), let $\mathcal{D}_{\valid}^{(b)}$ be the valid subset in bin $b$.
Define
\[
\texttt{MacroFP}\triangleq \frac{1}{|\mathcal{B}|}\sum_{b\in\mathcal{B}} \texttt{FPDiv}(\mathcal{D}_{\valid}^{(b)}),
\]
where $\texttt{FPDiv}(\cdot)=0$ if a bin contains fewer than 2 samples.

\subsection{Prefix-collapse metrics}
\label{app:metrics_prefix}

\paragraph{Prefixes and survivors.}
For a terminal $x=(x_1,\dots,x_{\ell(x)})$, define its length-$k$ prefix
$s_{0:k}(x)\triangleq(x_1,\dots,x_k)$ for $k\le \ell(x)$.
At depth $k$, the valid survivor multiset is
\[
\mathcal{D}_{\valid,k}\triangleq\{x\in\mathcal{D}_{\valid}:\ell(x)\ge k\},\qquad n_k\triangleq|\mathcal{D}_{\valid,k}|.
\]

\paragraph{Prefix survival (\texttt{Surv}$(k)$).}
\[
\texttt{Surv}(k)\triangleq \frac{n_k}{n_{\valid}}.
\]

\paragraph{Prefix entropy (\texttt{PefEnt}$(k)$) and effective prefix count (\texttt{Eff}$(k)$).}
Let $\hat p_k(p)$ be the empirical frequency of prefix $p$ among survivors:
\[
\hat p_k(p)\triangleq \frac{1}{n_k}\sum_{x\in\mathcal{D}_{\valid,k}}\mathbb{I}[s_{0:k}(x)=p].
\]
Define
\[
\texttt{PefEnt}(k)\triangleq -\sum_p\hat p_k(p)\log \hat p_k(p),\qquad
\texttt{Eff}(k)\triangleq \exp(\texttt{PefEnt}(k)).
\]

\paragraph{Top-1 prefix mass (\texttt{Top1}$(k)$).}
\[
\texttt{Top1}(k)\triangleq \max_p\hat p_k(p).
\]

\paragraph{Unique prefix rate (\texttt{UniqueRate}$(k)$).}
\[
\texttt{UniqueRate}(k)\triangleq \frac{|\{s_{0:k}(x):x\in\mathcal{D}_{\valid,k}\}|}{n_k}.
\]

\subsection{Distribution metrics for Expr24 (oracle reference)}
\label{app:metrics_expr24_dist}

For Expr24, valid means correct.
Let $\mathcal{U}_{\valid}\triangleq\{\mathrm{tuple}(x):x\in\mathcal{D}_{\valid}\}$ be the set of unique valid sequences.
Let the enumerated oracle set be $\mathcal{Y}^\star$.
We report
\[
\texttt{Unique}_{\checkmark}\triangleq |\mathcal{U}_{\valid}|,\qquad
\texttt{CovCount}\triangleq |\mathcal{U}_{\valid}\cap\mathcal{Y}^\star|,\qquad
\texttt{Cov}\triangleq \texttt{CovCount}/|\mathcal{Y}^\star|,
\]
and the sampling-cap-normalized variant
\[
\texttt{NormCov}\triangleq \texttt{CovCount}/\min(N,|\mathcal{Y}^\star|).
\]

\paragraph{Position-wise token marginals.}
Given a multiset of sequences $\mathcal{S}$ (oracle or sampled), define survivors at 0-indexed position $t$:
$\mathcal{S}_t\triangleq\{x\in\mathcal{S}:|x|>t\}$.
Let $C_t(v)$ be the count of token $v$ at position $t$ among $\mathcal{S}_t$ and $Z_t\triangleq\sum_v C_t(v)$.
If $Z_t>0$, define the empirical marginal $q_t(v)\triangleq C_t(v)/Z_t$.

\paragraph{Stabilized KL/JS.}
Let $\epsilon=10^{-9}$ and let $\mathcal{A}_t$ be the union support of oracle and sampled marginals at position $t$.
Define
\[
\mathrm{KL}_\epsilon(p\|q)\triangleq\sum_{v\in\mathcal{A}_t}(p(v)+\epsilon)\log\frac{p(v)+\epsilon}{q(v)+\epsilon},
\]
and $\mathrm{JS}_\epsilon(p,q)\triangleq\tfrac12\mathrm{KL}_\epsilon(p\|m)+\tfrac12\mathrm{KL}_\epsilon(q\|m)$ where $m=\tfrac12(p+q)$.

\paragraph{Scalar divergences.}
Let $p^\star_t$ be the oracle marginal at position $t$ computed from $\mathcal{Y}^\star$, and $\pi_t$ the sampled marginal at $t$ computed from $\mathcal{D}_{\valid}$ (duplicates kept).
Let $T_{\max}$ be the maximum length across oracle and sampled sequences.
We report
\[
\mathrm{KL}(\pi\!\to\!p^\star)\triangleq\frac{1}{T_{\max}}\sum_{t=0}^{T_{\max}-1}\mathrm{KL}_\epsilon(\pi_t\|p^\star_t),\quad
\mathrm{KL}(p^\star\!\to\!\pi)\triangleq\frac{1}{T_{\max}}\sum_{t=0}^{T_{\max}-1}\mathrm{KL}_\epsilon(p^\star_t\|\pi_t),
\]
\[
\texttt{JS}_{\text{tok}}\triangleq\frac{1}{T_{\max}}\sum_{t=0}^{T_{\max}-1}\mathrm{JS}_\epsilon(\pi_t,p^\star_t).
\]

\section{Derivations and Objective Details}
\label{app:deriv}

\subsection{Reference-prior reward shaping}
\label{app:reward_shaping}

We stabilize exploration by mixing
(i) a frozen reference-LM prior as a log-regularizer and (ii) an external task score. Following Method~\ref{method:gfn-llm}, we represent the LLM-GFlowNet state by a generated prefix.
Let $s_{0:k}$ denote the length-$k$ prefix state (with $s_{0:0}\equiv s_0$ as the root), and let termination occur by emitting $\top$ at $s_{0:\tau}$.
Let $q_{\mathrm{ref}}(\cdot\mid s_{0:k})$ be a frozen reference LM over $\mathcal{V}\cup\{\top\}$.

\paragraph{Reference log-score at a stop cut.}
We define the reference log-probability of stopping at $s_{0:k}$ as
\begin{equation}
\label{eq:app_ref_logP}
\log P_{\mathrm{ref}}(s_{0:k}^{\top})
\triangleq \sum_{t=0}^{k-1}\log q_{\mathrm{ref}}(s_{t+1}\mid s_{0:t}) + \log q_{\mathrm{ref}}(\top\mid s_{0:k})
\end{equation}

\paragraph{Mixed stop-reward and task-only component.}
Given an external task score $S(s_{0:k})$ (e.g., validity/QED/Expr-hit),
we define the mixed log stop-reward
\begin{equation}
\label{eq:app_mixed_reward}
\log R(s_{0:k}^{\top}) \triangleq \kappa \log P_{\mathrm{ref}}(s_{0:k}^{\top}) + \lambda S(s_{0:k}^{\top}),
\end{equation}
where $\kappa$ is a fixed scaling (typically $\kappa{=}1$) and $\lambda$ sets the task term scale
(empirically, $\lambda=50$).
Equivalently, define the \emph{task-only} log component
\begin{equation}
\label{eq:app_task_only_u}
u(s_{0:k}^{\top}) \triangleq \log R(s_{0:k}^{\top}) - \kappa \log P_{\mathrm{ref}}(s_{0:k}^{\top}),
\end{equation}
which is the part we ``absorb'' in RapTB (the reference-derived baseline remains intact).

\paragraph{Ablation: removing the reference prior.}
Table~\ref{tab:smiles_ablation_tb_ref} shows that dropping the reference-prior term can severely destabilize training (here shown for TB on SMILES), causing sharp validity collapse and degenerate length behavior.

\begin{table}[!htbp]
\centering
\small
\setlength{\tabcolsep}{4pt}
\renewcommand{\arraystretch}{1.08}
\caption{
\textbf{SMILES generation ablations (TB reference).}
Unless specified, all metrics are computed on valid samples.
\texttt{Len} denotes the mean token length of valid samples ($L_{\max}=10$).
}
\label{tab:smiles_ablation_tb_ref}
\vspace{-0.35em}
\begin{tabular}{@{}lccccc@{}}
\toprule
Method & Acc $\uparrow$ & Score $\uparrow$ & Entropy $\uparrow$ & FPDiv $\uparrow$ & Len \\
\midrule
TB & \textbf{0.998} & \textbf{0.717} & \textbf{2.503} & \textbf{0.807} & 3.065 \\
TB w/o ref & 0.381 & 0.601 & 0.418 & 0.425 & 10.000 \\
\bottomrule
\end{tabular}
\end{table}

\subsection{Terminable prefix-tree specialization of TB}
\label{app:deriv_rooted}

Following Method~\ref{method:gfn-llm}, $q_\theta(\cdot\mid s_{0:k})$ parametrizes the forward policy over
$\mathcal{V}\cup\{\top\}$. For readability, write
$q_\theta(s_{t+1}\mid \phi_t)$ for token actions and $q_\theta(\top\mid s_{0:k})$ for termination at prefix $s_{0:k}$.
On the prefix tree, each non-root prefix has a unique parent, hence backward factors are deterministic and vanish in log-space.

\paragraph{Terminal TB residual.}
For a realized termination index $\tau$, the TB log-residual is
\begin{equation}
\label{eq:app_tb_llm}
\Delta_{\mathrm{TB}}(\xi)
=
\log Z_\theta
+\sum_{t=0}^{\tau-1}\log q_\theta(s_{t+1}\mid s_{0:t}) + \log q_\theta(\top\mid s_{0:\tau}) - \log R(s_{0:\tau}^{\top})
\end{equation}
We minimize $\mathcal{L}_{\mathrm{TB}}=\mathbb{E}_\xi[\Delta_{\mathrm{TB}}(\xi)^2]$.

\paragraph{Residual at an intermediate prefix.}
In our implementation, the model outputs per-prefix stop logits $\log p_{\mathrm{term}}[k] \equiv \log q_\theta(\top\mid s_{0:k})$
and per-prefix stop-reward logs $\log r[k] \equiv \log R(s_{0:k}^{\top})$ for \emph{all} $k\in\{0,\dots,L{-}1\}$,
including $k=0$ (stopping immediately after the prompt).
Token-transition log-probabilities are stored as $\log p_F[t]\equiv \log q_\theta(s_{t+1}\mid s_{0:t})$ for steps
$t\in\{0,\dots,L{-}2\}$.

Define the TB-style residual at prefix $s_{0:k}$ by: 
\begin{equation}
\label{eq:app_stop_res_k_new}
\Delta_k^{\mathrm{TB}}(\xi)
\triangleq
\log Z_\theta
+\sum_{t=0}^{k-1}\log p_F[t]
+\log p_{\mathrm{term}}[k]
-\log r[k],
\qquad k\in\{0,\dots,L{-}1\},
\end{equation}
where 
$\Delta_0^{\mathrm{TB}}(\xi)=\log Z_\theta+\log p_{\mathrm{term}}[0]-\log r[0]$.

RapTB uses the \emph{rooted} version (cancels $\log Z_\theta$):
\begin{equation}
\label{eq:app_rooted_def_new}
\bar\Delta_k(\xi)
\triangleq
\Delta_k^{\mathrm{TB}}(\xi)-\Delta_0^{\mathrm{TB}}(\xi),
\qquad k\ge 1.
\end{equation}

\subsection{Absorbed suffix backups in RapTB}
\label{app:backup_details}

RapTB absorbs only the task-only component.
The reference term stays unchanged.
For a sampled trajectory $\xi=s_{0:\tau}$, define
\[
u_k\triangleq \lambda S(s_{0:k}^{\top}).
\]

\paragraph{Finite horizon.}
Let $K\in\{1,\dots,L_{\max}\}$ be the auxiliary horizon cap.
We define the auxiliary backup window end as $h\triangleq \min(\tau,K)$.
We compute suffix targets using indices $j\in[k,h]$.

\paragraph{Max and soft backups.}
Define
\begin{align}
\label{eq:app_backup_defs}
u_k^{\max}
&\triangleq \max_{j\in[k,h]} u_j,\\
u_k^{soft}
&\triangleq
\frac{1}{\beta}\log \sum_{j=k}^{h}\exp\!\Big(\beta u_j - \beta\rho\,(j-k)\Big),\qquad \beta>0,\ \rho\ge 0,\\
u_k^{tgt}
&\triangleq
\alpha\,u_k^{\max} + (1-\alpha)\,u_k^{soft},\qquad \alpha\in[0,1].
\end{align}
We compute $u_k^{soft}$ with LogSumExp trick to ensure numerical stability and prevent overflow.

\subsection{Practical details: prefix eligibility, scheduling, and stop gradients}
\label{app:raptb_impl}

\paragraph{Eligibility.}
We compute $\mathcal{L}_{\mathrm{aux}}$ only for prefixes $k\in\{1,\dots,h\}$.
We skip very short prefixes.
We set a minimum depth $k_{\min}\ge 1$.

\paragraph{Absorb terms}
In the main paper we use a mixed stop-reward
\[
\log R(s_{0:k}^{\top})
=
\kappa\log P_{\mathrm{ref}}(s_{0:k}^{\top})
+
u_k,
\qquad
u_k\triangleq \lambda S(s_{0:k}^{\top}).
\]
Absorption modifies only the task-only term.
It keeps the reference term fixed.
For a given prefix $k$, define a surrogate stop-reward
\[
\log \widetilde R(s_{0:k}^{\top})
\triangleq
\kappa\log P_{\mathrm{ref}}(s_{0:k}^{\top})
+
u_k^{tgt}.
\]
Let $\bar\Delta_k(\xi)$ be the rooted residual from Eq.~\eqref{eq:rooted}.
If we recompute the rooted residual at prefix $k$ using $\widetilde R$ (and keep the root term unchanged), we get
\begin{align*}
\bar\Delta^{Rap}_k(\xi)
&=
\bar\Delta_k(\xi)
+
\big(\log R(s_{0:k}^{\top})-\log \widetilde R(s_{0:k}^{\top})\big)\\
&=
\bar\Delta_k(\xi)
+
\big(u_k-u_k^{tgt}\big).
\end{align*}

\paragraph{Absorption Correction with Distance Discounting.}
For an eligible prefix at step $k$, we inject a correction term that pulls the trajectory flow towards the future target. This correction is discounted by the distance to the horizon $h$:
\begin{equation}
    \bar\Delta^{Rap}_k(\xi)= \bar\Delta_k(\xi)+\gamma^{(h-k)} \cdot \big(u_k-u_k^{tgt}\big).
\end{equation}
This ensures that credit assignment decays exponentially as the distance between the decision point and the realized outcome increases.

\paragraph{Auxiliary loss.}
For $k\le h$, let $u_k^{tgt}$ be computed by Eq.~\eqref{eq:app_backup_defs}.
We use
\begin{equation}
\label{eq:app_aux_fix}
\mathcal{L}_{\mathrm{aux}}(\xi)
\triangleq
\frac{\sum_{k=1}^{h} w_k\,\big(\bar\Delta^{Rap}_k(\xi)\big)^2}
{\sum_{k=1}^{h} w_k}.
\end{equation}

\paragraph{Stop-gradient on the termination head in the auxiliary branch.}
We stop gradients through termination log-probabilities only in the auxiliary branch.
Inside $\mathcal{L}_{\mathrm{aux}}$, we replace
\begin{equation}
\label{eq:app_detach_pterm}
\log q_\theta(\top\mid s_{0:k}) \mapsto \sg{\log q_\theta(\top\mid s_{0:k})}.
\end{equation}
The terminal TB loss uses full gradients.

\paragraph{Final objective.}
We optimize
\begin{equation}
\label{eq:app_final_fix}
\mathcal{L}_{\mathrm{RapTB}}
\triangleq
\mathbb{E}_{\xi\sim q_\theta^{\top}}\!\left[
\Delta^{\mathrm{TB}}(\xi)^2 + \eta\,\mathcal{L}_{\mathrm{aux}}(\xi)
\right].
\end{equation}

\begin{algorithm}[t]
\caption{RapTB auxiliary targets (single trajectory)}
\label{alg:raptb}
\begin{algorithmic}[1]
\STATE Roll out a terminable trajectory $\xi=(s_{0:0}\to\cdots\to s_{0:\tau})$.
\STATE Set $h$ as the end of the backup window.
\STATE Compute $u_k=\lambda S(s_{0:k}^{\top})$ for $k\le h$.
\STATE Compute $u_k^{\mathrm{tgt}}$ via Eq.~\eqref{eq:app_backup_defs}.
\STATE Compute $\bar\Delta_k(\xi)$ for eligible $k$.
\STATE In this branch, use $\sg{\log q_\theta(\top\mid s_{0:k})}$.
\STATE Apply the absorbed difference $u_k-u_k^{\mathrm{tgt}}$ when absorption is enabled.
\STATE Compute $\mathcal{L}_{\mathrm{aux}}(\xi)$ via Eq.~\eqref{eq:app_aux_fix}.
\STATE Train with $\mathcal{L}_{\mathrm{TB}}+\eta\mathcal{L}_{\mathrm{aux}}$.
\end{algorithmic}
\end{algorithm}

\subsection{Variance reduction view of RapTB}
\label{app:vr_view}

\paragraph{TB tail as a stochastic regression target.}
Fix a prefix cut at index $m$ along a sampled terminable trajectory
$\xi=(s_{0:0}\to s_{0:1}\to\cdots\to s_{0:\tau})$ on the prefix tree.
Starting from the terminal TB residual in Eq.~\eqref{eq:tb_llm}, we decompose it as
\begin{equation}
\label{eq:app_tb_cut}
\Delta_{\mathrm{TB}}(\xi)=X(s_{0:m}(\xi)) - Y_m(\xi),
\end{equation}
where the \emph{prefix term}
\begin{equation}
\label{eq:app_X_def}
X(s_{0:m}(\xi))
\triangleq
\log Z_\theta
+\sum_{t=0}^{m-1}\log q_\theta(s_{t+1}\mid s_{0:t}),
\end{equation}
 is deterministic given the prefix $s_{0:m}$, and the \emph{tail term}
\begin{equation}
\label{eq:app_Y_def}
Y_m(\xi)
\triangleq
\log R(s_{0:\tau}^{\top})
-\log q_\theta(\top\mid s_{0:\tau})
+\sum_{t=m}^{\tau-1}\log q_\theta(s_{t+1}\mid s_{0:t}),
\end{equation}
depends only on the sampled suffix $(s_{0:m}\to\cdots\to s_{0:\tau})$.

Conditioning on $s_{0:m}$, TB minimizes a conditional least-squares error:
\begin{align}
\label{eq:app_ls_decomp}
\mathbb{E}\!\left[\big(X(s_{0:m})-Y_m(\xi)\big)^2 \,\middle|\, s_{0:m}\right]
=\;&
\big(X(s_{0:m})-\mu(s_{0:m})\big)^2
+ \operatorname{Var}\!\left(Y_m(\xi)\,\middle|\, s_{0:m}\right),
\end{align}
where $\mu(s_{0:m})\triangleq \mathbb{E}[Y_m(\xi)\mid s_{0:m}]$ is the minimum-MSE target for that prefix.

\paragraph{Connecting to RapTB.}
Eq.~\eqref{eq:app_ls_decomp} highlights a core difficulty under terminal rewards: early prefixes are trained through a
single stochastic tail target $Y_m(\xi)$ whose conditional variance can be large, so credit assignment to early decisions is noisy
A natural response is to add subtrajectory consistency, but enforcing arbitrary-start windows introduces state-dependent boundary terms; in
terminable prefix trees, the commonly used SubTB in LLM-GflowNets ties these boundaries to $-\log q_\theta(\top\mid s)$, so heterogeneous
starts combined with discontinuous rewards can be absorbed by the shared termination head, inducing termination/length drift.
RapTB takes a conservative middle ground: it keeps terminal TB unchanged as the global anchor, and adds only \emph{root-start} residuals
$\bar\Delta_k=\Delta_k^{\mathrm{TB}}-\Delta_0^{\mathrm{TB}}$ (Appendix~\ref{app:deriv_rooted}), providing $O(\tau)$ prefix-local supervision without introducing a separate
flow head or arbitrary-start boundary heterogeneity.
To further reduce variance in auxiliary targets, RapTB densifies only the \emph{external} component of the stop-reward by absorbing high-reward
evidence along the sampled suffix (Appendix~\ref{app:backup_details}), while leaving any reference-derived baseline intact; this yields a lower-variance
proxy for the reward term inside rooted constraints without changing the terminal TB semantics.
Finally, we stop gradients through termination logits in the auxiliary branch so these additional prefix constraints cannot be satisfied by globally
shifting $q_\theta(\top\mid s)$, preventing auxiliary-driven length bias.

\subsection{LLM-SubTB analysis}
\label{app:deriv_naive_subtb}

\paragraph{SubTB on terminable prefix trees and the implicit local baseline.}
We use the Subtrajectory Balance (SubTB) objective for autoregressive prefix trees
\citep{madan2023learning,hu2024amortizing}.
For any subtrajectory window indexed by $0\le i<j\le\tau$ along a sampled trajectory $\xi=(s_0\!\to\!\cdots\!\to\!s_\tau)$,
\begin{equation}
\label{eq:app_subtb_window_pf}
\Delta^{\mathrm{SubTB}}_{i\to j}(\xi)
=
\sum_{k=i}^{j-1}\log P_F^\theta(s_{k+1}\mid s_{0:k})
+\big(\log P_F^\theta(\top\mid s_{0:j})-\log P_F^\theta(\top\mid s_{0:i})\big)
+\big(\log R(s_{0:i}^{\top})-\log R(s_{0:j}^{\top})\big),
\end{equation}
and the per-trajectory objective aggregates all window residuals. This form can be viewed as eliminating the state-dependent flow/normalizer $F_\theta(\cdot)$ by using the terminable identity
$R(s^\top)\approx F_\theta(s)\,P_F^\theta(\top\mid s)$ at optimum \citep{hu2024amortizing}; on prefix trees $P_B\equiv 1$, so
$\log P_F^\theta(\top\mid s)$ becomes the only learned term that can represent the missing \emph{local baseline} across different start states.

\paragraph{Termination drift.}
In stop-reward LLM settings, rewards $R(s_{0:k}^{\top})$ are often sparse and weakly prefix-dependent.
When $\log R(s_{0:i}^{\top})\approx \log R(s_{0:j}^{\top})$ for many $(i,j)$, the reward-difference term in
Eq.~\eqref{eq:app_subtb_window_pf} carries little signal, and the window residual is dominated by the token-transition sum and the
termination-logit difference
$\log P_F^\theta(\top\mid s_{0:j})-\log P_F^\theta(\top\mid s_{0:i})$.
Crucially, each $\log P_F^\theta(\top\mid s_{0:t})$ participates in \emph{many} windows (all windows spanning $t$),
so adjusting the termination head provides a high-leverage way to reduce a large number of squared window residuals simultaneously.
As a result, optimization can partially satisfy arbitrary-start consistency by  shifting
$\log P_F^\theta(\top\mid s)$, which miscalibrates stopping probabilities and manifests as length bias / termination drift.

\paragraph{How RapTB blocks this failure channel.}
RapTB prevents auxiliary consistency from being satisfied via termination-head drift in two complementary ways.
First, it restricts auxiliary consistency to \emph{rooted} (start-at-$i=0$) prefixes, which avoids imposing many arbitrary-start
windows that implicitly demand accurate local baselines for each intermediate start state.
Second, in the auxiliary branch it stops gradients through $\log P_F^\theta(\top\mid s)$, so auxiliary constraints cannot be optimized
by moving the termination head; length calibration is instead anchored by the terminal TB objective, while the auxiliary signal focuses
on improving prefix credit assignment without inducing systematic termination shifts.

\section{Implementation Details and Reproducibility}
\label{app:repro}

\paragraph{Config provenance.}
The hyperparameters in Tables~\ref{tab:repro_core_hparams}--\ref{tab:repro_schedulers} are taken from the task configs used to produce our main results.

\subsection{Model, decoding, and context-free grammar (CFG)}
\label{app:repro_cfg}

\paragraph{Backbone and parameterization.}
We fine-tune an autoregressive LLM with a terminable action at every prefix state.
The forward kernel is parameterized as in Sec.~\ref{sec:method} by (i) a token head $p_F(\cdot\mid s)$ and
(ii) a termination head $p_{\mathrm{term}}(s)$.
We use parameter-efficient fine-tuning (LoRA) unless otherwise specified.

\paragraph{LoRA fine-tuning details.}
We use parameter-efficient fine-tuning with LoRA on the LLM backbone.
Concretely, we apply LoRA to the attention and MLP projection modules
\texttt{\{q\_proj, k\_proj, v\_proj, o\_proj, gate\_proj, down\_proj, up\_proj\}}
with rank $r{=}16$, $\alpha{=}16$, dropout $0.1$, and no bias parameters.
The backbone is \texttt{meta-llama/Llama-3.2-1B}.
We optimize with AdamW (lr $10^{-4}$). Any auxiliary schedulers used in training (e.g., temperature/replay schedules)
are reported explicitly in Table~\ref{tab:repro_schedulers}.

\paragraph{Molecular fingerprints for similarity/diversity.}
For SMILES similarity used by SubM, we compute RDKit Morgan fingerprints with radius $2$ and $2048$ bits,
and use Tanimoto similarity for the facility-location diversity term. These settings are fixed across all SMILES experiments.

\paragraph{Context-free grammar (CFG) constrained decoding.}
We enforce syntactic constraints during generation using a grammar processor with an incremental EBNF parser.
At each step, the parser consumes the current prefix and returns the set of next tokens that keep the prefix valid under
the grammar. The processor helps decide tokens only, without masking or modifying their logits.
This is a decoding-time feasibility filter shared by all objectives (TB/SubTB/RapTB), so it does not change the training objective while substantially reducing invalid roll-outs.
Figures~\ref{fig:ebnf_smiles}--\ref{fig:ebnf_expr24} list the grammars used for SMILES and Expr24, respectively.

\begin{figure}[!htbp]
\centering
\begin{minipage}{0.97\linewidth}
\hrule \vspace{3pt}
\scriptsize
\begin{verbatim}
root ::= smiles

smiles ::= atom ( chain | branch )*

chain ::= (dot atom | bond? ( atom | ring_closure ) )+

branch ::= "(" ( ( dot | bond )? smiles )+ ")"

atom ::= organic_symbol | aromatic_symbol | atom_spec | wildcard

bond ::= "" | "=" | "#" | "$" | ":" | "@" | "@@"

dot ::= "."

wildcard ::=  "*"

atom_spec ::= "[" ( "se" | "as" | aromatic_symbol | element_symbol | wildcard ) chiral_class? h_count? ( charge | class? ) "]"

organic_symbol ::= "B" | "C" | "N" | "O" | "P" | "S" | "F" | "I" | "Br" | "Cl" | "At" | "Ts"

aromatic_symbol ::= "b" | "c" | "n" | "o" | "p" | "s"

element_symbol  ::= "A" ( "c" | "g" | "l" | "m" | "r" | "s" | "t" | "u" ) |
                    "B" ( "a" | "e" | "h" | "i" | "k" | "r" )? |
                    "C" ( "a" | "d" | "e" | "f" | "l" | "m" | "n" | "o" | "r" | "s" | "u" )? |
                    "D" ( "b" | "s" | "y" ) |
                    "E" ( "r" | "s" | "u" ) |
                    "F" ( "e" | "l" | "m" | "r" )? |
                    "G" ( "a" | "d" | "e" ) |
                    "H" ( "e" | "f" | "g" | "o" | "s" )? |
                    "I" ( "n" | "r" )? |
                    "K" "r"? |
                    "L" ( "a" | "i" | "r" | "u" | "v" ) |
                    "M" ( "c" | "g" | "n" | "o" | "t" ) |
                    "N" ( "a" | "b" | "d" | "e" | "h" | "i" | "o" | "p" )? |
                    "O" ( "g" | "s" )? |
                    "P" ( "a" | "b" | "d" | "m" | "o" | "r" | "t" | "u" )? |
                    "R" ( "a" | "b" | "e" | "f" | "g" | "h" | "n" | "u" ) |
                    "S" ( "b" | "c" | "e" | "g" | "i" | "m" | "n" | "r" )? |
                    "T" ( "a" | "b" | "c" | "e" | "h" | "i" | "l" | "m" | "s" ) |
                    "U" | "V" | "W" | "Xe" | "Y" "b"? |
                    "Z" ( "n" | "r" )

ring_closure ::= "%" [1-9] [0-9] | [0-9]

chiral_class ::= ( "@" ( "@" | "TH" [1-2] | "AL" [1-2] | "SP" [1-3] | "TB" ( "1" [0-9]? | "2" "0"? | [3-9] ) | "OH" ( "1" 

[0-9]? | "2" [0-9]? | "3" "0"? | [4-9] ) )? )?

charge   ::= "-" ( "-" | "0" | "1" [0-5]? | [2-9] )? | "+" ( "+" | "0" | "1" [0-5]? | [2-9] )?

h_count   ::= "H" [0-9]?

class    ::= ":" [0-9]+
\end{verbatim}
\vspace{2pt}\hrule
\end{minipage}
\caption{\textbf{EBNF grammar used for constrained SMILES decoding}.}
\label{fig:ebnf_smiles}
\end{figure}

\begin{figure}[!htbp]
\centering
\begin{minipage}{0.70\linewidth}
\hrule \vspace{3pt}
\scriptsize
\begin{verbatim}
root  ::= expr4
expr4 ::= num | num op num | num op num op num | num op num op num op num 
| num op num op num op num op num | num op num op num op num op num op num
op    ::= "+" | "-" | "*" | "/"
num   ::= [0-9]
\end{verbatim}
\vspace{2pt}\hrule
\end{minipage}
\caption{\textbf{EBNF grammar used for constrained Expr24 decoding.}}
\label{fig:ebnf_expr24}
\end{figure}

\begin{table}[!htbp]
\centering
\small
\setlength{\tabcolsep}{5pt}
\renewcommand{\arraystretch}{1.10}
\caption{\textbf{Core training and decoding hyperparameters.}
TB and SubTB share all hyperparameters with RapTB except the loss definition (and RapTB-specific coefficients).
}
\label{tab:repro_core_hparams}
\begin{tabular}{lcc}
\toprule
 & \textbf{SMILES} & \textbf{Expr24} \\
\midrule
Trainer steps (max) & 5000 & 5000 \\
Precision & Bf16 & Bf16 \\
Grad clip & 0.5 & 0.5 \\
Grad accumulation & 4 & 4 \\
Optimizer & AdamW ($\mathrm{lr}=10^{-4}$) & AdamW ($\mathrm{lr}=10^{-4}$) \\
\midrule
\textbf{Sampling (train)} &  &  \\
\quad \# trajectories per update ($n_{\mathrm{samples}}$) & 32 & 32 \\
\quad $p_F$ temperature mix & $T_{\mathrm{hi}}:1.5\to 1.0;\ T_{\mathrm{lo}}:0.8\to 1.0$ & $T_{\mathrm{hi}}:1.5\to 1.0;\ T_{\mathrm{lo}}:0.8\to 1.0$ \\
\quad low-temp probability & 0.666 & 0.666 \\
\quad replay mixture ratio & 0.7 & 0.7 \\
\midrule
\textbf{CFG decoding (grammar)} &  &  \\
Min/max length & 1 / 10 (15) & 3 / 9 \\
\bottomrule
\end{tabular}
\end{table}

\begin{table}[!htbp]
\centering
\small
\setlength{\tabcolsep}{4.5pt}
\renewcommand{\arraystretch}{1.10}
\caption{\textbf{RapTB-specific hyperparameters.} $k_{\min}$ is linearly scheduled by training step.}
\label{tab:repro_raptb_hparams}
\begin{tabular}{lcc}
\toprule
 & \textbf{SMILES} & \textbf{Expr24} \\
\midrule
Aux weight $\eta$ & 0.25 & 0.25 \\
Distance discount $\gamma$ & 0.99 & 0.99 \\
Detach $p_{\mathrm{term}}$ in aux & True & True \\
Absorb gate threshold $\varepsilon_{\mathrm{ab}}$ & $10^{-6}$ & $10^{-6}$ \\
Target mode & mix & mix \\
Mix weight $\alpha$ (max vs soft) & 0.5 & 0.8 \\
Soft backup $\beta$ & 5.0 & 3.0 \\
Distance penalty $\rho$ & 0.1 & 0.5 \\
$k_{\min}$ schedule & $5\to 2$ (5000 steps) & $7\to 3$ (5000 steps) \\
Aux horizon cap $K$ & $L_{\max}$ & $L_{\max}$ \\
\bottomrule
\end{tabular}
\end{table}

\subsection{SMILES constraints and validity evaluation}
\label{app:repro_smiles_constraints}

\paragraph{Constrained decoding and legal token list.}
We apply the grammar processor at \emph{every} decoding step.
In addition, we use a fixed allowlist of tokenizer tokens deemed legal for SMILES generation.
The combination of EBNF parsing and token allowed lists substantially reduces invalid generations,
without changing the learning objective.

\paragraph{Validity and scoring.}
We use RDKit-based validation to identify valid SMILES and to compute the task reward (QED in our setup).
Invalid generations receive an invalidity shaping schedule as configured in the reward module.

\subsection{CommonGen Diagnostic Subset Details}
\label{app:commongen_subset}

As discussed in Section~\ref{sec:results_commongen}, we utilize a fixed diagnostic subset to monitor optimization stability. The subset consists of 10 distinct concept-sets selected from the CommonGen validation split~\citep{lin2020commongen}.

Table~\ref{tab:commongen_concepts} lists the specific concept-sets used in our diagnostic experiments. Despite the uniform input size, these concepts cover diverse semantic domains (e.g., physical action, sports, indoor scenes), requiring the model to generate logically coherent sentences with natural termination points.

\begin{table}[h]
\centering
\small
\caption{\textbf{Diagnostic subset of CommonGen.} The table lists the 10 concept-sets used to monitor termination drift and prefix entropy. All sets require relational reasoning to link three distinct concepts into a coherent scenario.}
\label{tab:commongen_concepts}
\vspace{0.5em}
\setlength{\tabcolsep}{8pt}
\renewcommand{\arraystretch}{1.2}
\begin{tabular}{@{}cll@{}}
\toprule
\textbf{ID} & \textbf{Concept Set} & \textbf{Domain Context} \\
\midrule
1 & \{field, look, stand\} & Outdoor / Observation \\
2 & \{kid, room, dance\} & Indoor / Activity \\
3 & \{cat, pet, couch\} & Domestic / Animal \\
4 & \{climb, building, side\} & Urban / Action \\
5 & \{climb, wall, talk\} & Social / Action \\
6 & \{drive, snow, car\} & Travel / Environment \\
7 & \{talk, wear, phone\} & Communication \\
8 & \{hockey, rink, team\} & Sports \\
9 & \{ocean, surfer, surf\} & Nature / Sports \\
10 & \{stair, jump, flight\} & Motion \\
\bottomrule
\end{tabular}
\end{table}

\paragraph{Reward Definition for CommonGen Task}
\label{app:commongen_reward_def}
In the CommonGen task, we define a composite reward function designed to guide the model toward three distinct goals: structural validity, concept coverage, and linguistic quality. First, to ensure grammatical correctness, we enforce strict structural constraints where a generated sequence is deemed valid only if it meets specific criteria—such as containing at least one verb, proper capitalization, and appropriate terminal punctuation. Second, to maximize concept coverage, we compute rewards based on concept coverage, while adding a rigorous lemma-based Hard Coverage Bonus if all target concepts are successfully included. Finally, to promote fluency and semantic alignment with the ground truth, we incorporate a quality-based shaping term that calculates the BLEU scores between the generated prefix and the reference sentence, serving as a step-wise proxy for generation quality relative to human-written text.

\paragraph{Evaluation Protocol on Diagnostic Set.}
For each concept set, we generate $N=320$ samples using the learned policy. We record the raw logits of the termination head ($\log p_{\text{term}}$) at every step to visualize the drift reported in the main text. The small scale of this subset allows for this dense, step-wise instrumentation which would be storage-prohibitive on the full benchmark.

\subsection{Replay buffers}
\label{app:repro_replay_impl}

\paragraph{Reward-prioritized replay buffer (RP).}
We maintain a replay buffer keyed by the decoded prompt string.
For each prompt, the buffer stores up to $B$ trajectories using a min-heap over the reward proxy, thus keeping the current top-$B$ items.
Exact duplicates (identical decoded strings) are discarded.
To prevent near-duplicate high-reward items from dominating the buffer, we additionally apply a
near-duplicate filter using edit distance between the tokenized answers (excluding the termination token):
a new item is rejected if it is too similar to an existing buffer item, unless it has a higher reward
(or is explicitly forced to be added).
When enabled, we use a small buffer-augmentation trick for forced/validated items to ensure they are preferred.
In our configs, the per-prompt buffer capacity is $B=200$, and the near-duplicate tolerance is $0.25$.

\paragraph{Reward-prioritized replay training (PRT).}
The buffer is sufficiently populated, and replay sampling follows a
two-tier scheme inspired by \citet{pmlr-v202-shen23a}:
An $\alpha$ fraction of each replay minibatch is sampled uniformly from the top-$\beta$ reward tier
(i.e., the highest-reward $\lceil \beta|B|\rceil$ items), and the rest is sampled uniformly from the
remaining items; sampling falls back to uniform replay when prioritization is disabled or infeasible.

\subsubsection{Submodular Replay Details}
\label{app:subm_details}

\paragraph{Submodular replay (SubM): objective and selection.}
SubM replaces the ``keep top-$B$ by reward'' rule with a \emph{submodular subset selection} step that refreshes the
buffer to maximize a weighted objective combining quality, diversity, and length coverage.
for each candidate item $x$,
we define a \emph{static score}
$
s(x)=w_{\mathrm{rew}}\,r(x) + w_{\mathrm{val}}\mathbf{1}[\text{valid}(x)]
$
and maximize a facility-location style diversity term together with a concave length-bin coverage term.
Concretely, during greedy selection we use the per-candidate marginal gain
\begin{equation}
\label{eq:subm_gain_impl}
\Delta(x\mid S)
=
s(x)
+
 w_{\mathrm{div}}\sum_{u\in\mathcal{G}}\max\big\{0,\ \mathrm{sim}(u,x)-\mathrm{msim}(u,S)\big\}
+
 w_{\mathrm{len}}\alpha_{b(x)}\Big(\log(1+c_{b(x)}(S){+}1)-\log(1+c_{b(x)}(S))\Big),
\end{equation}
where $\mathrm{sim}(\cdot,\cdot)\in[0,1]$ is a task-dependent similarity, $b(x)$ is the length-bin index of $x$,
$c_b(S)$ is the current count in bin $b$, and $\alpha_b\ge 0$ controls how strongly we bias coverage toward specific length regimes.
In our configs we use uniform weights, so $\alpha_b=1$ for all bins.
Importantly, the implementation uses token length rather than
raw string length to define bins, avoiding the common mismatch between tokenizer length and character length.

\paragraph{Similarity backends.}
For SMILES, we canonicalize valid molecules with RDKit and compute Morgan fingerprints
(radius $2$, $2048$ bits); similarity is Tanimoto computed by RDKit bulk routines.
For Expr24 (string-domain tasks), we use $k$-gram shingles (default $k{=}2$) and Jaccard similarity between shingle sets.
These similarities are only used by the diversity term in Eq.~\eqref{eq:subm_gain_impl}.

\paragraph{Efficiency: cached coverage updates and greedy variants.}
To compute the facility-location marginal efficiently, we maintain $\mathrm{msim}(u,S)$ for each ground element $u\in\mathcal{G}$.
When evaluating a candidate $x$, we compute $\{\mathrm{sim}(u,x)\}_{u\in\mathcal{G}}$ via a bulk similarity call and accumulate
$\sum_u \max\{0,\ \mathrm{sim}(u,x)-\mathrm{msim}(u,S)\}$, then update $\mathrm{msim}(u,S)\leftarrow\max\{\mathrm{msim}(u,S),\mathrm{sim}(u,x)\}$ after selecting $x$.
We use standard greedy algorithm in our submodular replay method.

\paragraph{Validity gating for diversity.}
SubM optionally restricts the diversity optimization to a valid-heavy candidate pool: it keeps all valid items and only adds enough invalid items (ranked by static score)
to ensure the valid proportion is at least the specified ratio.
This prevents invalid strings from consuming the diversity budget while still allowing the buffer to remain populated
when valid samples are scarce.
In SMILES we set the validity-gating ratio to $0.0$, while in Expr24 we set the ratio to $1.0$, so the facility-location term is computed only among valid items.

\begin{table}[!htbp]
\centering
\small
\setlength{\tabcolsep}{5pt}
\renewcommand{\arraystretch}{1.10}
\caption{\textbf{Factor schedulers used in our training runs.}
Each factor uses linear interpolation from \texttt{start} to \texttt{end} over a fixed \texttt{horizon} (in steps).
}
\label{tab:repro_schedulers}
\vspace{-0.25em}
\begin{tabular}{llccc}
\toprule
Task & Factor & start & end & horizon \\
\midrule
SMILES & replay\_buffer & 0.50 & 0.25 & 5000 \\
SMILES & k\_min & 5 & 2 & 5000 \\
\midrule
Expr24 & replay\_buffer & 0.50 & 0.25 & 5000 \\
Expr24 & oracle\_buffer & 0.75 & 0.25 & 5000 \\
Expr24 & k\_min & 7 & 3 & 5000 \\
\bottomrule
\end{tabular}
\end{table}

\begin{table}[!htbp]
\centering
\small
\setlength{\tabcolsep}{5pt}
\renewcommand{\arraystretch}{1.10}
\caption{\textbf{Submodular replay hyperparameters (SMILES).}
We select a buffer of size $B$ using greedy maximization of a facility-location + length objective.
}
\label{tab:repro_subm_smiles}
\begin{tabular}{lc}
\toprule
Hyperparameter & Value \\
\midrule
Buffer capacity $B$ & 200 \\
Similarity backend & RDKit bulk Tanimoto (Morgan r=2, 2048-bit) \\
Weights $(w_{\mathrm{rew}}, w_{\mathrm{val}}, w_{\mathrm{div}}, w_{\mathrm{len}})$ & $(1.0, 1.0, 1.0, 1.0)$ \\
Length bin size &  1 token \\
Length alpha mode / power & uniform / 1.0 \\
Validity gating ratio & 0.0 \\
Selection strategy & standard greedy \\
\bottomrule
\end{tabular}
\end{table}

\begin{table}[!htbp]
\centering
\small
\setlength{\tabcolsep}{5pt}
\renewcommand{\arraystretch}{1.10}
\caption{\textbf{Submodular replay hyperparameters (Expr24).}
We select a buffer of size $B$ using greedy maximization of a facility-location + length objective.
}
\label{tab:repro_subm}
\begin{tabular}{lc}
\toprule
Hyperparameter & Value \\
\midrule
Buffer capacity $B$ & 200 \\
Refresh period $K_{\mathrm{buf}}$ & 1 \\
Similarity backend & $k$-gram shingles + Jaccard \\
Weights $(w_{\mathrm{rew}}, w_{\mathrm{val}}, w_{\mathrm{div}}, w_{\mathrm{len}})$ & $(1.0, 1.0, 1.0, 0.0)$ \\
Length bin size & - \\
Length alpha mode / power & - \\
Validity gating ratio & 1.0 \\
Selection strategy & standard greedy \\
\bottomrule
\end{tabular}
\end{table}

\end{document}